\documentclass[10pt,twocolumn,letterpaper]{article}

\usepackage[pagenumbers]{cvpr} 

\definecolor{cvprblue}{rgb}{0.21,0.49,0.74}
\usepackage[pagebackref,breaklinks,colorlinks,allcolors=cvprblue]{hyperref}

\usepackage{multirow}
\definecolor{mycolor_blue}{HTML}{E7EFFA}
\definecolor{mycolor_green}{HTML}{E6F8E0}
\definecolor{mycolor_gray}{HTML}{ECECEC}
\definecolor{mycolor_red}{HTML}{FFE6E6}
\definecolor{mycolor_yellow}{HTML}{FFFFCC}
\definecolor{mycolor_purple}{HTML}{E6E6FF}

\usepackage{graphicx}
\usepackage{xcolor} 
\usepackage{colortbl}
\usepackage{tcolorbox}  
\usepackage{titletoc}
\newcommand{\worldwideweb}{\raisebox{-1.5pt}{\includegraphics[height=1.05em]{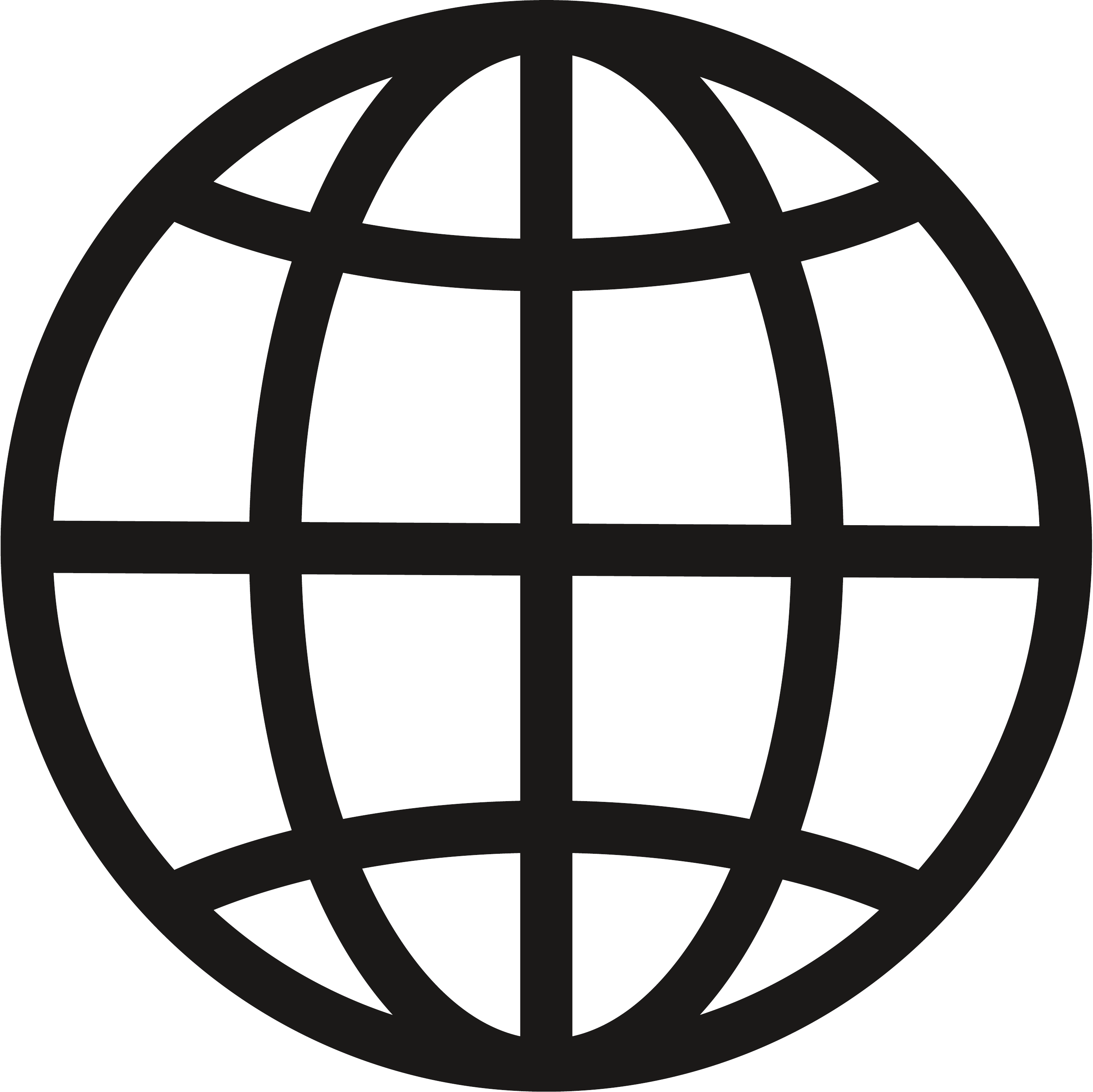}}\xspace}
\newcommand{\github}{\raisebox{-1.5pt}{\includegraphics[height=1.05em]{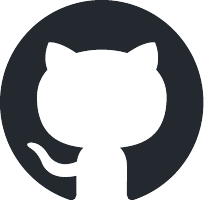}}\xspace}
\newcommand{\huggingface}{\raisebox{-1.5pt}{\includegraphics[height=1.05em]{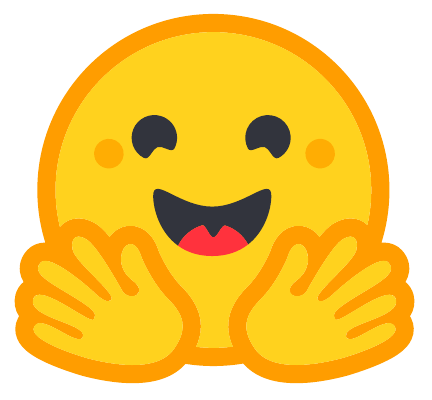}}\xspace}

\title{T2V-CompBench: A Comprehensive Benchmark for \\Compositional Text-to-video Generation}

\author{
  \bf{Kaiyue Sun}\textsuperscript{1} \quad 
  \bf{Kaiyi Huang}\textsuperscript{1} \quad 
  \bf{Xian Liu}\textsuperscript{2}  \quad 
  \bf{Yue Wu}\textsuperscript{3} \quad 
  \bf{Zihan Xu}\textsuperscript{1} \quad 
  \bf{Zhenguo Li}\textsuperscript{3}  \quad 
  \bf{Xihui Liu}\textsuperscript{1} \\
\textsuperscript{1}The University of Hong Kong  \quad
\textsuperscript{2}The Chinese University of Hong Kong  \quad 
\textsuperscript{3}Huawei Noah’s Ark Lab \\
\\
{\worldwideweb \href{https://t2v-compbench-2025.github.io}{{\text{Project Page}}}} \quad \quad 
{\github \href{https://github.com/KaiyueSun98/T2V-CompBench/tree/V2}{{\text{Evaluation Code}}}}
\quad \quad
{\huggingface \href{https://huggingface.co/spaces/Kaiyue/T2V-CompBench_Leaderboard}{{\text{Leaderboard}}}}
}

\begin{document}

\twocolumn[{%
\renewcommand\twocolumn[1][]{#1}%
\maketitle
\vspace{-2em}
\includegraphics[width=\linewidth]{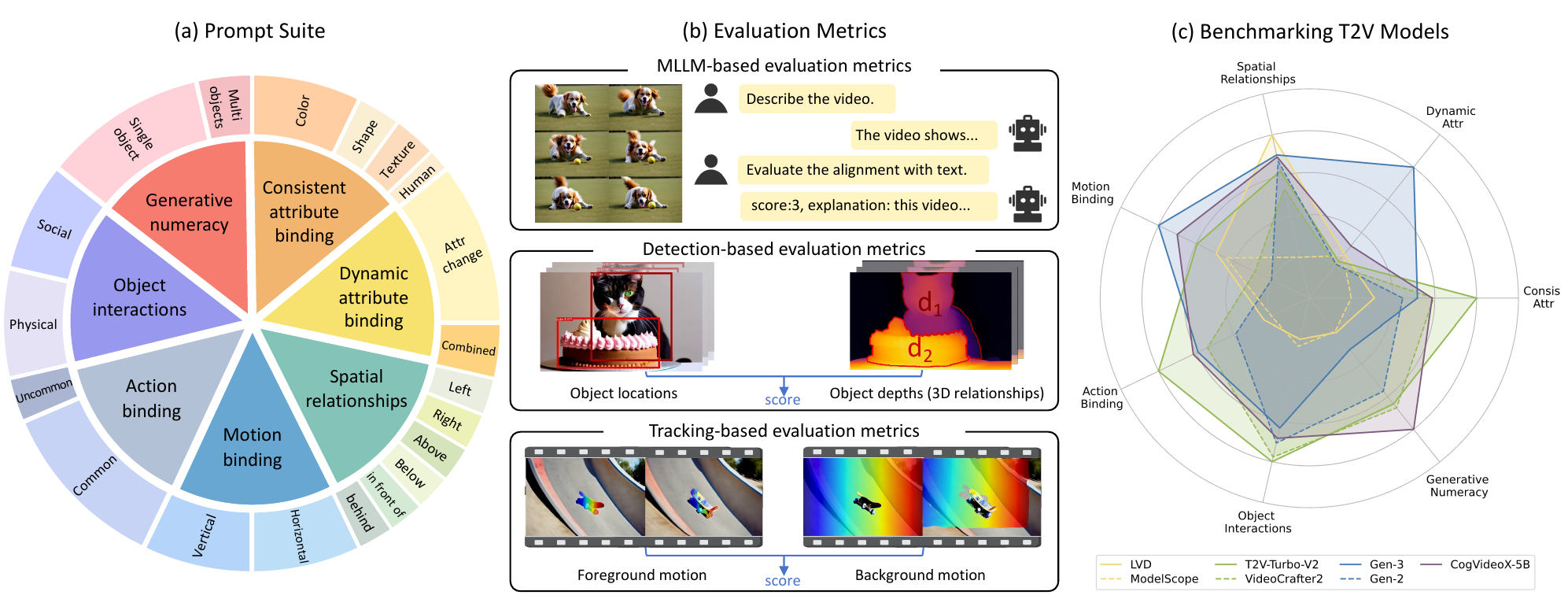}
\captionof{figure}{\textbf{Overview of T2V-CompBench.} We propose T2V-CompBench, a comprehensive compositional text-to-video generation benchmark that consists of seven categories: \textbf{consistent attribute binding, dynamic attribute binding, spatial relationships, motion binding, action binding, object interactions, and generative numeracy}. We propose three types of evaluation metrics: MLLM-based, Detection-based, and Tracking-based metrics. We benchmark various text-to-video generation models.\vspace{2em}
}
\label{fig:teaser}
}]

\renewcommand{\thefootnote}{\fnsymbol{footnote}}

\begin{abstract}
Text-to-video (T2V) generative models have advanced significantly, yet their ability to compose different objects, attributes, actions, and motions into a video remains unexplored. Previous text-to-video benchmarks also neglect this important ability for evaluation. In this work, we conduct the first systematic study on compositional text-to-video generation. We propose T2V-CompBench, the first benchmark tailored for compositional text-to-video generation. T2V-CompBench encompasses diverse aspects of compositionality, including consistent attribute binding, dynamic attribute binding, spatial relationships, motion binding, action binding, object interactions, and generative numeracy. We further carefully design evaluation metrics of multimodal large language model (MLLM)-based, detection-based, and tracking-based metrics, which can better reflect the compositional text-to-video generation quality of seven proposed categories with 1400 text prompts. The effectiveness of the proposed metrics is verified by correlation with human evaluations. We also benchmark various text-to-video generative models and conduct in-depth analysis across different models and various compositional categories. We find that compositional text-to-video generation is highly challenging for current models, and we hope our attempt could shed light on future research in this direction.
\end{abstract}    
\section{Introduction}
\label{sec:intro}
Text-to-video (T2V) generation has made significant progress in recent years~\cite{blattmann2023align, blattmann2023stable, ho2022imagen, singer2022make, wu2021godiva,wu2022nuwa,hong2022cogvideo,villegas2022phenaki, zhou2022magicvideo, khachatryan2023text2video, luo2023videofusion, he2022latent, wang2023modelscope}. However, generating videos that accurately depict multiple objects, attributes, and motions in complex and dynamic scenes based on fine-grained text prompts remains a challenging task. 
In this work, we aim to conduct a systematic study on compositional T2V generation.

Compositional text-to-image (T2I) generation, which aims to compose multiple objects, attributes, and their relationships into complex scenes, has been widely studied in previous methods~\cite{liu2022compositional,feng2022training,chefer2023attend}. Benchmarks for compositional T2I generation~\cite{huang2024t2i} have been accepted as an important evaluation dimension for T2I foundation models~\cite{chen2024pixart, betker2023improving, esser2024scaling}.
However, most works on T2V generation focus on generating videos with simple text prompts, neglecting the significance of compositional T2V generation.
Moreover, existing video generation benchmarks~\cite{huang2023vbench, liu2023evalcrafter, liu2024fetv} primarily evaluate video quality, motion quality, and text-video alignment with single-object text prompts, and benchmarks for compositional T2V generation have not been systematically and extensively investigated in previous literature.

\begin{figure*}[ht]   
  \centering
   \includegraphics[width=1.0\linewidth]{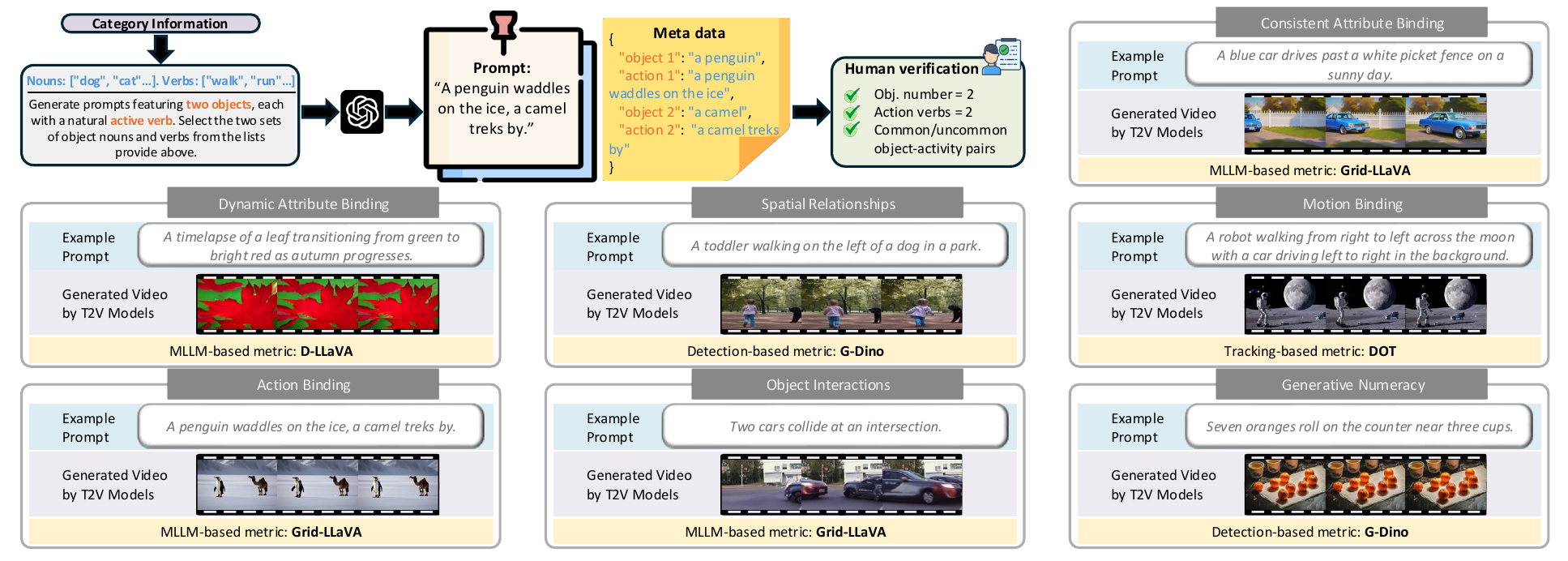}
   \caption{Prompt generation process and illustrations of the seven compositional categories.
   }
   \label{fig:prompt_generation_process}
\end{figure*}

To this end, we propose T2V-CompBench, a comprehensive benchmark designed for compositional text-to-video generation. This benchmark emphasizes compositionality through multiple objects with attributes, quantities, actions, interactions, and spatio-temporal dynamics. 
We design a prompt suite composed of seven categories, where each category consists of 200 text prompts for video generation. When constructing the prompts, we emphasize temporal dynamics and guarantee that each prompt contains at least one active verb. The seven categories are as follows and examples are illustrated in Figure~\ref{fig:prompt_generation_process}:
\textbf{1) Consistent attribute binding.} This category includes prompts featuring two objects, each with a distinct attribute. The attributes associated with each object are consistent throughout the video.
\textbf{2) Dynamic attribute binding.} Prompts in this category focus on dynamic attribute binding for objects, where the attributes change over time.
\textbf{3) Spatial relationships.} In this category, each prompt mentions two objects and specifies the spatial relationship between them. 
\textbf{4) Motion binding.} 
Each prompt in this category includes one or two objects and a moving direction is specified for each object.
\textbf{5) Action binding.} Prompts in this category describe two objects, each with a distinct action.
\textbf{6) Object interactions.} This category tests the models' abilities to understand and generate dynamic interactions between multiple objects, including physical interactions and social interactions.
\textbf{7) Generative numeracy.} The text prompts in this category include one or two objects with quantities ranging from one to eight. %

Another challenge lies in the evaluation of compositional T2V generation. Commonly used metrics, such as Inception Score~\cite{salimans2016improved}, Fr\'echet Inception Distance (FID)~\cite{heusel2017gans}, Fr\'echet Video Distance (FVD)~\cite{unterthiner2019fvd}, and CLIPScore~\cite{hessel2021clipscore}, cannot fully reflect the compositionality of T2V generation models. 
Evaluating compositionality of T2V models requires a fine-grained understanding of not only objects and attributes in each frame but also the dynamics and motions across frames. It is orders of magnitude more complex than evaluating T2I models.

To address this challenge, we take temporal dynamics across frames into consideration and design different metrics to evaluate different categories in the benchmark. Specifically, we design multimodal large language model (MLLM)-based metrics, including image-LLM and video-LLM, to evaluate consistent attribute binding, dynamic attribute binding,  action binding, and object interactions. We devise detection-based metrics to evaluate spatial relationships and generative numeracy. We propose tracking-based metrics to evaluate motion binding. The effectiveness of our proposed metrics is validated by computing the correlation with human evaluations. We evaluate various T2V generation models on T2V-CompBench and analyze their performances on different compositional categories.

The contributions of our paper are three-fold. \textbf{1)} To our best knowledge, we are the first to propose a benchmark for compositional text-to-video generation, featuring seven categories with 1400 text prompts. \textbf{2)} We propose comprehensive evaluation metrics for the seven categories and verify their effectiveness through correlation with human evaluations. \textbf{3)} We benchmark various T2V models and provide a systematic study with insightful analysis, which will inspire future research in this direction.

\section{Related Work}
\label{sec:related_work}
\subsection{Text-to-video Generation.} 
Recently, diffusion models have gained significant attention in the field of text-to-video generation, building on the success of text-to-image models. 
Existing text-to-video diffusion models can be roughly categorized into two types, namely the diffusion unet-based~\cite{ho2022imagen, singer2022make, zhou2022magicvideo, khachatryan2023text2video, luo2023videofusion, blattmann2023align, he2022latent, wang2023modelscope} and the Diffusion Transformer (DiT)- based~\cite{ma2024latte,opensora, pku_yuan_lab_and_tuzhan_ai_etc_2024_10948109, yang2024cogvideox, genmo2024mochi}. 
In this paper, we evaluate a list of text-to-video diffusion models, including officially open-sourced models: ModelScope~\cite{wang2023modelscope}, ZeroScope, LVD~\cite{lian2023llm}, AnimateDiff~\cite{guo2023animatediff}, MagicTime~\cite{yuan2024magictime}, Show-1~\cite{zhang2023show}, VideoCrafter2~\cite{chen2024videocrafter2}, VideoTetris~\cite{tian2024videotetris}, Vico~\cite{yang2024vico}, T2V-Turbo-V2~\cite{li2024t2vturbo}, Latte~\cite{ma2024latte}, Open-Sora~\cite{opensora}, Open-Sora-Plan~\cite{pku_yuan_lab_and_tuzhan_ai_etc_2024_10948109},  CogVideoX-5B~\cite{yang2024cogvideox}, and Mochi~\cite{genmo2024mochi}, and commercial models: Pika-1.0~\cite{pika}, Gen-2~\cite{gen2}, Gen-3~\cite{gen3}, Dreamina 1.2~\cite{Dreamina}, PixVerse-V3~\cite{PixVerse}, and Kling-1.0~\cite{kling}.
This comprehensive evaluation ensures the diversity of T2V approaches and provides insights into their capabilities and limitations across different architectures. 

\subsection{Compositional Text-to-image Generation.}
Recent studies have delved into compositionality in text-to-image generation~\cite{liu2022compositional,feng2022training,li2022stylet2i,wu2023harnessing, huang2024t2i, patel2023eclipse, liu2024referee, chefer2023attend,park2021benchmark, lian2023llm, chen2024training, wang2023compositional, meral2023conform, kim2023dense, rassin2024linguistic, gani2023llm, li2023gligen, wang2024divide, yang2024mastering}.
T2I-CompBench~\cite{huang2024t2i} proposed the first comprehensive benchmark to evaluate compositionality in text-to-image models, focusing on attributes binding, relationships, and complex compositions. While these evaluations are tailored exclusively to the image domain, video generation requires a deeper consideration of spatio-temporal dynamics. 
Our work pioneers the development of benchmarking compositional text-to-video generation. 

\subsection{Benchmarks for Text-to-video Generation.}
Previous methods evaluate T2V models from the perspectives of video quality and text-video alignment. For video quality, the commonly used metrics, such as Inception Score (IS)~\cite{salimans2016improved} and Fr\'echet Video Distance (FVD)~\cite{unterthiner2019fvd} are adopted to evaluate video diversity and fidelity. For text-video alignment, CLIPScore~\cite{hessel2021clipscore} is proposed to measure the similarity of the text prompt and the frames, using the pre-trained CLIP model~\cite{radford2021learning}. However, these metrics are not suitable for complex compositional prompts.

Recent T2V benchmarks design text prompts and metrics to evaluate the video quality and text-video alignment in open domains.
VBench~\cite{huang2023vbench} and EvalCrafter~\cite{liu2023evalcrafter} propose comprehensive benchmarks to evaluate T2V models from various perspectives. 
FETV~\cite{liu2024fetv} categorizes prompts based on major content, controllable attributes, and prompt complexity. 
ChronoMagic-Bench~\cite{yuan2024chronomagic} evaluates T2V models' ability in generating time-lapse videos.
However, most prompts provided in these benchmarks focus on single-object rather than composition of multiple objects. Although some of them involves evaluation dimensions that include multiple objects, their prompts, such as ``a bird and a cat'' in ``multiple objects'' dimension of VBench~\cite{huang2023vbench}, do not reflect the dynamics in videos.
ChronoMagic-Bench~\cite{yuan2024chronomagic} emphasizes the generation of natural metamorphic time-lapse videos, while excluding unnatural attribute-change videos, which are also essential for evaluating the design and creative capabilities of T2V models.
A comprehensive definition of compositionality in text-to-video generation is currently lacking in the literature. Therefore, we introduce the first benchmark for evaluating compositional text-to-video generation, with tailored evaluation metrics that we validate through extensive human correlation studies. 
\section{Benchmark Construction}
\label{sec:new_dataset}

\subsection{Problem Definition and Categorization}
Compositional T2V generation has not been comprehensively explored in prior research. Therefore, we first clarify the problem definition and categorization. Previous literature on compositional T2I generation~\cite{chefer2023attend,feng2022training,huang2024t2i} typically focus on attribute binding, object relationships and numeracy. However, in the context of compositional T2V, we need to consider the composition in both spatial and temporal dimensions. In spatial dimension, we follow the framework established in compositional T2I and define categories of consistent attribute binding, spatial relationships, and numeracy. These categories require the generated video frames consistently align with the text prompts. In temporal dimension, we introduce categories of dynamic attribute binding, motion binding, action binding, and object interactions. These categories specifically address whether the temporal dynamics of the video follow the description of the text prompts. 

\subsection{Prompt Categories}
\noindent \textbf{Consistent Attribute Binding}. We define four attribute types including color, shape, texture and human-related attributes. 
Each prompt has two objects, two attributes, and at least one active verb, with each attribute associates with a specific object.
Among all the prompts, 20\% are challenging and uncommon cases that aim to test the model's ability to generalize to unseen combinations, such as ``Blue apple bouncing near a pink tree''.

\noindent \textbf{Dynamic Attribute Binding}. This category focuses on how the attributes of objects change over time. For example, ``Green avocado darkens to black as the tomato beside it ripens to a deep red''. We define four aspects inspired by TempCompass~\cite{liu2024tempcompass}: color \& light change, shape \& size change, texture change, and combined change. 80\% of the prompts describe common attribute changes in real world, while the remaining 20\% are less common or artificial.

\noindent \textbf{Spatial Relationships}.
This category requires the model to generate two objects with specified spatial relationships across the video. We define six types of spatial relationships: ``on the left of'', ``on the right of'', ``above'', ``below'', ``in front of'', and ``behind''. We construct contrastive ``left'' and ``right'' prompts by inverting the spatial relationships.

\noindent \textbf{Motion Binding}. 
Prompts in this category contain one or two objects with specified moving directions. We define four types of moving directions: ``leftwards'', ``rightwards'', ``upwards'' and ``downwards''. Each object in the prompt moves in one of the directions. 

\noindent \textbf{Action Binding}.
This category tests the models' abilities to bind actions to corresponding objects when there are multiple objects and multiple actions described. Two objects are involved in the prompt, each engaged in an activity. This category contains 80\% common prompts and 20\% uncommon prompts, which can be further divided into uncommon object coexistence and uncommon object-activity pairs.

\noindent \textbf{Object Interactions}. 
This category tests the models’ abilities to understand and generate dynamic interactions, including physical interactions causing motion change or state change and social interactions between live entities. 

\noindent \textbf{Generative Numeracy}. 
To analyze the models' abilities to generate correct number of objects, we construct prompts in quantity groups. In each group, the same noun is paired with different quantities, such as ``Three/Four/Five dogs run through a field''. We also construct prompts with two objects like ``Three/Four cows graze in the pasture, and one/two sheep wanders/wander nearby''.

\begin{figure}
    \centering
    \vspace{-5pt}
    \includegraphics[width=\linewidth]{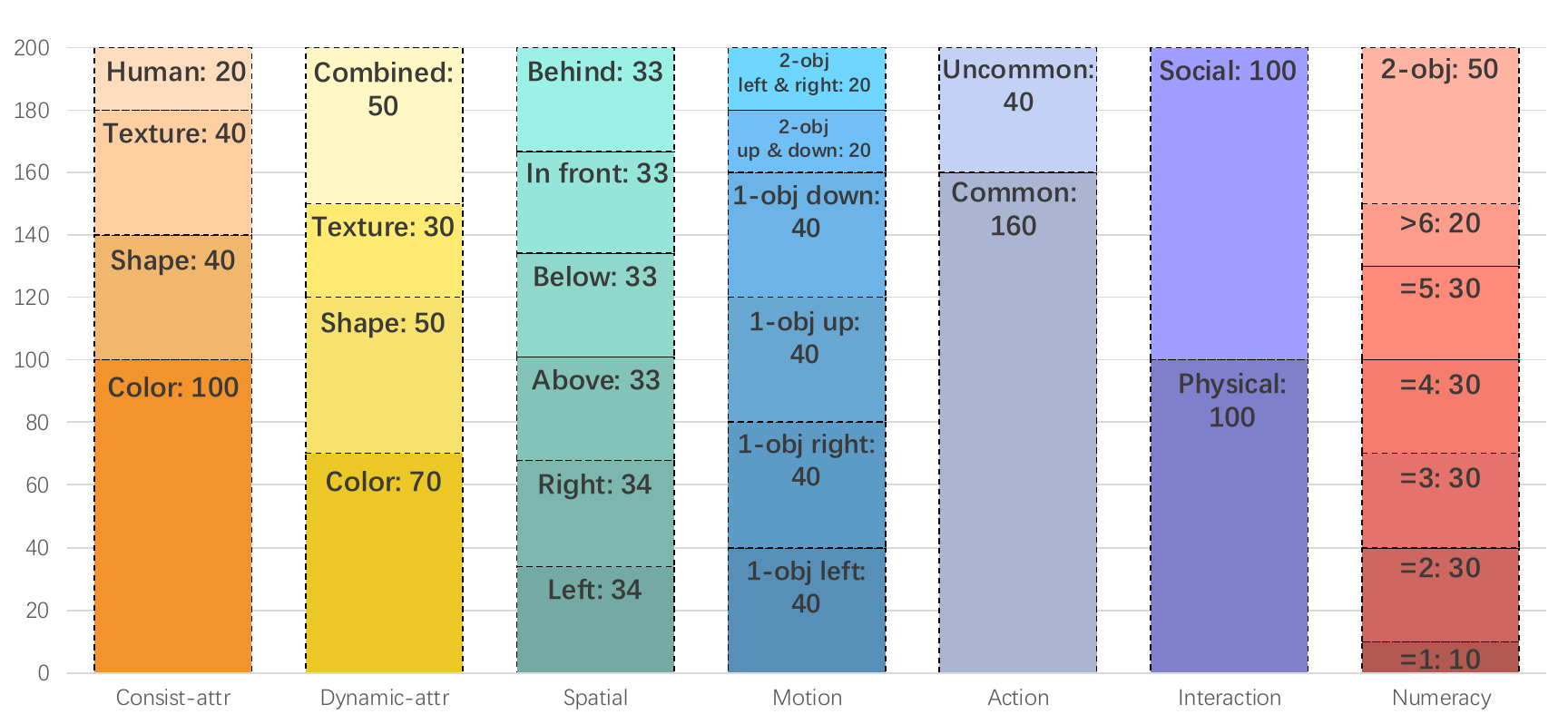}
    \vspace{-10pt}
    \caption{\textbf{Illustration of prompt categories.} We show the number of prompts for the seven categories and their respective subgroups.}
    \vspace{-10pt}
    \label{fig:prompt_group}
\end{figure}

We prepare 200 prompts for each of the seven categories.
Figure~\ref{fig:prompt_group} displays the prompt subgroups along with the number of prompts associated with each subgroup.

\subsection{Prompt Suite Generation}
\begin{figure}[htbp]
    \begin{minipage}[t]{0.5\linewidth}
        \centering
        \includegraphics[width=\textwidth]{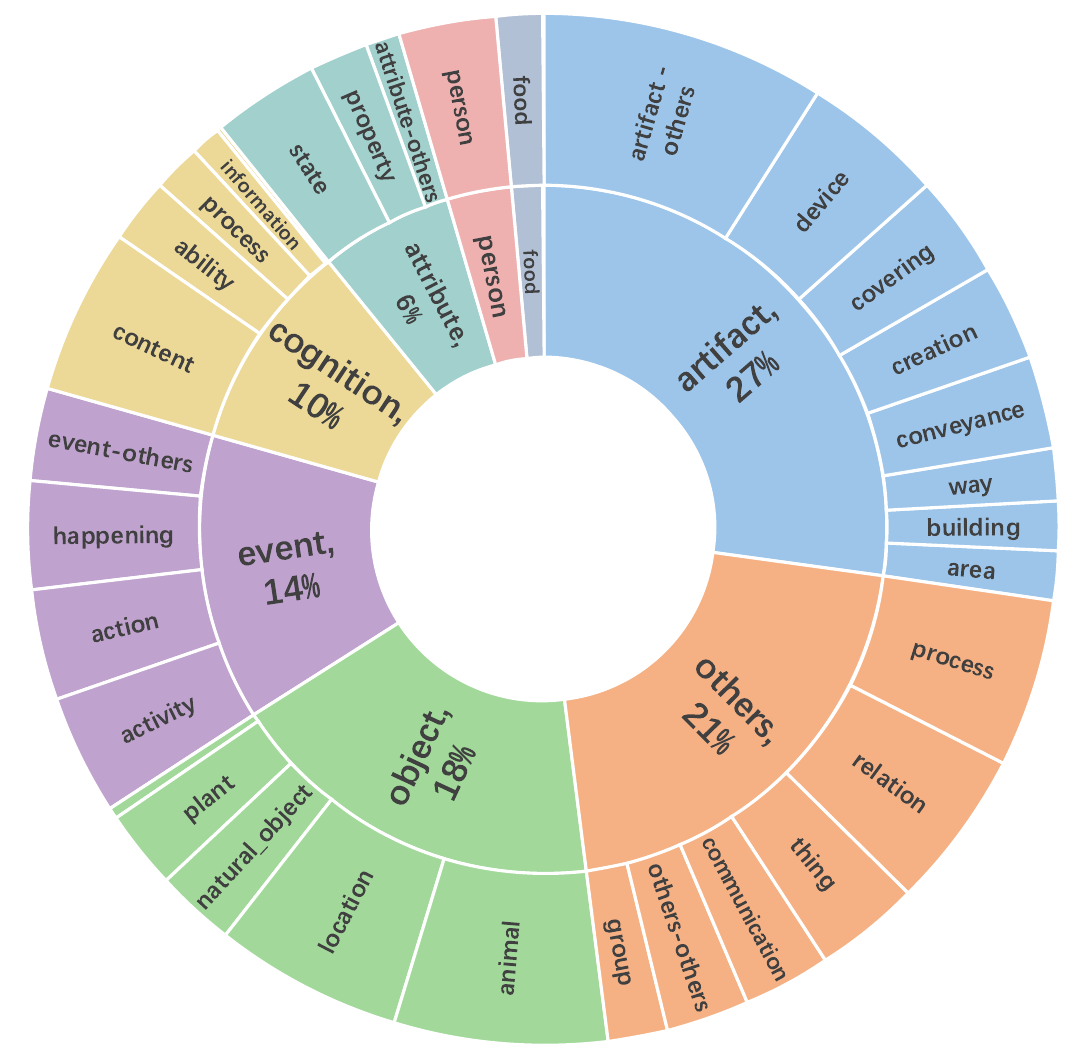}
        \centerline{(a) Meta types of nouns}
    \end{minipage}%
    \begin{minipage}[t]{0.5\linewidth}
        \centering
        \includegraphics[width=\textwidth]{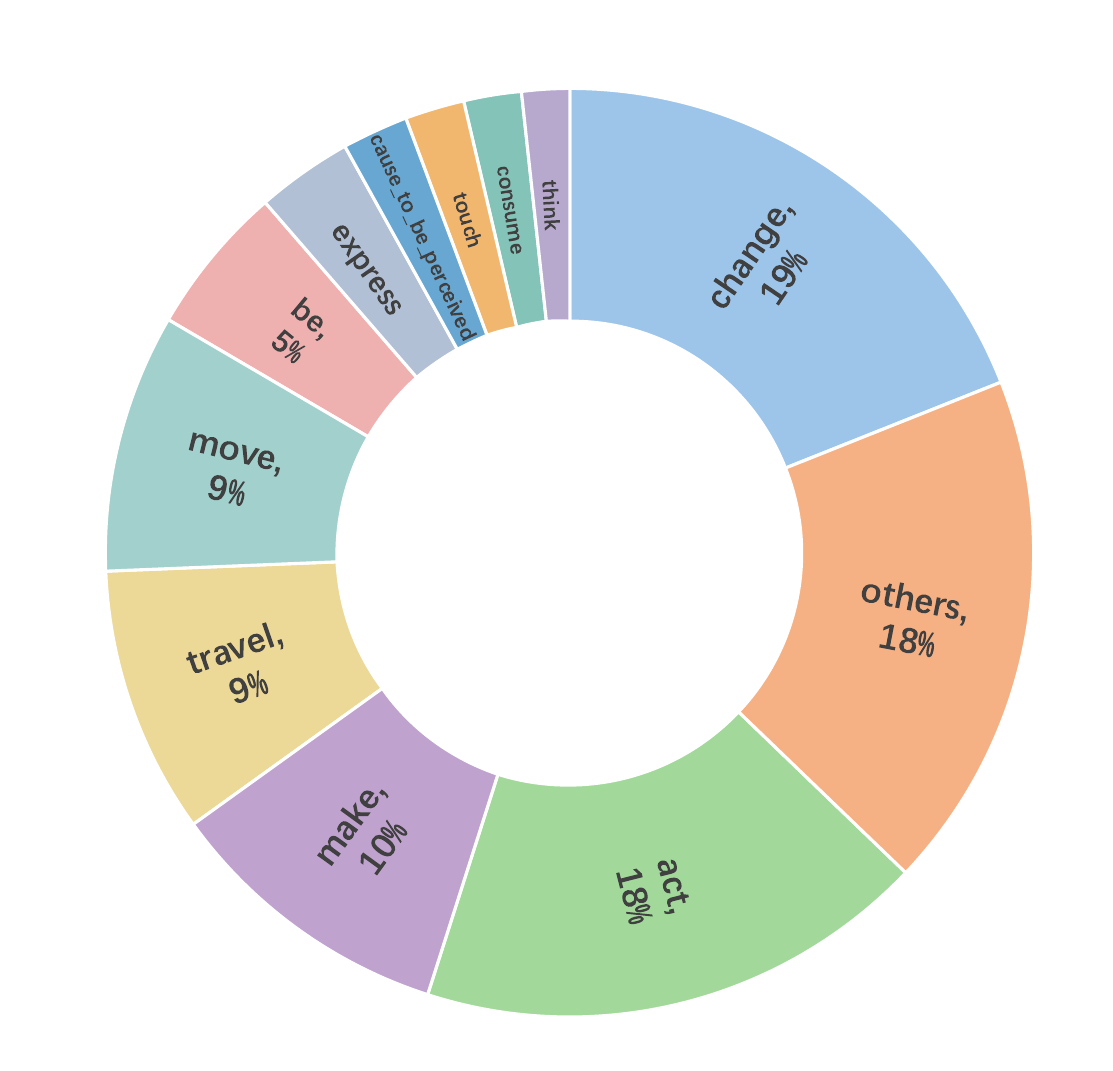}
        \centerline{(b) Meta types of verbs}
    \end{minipage}
    \caption{\textbf{Word distributions of real-user prompts.} We show the types of nouns and verbs of real-user prompts from VidProM~\cite{wang2024vidprom}.}
\label{fig:vidprom}
\end{figure}

\noindent \textbf{Vocabulary Construction}.
To ensure our prompt suite matches the real users' desires, we analyze 1.67 million unique T2V prompts collected by VidProM~\cite{wang2024vidprom} from Pika Discord channels. We use WordNet~\cite{miller1995wordnet} to identify the metaclasses of nouns and verbs, and their distributions are visualized in Figure~\ref{fig:vidprom}. Firstly, as our goal is to benchmark T2V models ability to compose multiple concepts, we focus on entry-level objects that occur with high frequency. Secondly, to facilitate evaluation, we select ``thing'' categories, i.e. individual objects that can be easily labeled with bounding boxes like ``car'' and ``dog'', rather than ``stuff'' categories, i.e. objects without clear boundaries such as ``sky''. Based on these principles, we analyze real user prompts to identify high-frequency nouns that belong to specific metaclasses, such as person, artifacts (e.g., conveyance and device), animals, plants, and food. 
We identify high-frequency verbs and attributes in a similar way. To account for the inherent dynamics of videos, we pair objects with vibrant active verbs, like those in the metaclasses of ``move'', ``travel'', and ``act'', while avoiding static actions like ``think'', ``see'' or ``rest''. Attributes are also drawn from relevant metaclasses. In total, we collected 260 object nouns, 200 active verbs, and 80 attributes in color, shape, and texture. The prompt suite is generated using these words that reflect the interests of real users. For more details about word selection, please refer to Appendix~\ref{app:word_selection}.

\noindent\textbf{Prompt Generation.} Since we evaluate compositional T2V generation, the prompts must follow specified requirements. As a result, sourcing from existing dataset captions is challenging. Additionally, using a fixed template for automatic prompt generation is not ideal, since real user prompts are free-form and diverse. Therefore, we have opted to use GPT-4~\cite{openai2024gpt4} to generate the prompts.
The 200 prompts in each category are generated by prompting GPT-4~\cite{openai2024gpt4} with the collected high-frequency words and the specific requirements for the category. Although not all prompt categories are designed to evaluate actions and motions, we ensure all prompts in our benchmark contain at least one active verb, to prevent the T2V model from generating static videos. GPT-4~\cite{openai2024gpt4} returns both prompts and the parsed meta information for the prompts that facilitates evaluation. 
All the generated prompts are verified by humans, and improper prompts are filtered out. For more details about prompt generation, please refer to Appendix~\ref{app:promptgen_llm}.

\subsection{Prompt Suite Statistics} 
\begin{figure}[htbp]
   \centering
   \vspace{-5pt}
   \includegraphics[width=\linewidth]{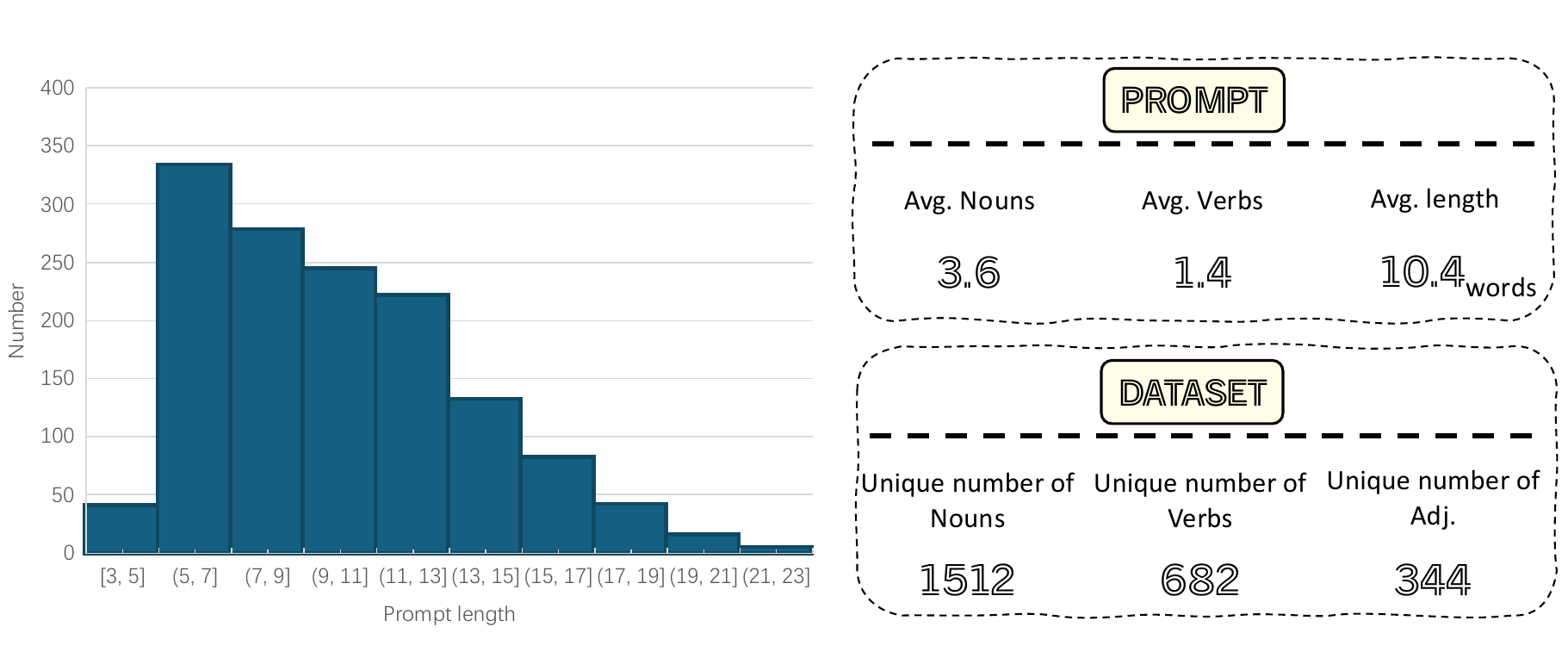}
   \vspace{-10pt}
   \caption{\textbf{Statistics on benchmark prompts.} \textbf{\textit{Left}}: Prompt length distribution. \textbf{\textit{Right}}: Prompt Suite statistics.}
   \vspace{-10pt}
   \label{fig:prompt_statistics}
\end{figure}

T2V-CompBench stands out for its focus on multiple objects and temporal dynamics. Figure~\ref{fig:prompt_statistics} shows the statistics on benchmark prompts: (1) In contrast to previous benchmarks, which predominantly focus on single-object, our prompts involve more than two nouns on average, with each prompt containing approximately 3.6 nouns. (2) T2V-CompBench considers temporal dynamics, with all prompts containing verbs, averaging at 1.4 per prompt. (3) To prevent T2V models from being distracted by irrelevant contents, we avoid using excessively long prompts, the average length is 10.4 words, ranging from 3 to 23. (4) Since we utilize an LLM to generate free-form prompts, many auxiliary words, such as those representing time and spot, are produced in addition to the provided words we instruct the LLM to select from. As a result, there is a rich variety of nouns, verbs, and adjectives present in the generated dataset, ensuring it encompasses a diverse range of topics. For more information about T2V-CompBench prompts, please refer to Appendix~\ref{t2v-compbench_metaclass}

\section{Evaluation Metrics}
\label{sec:evaluation_metrics}
We observe that the evaluation metrics for compositional T2I generation~\cite{huang2024t2i} cannot be directly adopted for evaluating compositional T2V generation, due to the large number of frames and complex spatio-temporal dynamics in videos. 
Most T2V models generate short videos in 2-5 seconds. For a fair comparison, we evenly extract 6 frames for MLLM-based evaluation, and 16 frames for detection-based evaluation, and sample the videos to a frame rate of 8 frames per second (FPS) for tracking-based evaluation. Figure~\ref{fig:teaser}(b) illustrates the three types of evaluation metrics.

\subsection{MLLM-based Evaluation Metrics}
Multimodal Large Language Models (MLLMs) have shown great capabilities in understanding complex contents in images and videos~\cite{zhu2023minigpt4,liu2023improvedllava,openai2024gpt4,xu2024pllava}. Inspired by their effectiveness in video understanding, we exploit MLLMs as evaluators for compositional text-to-video generation. 

\noindent \textbf{Video LLM-based metrics for consistent attribute binding, action binding, and object interactions.}
To handle the complex spatio-temporal information in videos, we investigate video LLMs such as Image Grid~\cite{kim2024imagegrid} and PLLaVA~\cite{xu2024pllava}, which extends LLaVA~\cite{liu2023improvedllava} from single image input to multi-frame input.
We empirically find that Image Grid performs better than PLLaVA in our compositional categories.
Specifically, Image Grid uniformly samples 6 frames from the video to form an image grid as the input to LLaVA~\cite{liu2023improvedllava}. Additionally, we boost the ability of video LLMs and avoid hallucinations by the chain-of-thought mechanism~\cite{wei2022chain} along with disentangled questions, where we first ask the MLLM to describe the video content, and then request it to grade each aspect of the text-video alignment. We denote this metric as \textit{Grid-LLaVA}.
\begin{itemize}
\item To evaluate consistent attribute binding, we use GPT-4~\cite{openai2024gpt4} to parse the prompts into disentangled phrases (\textit{e.g.}, ``A blue car drives past a white picket fence on a sunny day'' is parsed into ``a blue car'' and ``a white picket fence''), and then ask the video LLM to assign a matching grade for each disentangled phrase in relation to the Image Grid. The grades for both phrases are combined and averaged to produce a final numerical score.
\item For action binding, we use GPT-4~\cite{openai2024gpt4} to extract objects and their actions.
For example, given the prompt ``A dog runs through a field while a cat climbs a tree'', we extract the phrases ``a dog'', ``a dog runs through a field'', ``a cat'', and ``a cat climbs a tree''. We then ask the video LLM to check the presence of objects and evaluate the alignment between each object-action pair and the Image Grid.
\item For object interactions, we prompt the video LLM to check the presence of objects and then assess the quality of the interaction based on the text. This evaluation includes examining dynamics of objects, the overall development and outcome of the interaction process.
\end{itemize}

\noindent  \textbf{Image LLM-based metrics for dynamic attribute binding.} 
Evaluating dynamic attribute binding, such as the prompt ``Bright green leaf wilts to brown'' is challenging as it requires a deep understanding of dynamic changes across frames.
We find that current video LLMs perform poorly in this area, so we develop a frame-by-frame evaluation metric based on an Image LLM such as LLaVA~\cite{liu2023improvedllava}.
We utilize GPT-4~\cite{openai2024gpt4} to parse the initial state (``bright green leaf'') and the final state (``brown leaf''). We then prompt LLaVA~\cite{liu2023improvedllava} to score the alignment between each frame and each of the two states. Our scoring function is designed to encourage the first frame to align with the initial state, the last frame to align with the final state, and the middle frames to be in between. 
We denote this metric as \textit{D-LLaVA}.

For more details about using MLLMs as evaluation metrics, please refer to Appendix~\ref{app:mllm}.

\subsection{Detection-based Evaluation Metrics}
Most vision-language models face difficulties with spatial relationships and numeracy-related understandings. So we introduce the object detection model GroundingDINO (\textit{G-Dino})~\cite{liu2023grounding} to detect objects for each frame, filter out duplicate bounding boxes with high intersection-over-union (IoU), and then define rule-based metrics based on the object detection results.

\noindent  \textbf{2D Spatial Relationships.} For 2D spatial relationships including ``left'', ``right'', ``above'', and ``below'', we define rule-based metrics for each frame similar to T2I-CompBench~\cite{huang2024t2i}. Specifically, for each pair of objects, we denote their centers as $(x_1, y_1)$ and $(x_2, y_2)$, respectively. The first object is on the left of the second object if $x_1 < x_2,$ and $|x_1-x_2| > |y_1-y_2|$.
The rule is similar for other 2D spatial relationships. If there is more than one pair of objects in a frame, we select the most probable one based on their IoU and confidence scores. The per-frame score is ($1-$IoU) if there is an object pair that satisfies the relationship, or 0 if no object pair satisfies the relationship. The video-level score is the average of per-frame scores.

\noindent  \textbf{3D Spatial Relationships.} 3D spatial relationships (``in front of'', ``behind'') cannot be identified by purely 2D bounding boxes. With the 2D boxes detected by GroundingDINO~\cite{liu2023grounding}, we further leverage Segment Anything~\cite{kirillov2023seganysam} to predict masks of the specified objects and then leverage Depth Anything~\cite{yang2024depth} to predict depth maps. The depth of an object is defined as the average depth values of the pixels inside the object mask. We define per-frame score based on the IoU and relative depth between two objects, and the video-level score is the average of per-frame scores.

\noindent \textbf{Generative Numeracy.} To evaluate generative numeracy, 
we count the number of objects detected for each object class. If the detected quantity matches the number in text prompt, we assign a score of 1 for that object class. Otherwise, we assign a score of 0. The frame-level score is calculated as the average score of all object classes, and the video-level score is the average of all frames.

\subsection{Tracking-based Evaluation Metrics}
The evaluation metric for motion binding aims to identify the moving direction of objects in videos. However, in many cases, object motions are entangled with camera motions, making it difficult to determine the true direction of an object's movement.
In videos, the actual moving direction of an object is the relative moving direction between the foreground object and the background. Therefore, we introduce a tracking-based method to determine the moving directions of the foreground and background separately.
Specifically, we use GroundingSAM~\cite{ren2024groundedsam} to obtain masks for the foreground objects and the background. Then, we apply DOT~\cite{moing2023dense} to track points in both the foreground and background throughout the video. We compute the average motion vectors for the points in both the foreground and background, and the difference between these two vectors gives us the actual moving direction of the object. The final score reflects whether this actual moving direction aligns with the motion described in the text prompts. We denote this metric as \textit{DOT}.
\section{Experiments}
\label{sec:experiments}
 
\subsection{Evaluated Text-to-video Models}
We evaluate the performance of 23 T2V models on T2V-CompBench, including 17 open-source models and 6 commercial models.
Among these models,
\begin{itemize}
\item \textbf{ModelScope}~\cite{wang2023modelscope} is a diffusion unet-based T2V generation model that evolves from a T2I model by incorporating spatio-temporal blocks, ensuring consistent frame generation and smooth movement transitions.

\item \textbf{AnimateDiff}~\cite{guo2023animatediff} features a plug-and-play motion module that can be applied to diffusion unet-based T2I models to produce temporally smooth animations.

\item \textbf{Show-1}~\cite{zhang2023show} is a high-quality T2V diffusion model with unet backbone that initially employs pixel-based diffusion to create a low-resolution video, followed by a latent-based approach to upsample the video to high resolution.

\item \textbf{VideoCrafter2}~\cite{chen2024videocrafter2} is a diffusion unet-based video generation model that utilizing low-quality videos and synthesized high-quality images to create a video model with high visual quality and precise text-video alignment.

\item \textbf{Latte}~\cite{ma2024latte} is a novel Diffusion Transformer (DiT)-based video generation model, which extracts spatio-temporal tokens from input videos and utilizes Transformer blocks to model video distribution in latent space. 

\item \textbf{Open-Sora}~\cite{opensora} and \textbf{Open-Sora-Plan}~\cite{pku_yuan_lab_and_tuzhan_ai_etc_2024_10948109} are both DiT-based open-source video generation projects, aiming at reproducing Sora.

\item \textbf{CogVideoX-5B}~\cite{yang2024cogvideox} is a DiT-based large scale T2V model designed to generate significant movement.

\item \textbf{Mochi}~\cite{genmo2024mochi} is also a large scale T2V model built on Asymmetric DiT architecture, with emphasis on quality of motion and prompt following.

\item \textbf{LVD}~\cite{lian2023llm} is specifically designed to leverage LLM-guided layout planning for videos with multiple objects. Here, we use the version adapted from ModelScope~\cite{wang2023modelscope}. 
\textbf{ZeroScope}~\cite{Zeroscope} is optimized from ModelScope~\cite{wang2023modelscope} to create high-quality 16:9 compositions and smooth videos.

\item \textbf{VideoTetris}~\cite{tian2024videotetris} is proposed to handle compositional T2V generation that involves multiple objects and dynamic changes in object numbers.
\textbf{Vico}~\cite{yang2024vico} is a framework that explicitly ensures that complex compositional interactions between multiple concepts and actions can be represented properly.
\textbf{T2V-Turbo-V2}~\cite{li2024t2vturbo} introduces a consistency distillation process of a T2V model from a pretrained T2V model by integrating various supervision signals. These three model are all adapted or distilled from VideoCrafter2~\cite{chen2024videocrafter2}. 

\item \textbf{MagicTime}~\cite{yuan2024magictime} is trained from AnimateDiff~\cite{guo2023animatediff}. It is designed to generate metamorphic time-lapse videos, so we specifically test it on dynamic attribute binding.
\end{itemize}

\noindent To organize the T2V models, we categorize them into five groups: 
\textbf{1) DiT-based models:} Latte~\cite{ma2024latte}, Open-Sora 1.1 and 1.2~\cite{opensora}, Open-Sora-Plan v1.0.0 and v1.3.0~\cite{pku_yuan_lab_and_tuzhan_ai_etc_2024_10948109}, CogVideoX-5B~\cite{yang2024cogvideox} and Mochi~\cite{genmo2024mochi}. 
\textbf{Models with the same foundation model:} 
\textbf{2)} ModelScope~\cite{wang2023modelscope}, ZeroScope~\cite{Zeroscope}, and LVD~\cite{lian2023llm}; \textbf{3)} AnimateDiff~\cite{guo2023animatediff} and MagicTime~\cite{yuan2024magictime}; 
\textbf{4)} Videocrafter2~\cite{chen2024videocrafter2}, VideoTetris~\cite{tian2024videotetris}, Vico~\cite{yang2024vico}, and T2V-Turbo-V2~\cite{li2024t2vturbo}.\footnote{Show-1~\cite{zhang2023show} and videocrafter2~\cite{chen2024videocrafter2} are concurrent diffusion unet-based models, so we place them together.}
\textbf{5)} \textbf{Commercial models:} Pika-1.0~\cite{pika}, Gen-2~\cite{gen2}, Gen-3~\cite{gen3}, Dreamina 1.2~\cite{Dreamina}, PixVerse-V3~\cite{PixVerse}, and Kling-1.0~\cite{kling}.

We follow the official default implementations of these T2V models in evaluation, please refer to Appendix~\ref{app:implementation}.

\subsection{Evaluation Metrics}
\begin{table*}[htbp]
\centering
\vspace{5pt}
\resizebox{\linewidth}{!}{%
\begin{tabular}{l cccc cccc cccc cc}
\toprule
\multirow{2}{*}{ \textbf{Metric}} & \multicolumn{2}{c}{\textbf{Consist-attr}} & \multicolumn{2}{c}{\textbf{Dynamic-attr}} & \multicolumn{2}{c}{\textbf{Spatial}} & \multicolumn{2}{c}{\textbf{Motion}} & \multicolumn{2}{c}{\textbf{Action}} & \multicolumn{2}{c}{\textbf{Interaction}} &\multicolumn{2}{c}{\textbf{Numeracy}}\\
\cmidrule(lr){2-3}\cmidrule(lr){4-5}\cmidrule(lr){6-7}\cmidrule(lr){8-9}\cmidrule(lr){10-11}\cmidrule(lr){12-13} \cmidrule(lr){14-15}
 & $\tau$($\uparrow$) & $\rho$($\uparrow$) & $\tau$($\uparrow$) & $\rho$($\uparrow$) & $\tau$($\uparrow$) & $\rho$($\uparrow$)& $\tau$($\uparrow$) & $\rho$($\uparrow$) & $\tau$($\uparrow$) & $\rho$($\uparrow$) & $\tau$($\uparrow$) & $\rho$($\uparrow$) &
 $\tau$($\uparrow$) & $\rho$($\uparrow$) \\
    \midrule
    \hspace{-.5em} \textit{conventional metrics} \\
    CLIP&$0.3667$&$0.4859$&$-0.0096$&$-0.01401$&$0.2395$&$0.3343$&$0.1381$&$0.1818$&$0.2796$&	$0.3799$& $0.1187$ & $0.1587$ &$0.0560$&$0.0821$  \\
    
    B-CLIP &$0.2609$&$0.3562$&$0.2100$&$0.2917$&$0.1247$&$0.1647$&$-0.0582$&$-0.0889$&$0.0915	$&$0.1246$& $0.1455$ & $0.2103$ &$0.0694$&$0.0829$\\

    B-BLEU &$0.2030$&$0.2777$&$0.0854$&$0.1041$&$0.1006$&$0.1396$&$-0.1450$&$-0.1978$&$0.2505$&$0.3577$& $0.1040$ & $0.1479$ &$0.1275$&$0.1770$ \\
    
    B-VQA&$0.5194$&$0.6964$&$-$&$-$&$-$&$-$&$-$&$-$&$-$&$-$&$-$&$-$&$-$&$-$\\
    ViCLIP&$0.4520$&$0.6116$&$0.0079$&$0.0074$&$0.2257$&$0.3222$&$0.0834$&$0.1144$&$0.2481$&$0.3361$& $0.2308$ & $0.3229$ &$0.1036$&$0.1421$ \\
    
\midrule
\hspace{-.5em} \textit{MLLM-based metrics} \\
    PLLaVA &$0.2715$&$0.3105$&$0.1845$&$0.2201$&$-0.1252$&$-0.1509$&$0.0401$&$0.0498$&$0.4326$&$0.5066$& $-0.2233$ & $-0.2607$  &$0.2253$&$0.3142$ \\

    LLaVA & $0.6373$ & $0.7461$ & $-$ & $-$ &$0.4258$&$0.5332$&$-$&$-$& $0.5272$ & $0.6714$ & $0.4213$  & $0.5358$ &$0.3212$&$0.4540$\\
     
    \textbf{Grid-LLaVA} &$\textbf{0.6636}$&$\textbf{0.7893}$& $0.1435$ & $0.1678$ &$0.4815$&$0.5763$&$0.1349$&$0.1619$&$\textbf{0.5969
}$&$\textbf{0.7353}$&$\textbf{0.4557}$&$\textbf{0.5925}$&$0.2266$&$0.2809$ \\

     \textbf{D-LLaVA} &$-$&$-$&$\textbf{0.4362}$&$\textbf{0.5061}$&$-$&$-$&$-$&$-$&$-$&$-$&$-$&$-$&$-$&$-$\\

\midrule
\hspace{-.5em} \textit{detection-based metrics} \\
    VPEval-S&$-$&$-$&$-$&$-$&$0.3137$&$0.3853$&$-$&$-$&$-$&$-$&$-$&$-$&$-$&$-$ \\
    M-GDino&$-$&$-$&$-$&$-$&$-$&$-$&$0.2059$&$0.2306$&$-$&$-$&$-$&$-$&$-$&$-$ \\

    \textbf{G-Dino} &$-$&$-$&$-$&$-$&$\textbf{0.5769}$&$\textbf{0.7057}$&$-$&$-$&$-$&$-$&$-$&$-$&$\textbf{0.4063}$&$\textbf{0.5378}$ \\

\midrule
\hspace{-.5em} \textit{tracking-based metric} \\
    \textbf{DOT} &$-$&$-$&$-$&$-$&$-$&$-$&$\textbf{0.4523}$&$\textbf{0.5366}$&$-$&$-$&$-$&$-$&$-$&$-$ \\
    \bottomrule
\end{tabular}%
}
\caption{\textbf{The correlation between automatic evaluations and human evaluations.} Our proposed evaluation metrics show enhanced performance in Kendall's $\tau$ and Spearman's $\rho$.
} 
\label{tab:human_corr}
\end{table*}

\noindent \textbf{Conventional Metrics.}
We compare our proposed metrics with five metrics widely used in previous studies: \textbf{1)} CLIPScore~\cite{hessel2021clipscore} (denoted as \textit{CLIP}) calculates the cosine similarity between CLIP text and image embeddings. 
\textbf{2)} BLIP-CLIP~\cite{chefer2023attend} (denoted as \textit{B-CLIP}) applies BLIP~\cite{li2022blip} to generate captions for images, and then calculates the text-text cosine similarity between the CLIP embeddings of captions and input prompts. 
\textbf{3)} BLIP-BLEU~\cite{liu2023evalcrafter} 
(denoted as \textit{B-BLEU}) employs BLIP2~\cite{li2023blip} for caption generation, then calculates the BLEU~\cite{papineni2002bleu} similarity between the captions and input prompts. Here, we use the same implementation as EvalCrafter~\cite{liu2023evalcrafter}, which averages five captions generated by BLIP2~\cite{li2023blip}.
\textbf{4)} BLIP-VQA~\cite{huang2024t2i} (denoted as
\textit{B-VQA}) leverages the visual question answering (VQA) ability of BLIP2~\cite{li2022blip} to evaluate the text-image alignment, focusing specifically on attribute binding.
The video-level scores of these metrics are calculated by averaging across all frames.
We also include \textbf{5)} \textit{ViCLIP} score~\cite{huang2023vbench}, which measures text-video alignment by calculating the similarity between text and video features extracted by ViCLIP model~\cite{wang2023internvid}.

In addition, VPEval~\cite{Cho2023VPT2I} introduces detection-based metrics using GroundingDINO~\cite{liu2023grounding} to evaluate spatial relationships in T2I generation. VideoDirectorGPT~\cite{Lin2023VideoDirectorGPT} adapts VPEval~\cite{Cho2023VPT2I} to evaluate object movement direction in videos by obtaining the object locations in the first and last frames. 
Therefore, we include the detection-based metrics from VPEval~\cite{Cho2023VPT2I}: \textbf{6)} \textit{VPEval-S}, and from VideoDirectorGPT~\cite{Lin2023VideoDirectorGPT}: \textbf{7)} \textit{M-GDino}, as compared metrics to evaluate the spatial relationships and motion binding in our benchmark. Similarly, \textit{VPEval-S} is adapted for T2V evaluation by averaging the scores over all frames. 

\noindent \textbf{Our Proposed Metrics.}
As introduced in Section~\ref{sec:evaluation_metrics}, 
the image LLM-based metric \textit{D-LLaVA} is designed to evaluate dynamic attribute binding.
The detection-based metric \textit{G-Dino} is designed for spatial relationships and numeracy.
The tracking-based metric \textit{DOT} is designed for motion binding.
Additionally, we test the video LLM-based metrics \textit{Grid-LLaVA} and \textit{PLLaVA} for all categories. We also test the image LLM-based metric \textit{LLaVA}, which evaluates the text-video alignment on a frame basis for all categories except dynamic attribute binding and motion binding.

In next section, we identify the best metric for each category by analyzing the correlation between results given by these metrics and human annotators.

\subsection{Human Evaluation Correlation Analysis}
In this section, we conduct human evaluations and compute the correlation between scores from automatic metrics and humans to identify the best metric for each category.

\noindent \textbf{Human Evaluation.} 
For the human evaluation of each category, we randomly select 15 prompts out of 200 prompts and use 6 T2V models to generate a total of 90 videos. Additionally, we include 10 ground truth videos for dynamic attribute binding and 11 for object interactions. The total number of videos for human evaluation is 651. We employ the platform of Amazon Mechanical Turk, where we ask three annotators to score the text-video alignment for each video. We then average across the three scores for each text-video pair and calculate the correlation between these human scores and the automatic evaluation scores with Kendall's $\tau$ and Spearman's $\rho$. For more details about human evaluation, please refer Appendix~\ref{app:human_evaluation}.

\noindent \textbf{Comparisons across Evaluation Metrics.}
The correlation results are presented in Table~\ref{tab:human_corr}. These results validate the effectiveness of our proposed metrics, which are highlighted in bold. Specifically, Grid-LLaVA excels in consistent attribute binding, action binding and object interactions, D-LLaVA is the best for dynamic attribute binding, G-Dino is the top metric for spatial relationships and generative numeracy, and DOT is the most reliable for motion binding.
In contrast, CLIP, ViCLIP, B-CLIP and B-BLEU show relatively low human correlations, indicating their limitations in capturing the fine-grained attributes, spatial relationships and dynamics in complex videos. Although Grid-LLaVA is the top metric for three categories, its performance is marginally inferior to those best metrics in other categories in capturing temporal changes in attributes, understanding spatial relationships, motion direction, and numeracy. 
LLaVA performs reasonably well across all the categories it evaluates, which reveals its abilities in capturing attributes, spatial layout, numeracy and object relationships. It falls short compared to Grid-LLaVA in action binding and object interactions.  This is because Grid-LLaVA has the advantage of analyzing multiple frames simultaneously, which allows it to account for temporal changes rather than relying solely on a static frame.

\begin{table*}[tbp]
\centering
\resizebox{\linewidth}{!}{%
\begin{tabular}
{llccccccccccccccccccc}
\toprule 
\multicolumn{2}{c}{{\textbf{Model}}} & \multicolumn{1}{c}{\textbf{Consist-attr}} & \multicolumn{1}{c}{\textbf{Dynamic-attr}} & \multicolumn{1}{c}{\textbf{Spatial}} & \multicolumn{1}{c}{\textbf{Motion}} & \multicolumn{1}{c}{\textbf{Action}} & \multicolumn{1}{c}{\textbf{Interaction}} & \multicolumn{1}{c}{\textbf{Numeracy}}    \\
\cmidrule(lr){1-2} \cmidrule(lr){3-3} \cmidrule(lr){4-4} \cmidrule(lr){5-5} \cmidrule(lr){6-6} \cmidrule(lr){7-7} \cmidrule(lr){8-8} \cmidrule(lr){9-9}
\multicolumn{2}{c}{Metric} & Grid-LLaVA   & D-LLaVA    & G-Dino & DOT  & Grid-LLaVA & Grid-LLaVA & G-Dino  \\
\midrule
\multicolumn{2}{l}{\hspace{-.5em} \textit{diffusion unet-based}} \\
& ModelScope~\cite{wang2023modelscope}                                       & 0.5148 & 0.0161 &  0.4118  &  0.2408  & 0.3639 & 0.4613 & 0.1986 \\
& ZeroScope~\cite{Zeroscope}                                                 & 0.4011 & 0.0091 &  0.4287  &  0.2454  & 0.3661 & 0.4196 & 0.2408 \\
& LVD~\cite{lian2023llm}                                                     & 0.5439 & \cellcolor{mycolor_blue}{0.0171} &  \cellcolor{mycolor_blue}{0.5405}  &  0.2457  & 0.3802 & 0.4502 & 0.2008 \\
\cmidrule{2-9}
& AnimateDiff~\cite{guo2023animatediff}                                      & 0.4325 & 0.0097 &  0.3920  &  0.2227  & 0.2844 & 0.3970 & 0.1767 \\
& MagicTime~\cite{yuan2024magictime}                                         &    -   & 0.0151 &    -     &    -     &   -    &    -   &   -    \\
\cmidrule{2-9}
& Show-1~\cite{zhang2023show}                                                & 0.5670 & 0.0115 &  0.4544  &  0.2291  & 0.3881 & 0.6244 & 0.3086 \\
& VideoCrafter2~\cite{chen2024videocrafter2}                                 & 0.6182 & 0.0103 &  0.4838  &  0.2259  & 0.5030 & 0.6365 & 0.3330 \\
& VideoTetris~\cite{tian2024videotetris}                                              & 0.6211 & 0.0104 &  0.4832  &  0.2249  & 0.4939 & \cellcolor{mycolor_blue}{0.6578} & \cellcolor{mycolor_blue}{0.3467} \\
& Vico~\cite{yang2024vico}                                                   & 0.5887 & 0.0107 &  0.4974  &  0.2219  & 0.5111 & 0.5957 & 0.3230 \\
& T2V-Turbo-V2~\cite{li2024t2vturbo}                                         & \cellcolor{mycolor_blue}{0.6723} & 0.0127 &  0.5025  &  \cellcolor{mycolor_blue}{0.2556}  & \cellcolor{mycolor_blue}{0.6087} & 0.6439 & 0.3261 \\
\midrule
\multicolumn{2}{l}{\hspace{-.5em} \textit{DiT-based}} \\
& Latte~\cite{ma2024latte}                                                   & 0.4713 & 0.0080 &  0.4340  &  0.2155  & 0.4146 & 0.4146 & 0.2320 \\
& Open-Sora 1.1~\cite{opensora}                                              & 0.5414 & 0.0109 &  0.5406  &  0.2261  & 0.5037 & 0.5565 & 0.2259 \\
& Open-Sora 1.2~\cite{opensora}                                              & 0.5639 & 0.0189 &  0.5063  &  0.2468  & 0.4833 & 0.5039 & \cellcolor{mycolor_yellow}{0.3719} \\
& Open-Sora-Plan v1.0.0~\cite{pku_yuan_lab_and_tuzhan_ai_etc_2024_10948109}  & 0.4246 & 0.0086 &  0.4520  &  0.2148  & 0.4009 & 0.4150 & 0.2331 \\
& Open-Sora-Plan v1.3.0~\cite{pku_yuan_lab_and_tuzhan_ai_etc_2024_10948109}  & 0.6076 & 0.0119 &  0.5162  &  0.2377  & 0.4524 & 0.4483 & 0.2952 \\
& CogVideoX-5B~\cite{yang2024cogvideox}                                      & \cellcolor{mycolor_yellow}{0.6164} & 0.0219 &  0.5172  &  \cellcolor{mycolor_yellow}{0.2658}  & \cellcolor{mycolor_yellow}{0.5333} & \cellcolor{mycolor_yellow}{0.6069} & 0.3706 \\
& Mochi~\cite{genmo2024mochi}                                                & 0.5973 & \cellcolor{mycolor_yellow}{0.0246} &  \cellcolor{mycolor_yellow}{0.5480}  &  0.2334  & 0.4759 & 0.5381 & 0.2718 \\
\midrule
\midrule
\multicolumn{2}{l}{\hspace{-.5em} \textit{commercial}} \\
& Pika-1.0~\cite{pika}                                                       & 0.5536 & 0.0128 &  0.4650  &  0.2234  & 0.4250 & 0.5198 & 0.3870 \\
& Gen-2~\cite{gen2}                                                          & 0.5795 & 0.0109 &  0.5126  &  0.2173  & 0.4413 & 0.6144 & 0.3039 \\
& Gen-3~\cite{gen3}                                                          & 0.5980 & \cellcolor{mycolor_red}{\textbf{0.0687}} &  0.5194  &  0.2754  & 0.5233 & 0.5906 & 0.2306 \\
& Dreamina 1.2~\cite{Dreamina}                                                   & 0.6913 & 0.0051 &  0.5773  &  0.2361  & 0.5924 & 0.6824 & 0.4380 \\
& PixVerse-V3~\cite{PixVerse}                                                & \cellcolor{mycolor_red}{\textbf{0.7060}} & 0.0624 &  \cellcolor{mycolor_red}{\textbf{0.5979}}  &  \cellcolor{mycolor_red}{\textbf{0.2867}}  & \cellcolor{mycolor_red}{\textbf{0.8722}} & \cellcolor{mycolor_red}{\textbf{0.8309}} & \cellcolor{mycolor_red}{\textbf{0.6066}} \\
& Kling-1.0~\cite{kling}                                                     & 0.6931 & 0.0098 &  0.5690  &  0.2562  & 0.5787 & 0.7128 & 0.4413 \\

\bottomrule
\end{tabular}%
}
\caption{\textbf{T2V-CompBench evaluation results using proposed metrics}. Scores are normalized between 0 and 1. A higher score indicates better performance.  \textbf{Bold} signifies the highest score within each category. \colorbox{mycolor_blue}{Blue} highlights the top score among diffusion unet-based models. \colorbox{mycolor_yellow}{Yellow} highlights the top score among DiT-based models. \colorbox{mycolor_red}{Red} highlights the top score among commercial models.
}
\label{tab:benchmark}
\end{table*}
\subsection{Quantitative Evaluation}
The performance of the models in T2V-CompBench evaluated by our proposed metrics is shown in Table~\ref{tab:benchmark}. A comparison of different models reveals the following findings:

\noindent \textbf{\#1: Top Performers.} 
T2V-Turbo-V2~\cite{li2024t2vturbo} exhibits the best overall performance in diffusion unet-based models.
CogVideoX-5B~\cite{yang2024cogvideox} shows the best overall performance in DiT-based models.
PixVerse-V3~\cite{PixVerse} excels in all models.

\noindent \textbf{\#2: Adapted Models.}
Models adapted from VideoCrafter2~\cite{chen2024videocrafter2}, including VideoTetris~\cite{tian2024videotetris}, Vico~\cite{yang2024vico}, and T2V-Turbo-V2~\cite{li2024t2vturbo}, improves in most of the categories.
MagicTime~\cite{yuan2024magictime} shows a significant improvement in dynamic attribute binding compared to AnimateDiff~\cite{guo2023animatediff}.
LVD~\cite{lian2023llm} demonstrates enhancements in almost all categories compared to ModelScope~\cite{wang2023modelscope}. Its design, which leverages LLM-guided layout planning, allows it to excel in spatial relationships in diffusion unet-based models.

\noindent \textbf{\#3: Models with Strong Dynamic Capabilities.}
Among open-source models, T2V-Turbo-V2\cite{li2024t2vturbo} and CogVideoX-5B\cite{yang2024cogvideox} excel in motion binding and action binding. They show strong performance in dynamic attribute binding and interactions as well. The commercial models Gen-3\cite{gen3} and PixVerse-V3\cite{PixVerse} lead all models in dynamic attribute binding and motion binding. This can be attributed to their capabilities to generate videos with a high dynamic degree.

\noindent \textbf{\#4: Other Noteworthy Results.}
Other notable models include Open-Sora 1.2~\cite{opensora}, which excels in numeracy among open-source models, and VideoTetris~\cite{tian2024videotetris}, which demonstrates strong capabilities in numeracy and interaction among diffusion unet-based models.

\subsection{Qualitative Evaluation}
\begin{figure*}[h]
    \centering
    \vspace{-20pt}
    \includegraphics[width=0.88\linewidth]{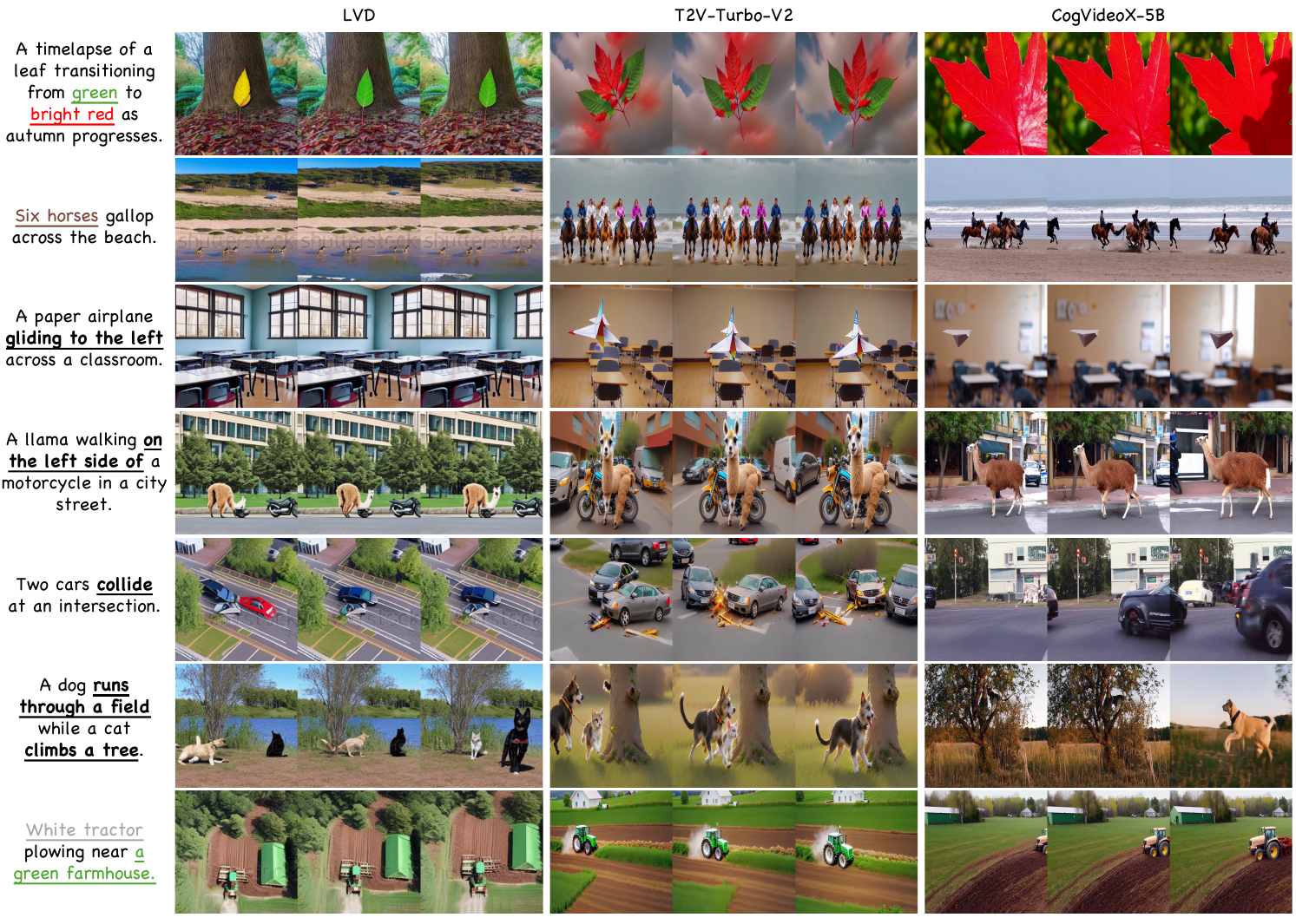}
    \vspace{-7pt}
    \caption{\textbf{Qualitative comparison with different open-source T2V models for the seven compositional categories in T2V-CompBench.} We show text-to-video examples generated by LVD~\cite{lian2023llm}, T2V-Turbo-V2~\cite{li2024t2vturbo}, and CogVideoX-5B~\cite{yang2024cogvideox}.}
    \vspace{-4pt}
    \label{fig:qualitative_1}
\end{figure*}

\begin{figure*}[h]
    \centering
    \vspace{0pt}
    \includegraphics[width=0.88\linewidth]{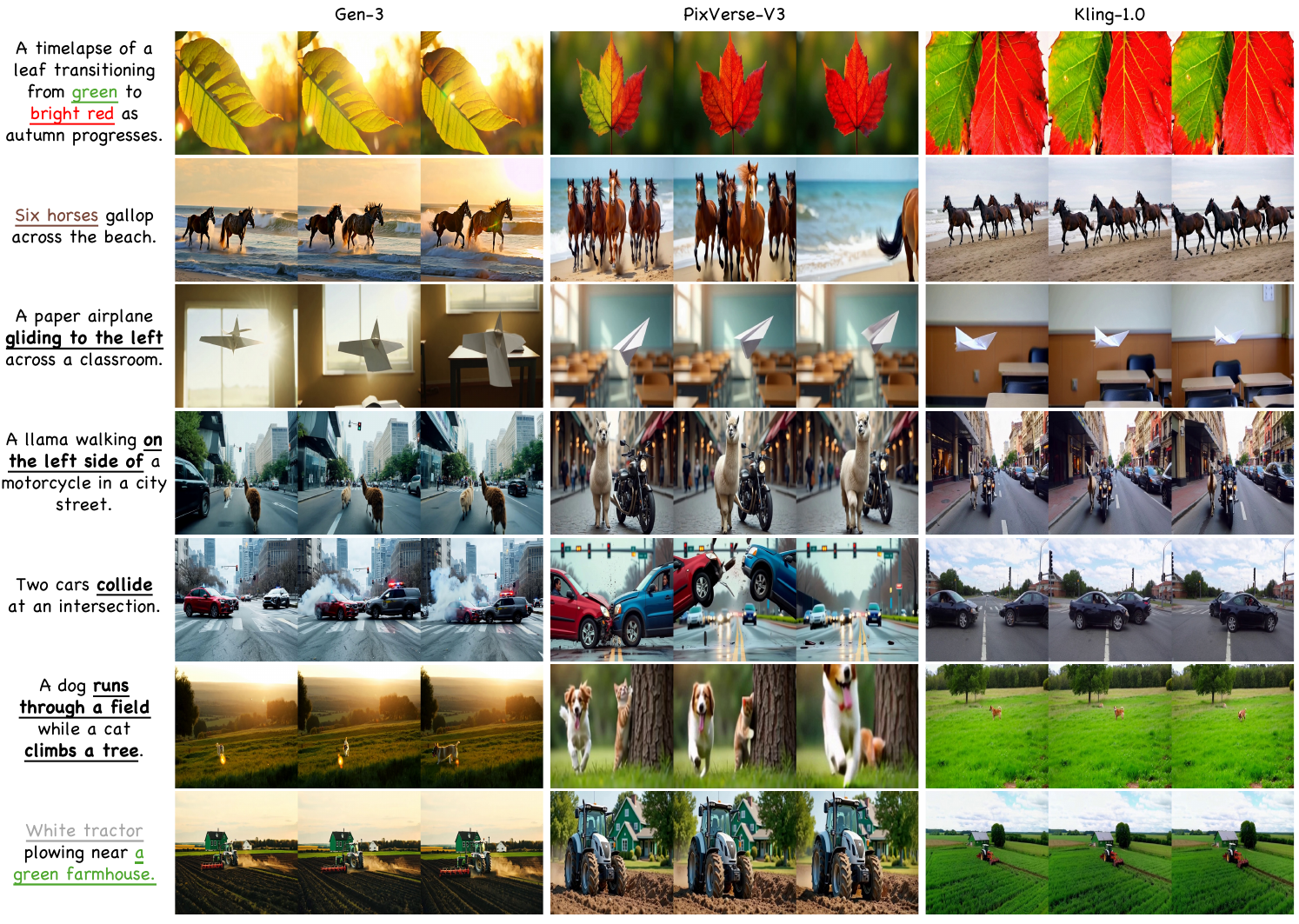}
    \vspace{-7pt}
    \caption{\textbf{Qualitative comparison with different commercial T2V models for the seven compositional categories in T2V-CompBench.} We show text-to-video examples generated by Gen-3~\cite{gen3}, PixVerse-V3~\cite{PixVerse}, and Kling-1.0~\cite{kling}.
    }
    \vspace{-40pt}
    \label{fig:qualitative_2}
\end{figure*}

The challenging cases for the seven compositional categories are illustrated in Figure~\ref{fig:qualitative_1} and~\ref{fig:qualitative_2}, with the difficulty level decreasing from top to bottom rows. Figure~\ref{fig:qualitative_1} shows example videos from open-source models, while Figure~\ref{fig:qualitative_2} shows those from commercial models. In our evaluation of various categories, we identify the following insights:

\noindent \textbf{\#1: Dynamic Attribute Binding is the most challenging category.}
As shown in row \#1 of Figure~\ref{fig:qualitative_1} and~\ref{fig:qualitative_2}, the evaluated T2V models tend to focus on certain keywords in the prompts, but overlook the required transitions in attributes. Consequently, they usually generate fixed objects or attributes that do not have any change through the video.

\noindent \textbf{\#2: T2V models struggle with generating correct spatial relationships, motion directions, and quantities.}
The second most challenging categories include spatial relationships, motion binding, and numeracy (rows \#4, \#3, and \#2 in Figure~\ref{fig:qualitative_1} and~\ref{fig:qualitative_2}). In spatial relationships, most T2V models struggle to differentiate between locality terms such as ``left'' and ``right'', resulting in random spatial layouts in the generated videos. 
The issue is even more pronounced in motion binding, where models can hardly ever understand moving directions, such as ``sail to the left'' or ``flying right towards''. Most T2V models not only fail to move in the correct directions but also have difficulties generating significant movement for objects.
In addition, generating videos with correct number of objects requires accurate counting. While the T2V models perform well when the quantities are fewer than three, they often fail to accurately generate larger quantities of objects. 

\noindent \textbf{\#3: While Object Interactions, Action Binding, and Consistent Attribute Binding are generally easier to handle, T2V models still encounter challenging scenarios.}
Text-to-video examples from object interactions, action binding, and consistent attribute binding are shown in rows \#5-7 in Figure~\ref{fig:qualitative_1} and Figure~\ref{fig:qualitative_2}.
In object interactions, some T2V models tend to produce static videos that do not depict the full interaction process. Regarding action binding, models may find it difficult to generate correct actions. For instance, given the prompt ``A dog runs through a field while a cat climbs a tree'', the models might incorrectly show both animals running instead of representing their respective actions, or they may only depict one animal while ignoring the other. In terms of consistent attribute binding, models sometimes fail to accurately associate attributes with the correct objects or overlook certain object entirely.

For additional analysis of the evaluation results, please refer to Appendix~\ref{app:result}.

\section{Conclusion}
\label{sec:conclusion}
We conduct the first systematic study on compositionality in text-to-video generation. We propose T2V-CompBench, a comprehensive benchmark for compositional text-to-video generation, with 1400 prompts in seven categories. We further design a suite of evaluation metrics for the seven categories, all of them are validated by correlations with human evaluation. Finally, we benchmark various text-to-video generation models with different architectures and provide insightful analysis and findings based on the results.
Compositional text-to-video generation is highly challenging for current models, and we hope our work will inspire future works to improve the compositionality of text-to-video models. For limitations and social impacts of our work, please refer to Appendix~\ref{app:limitations} and~\ref{app:societal_impacts}.
\newpage
{
    \small
    \bibliographystyle{ieeenat_fullname}
    \bibliography{main}
}

\clearpage
\maketitlesupplementary
\appendix

\section*{Appendix Contents}
\startcontents
\printcontents{ }{1}{}

\section{Prompt Suite}
\label{app:prompt_suite}
\subsection{Vocabulary Construction}
\label{app:word_selection}
We utilize WordNet~\cite{miller1995wordnet} to group words from real-user prompts provided by VidProM~\cite{wang2024vidprom} into multi-level meta classes. From these groups, we select high-frequency object nouns, active verbs, and adjectives.\\

\begin{figure}[htbp]
    \centering
    \vspace{-5pt}
    \includegraphics[width=\linewidth]{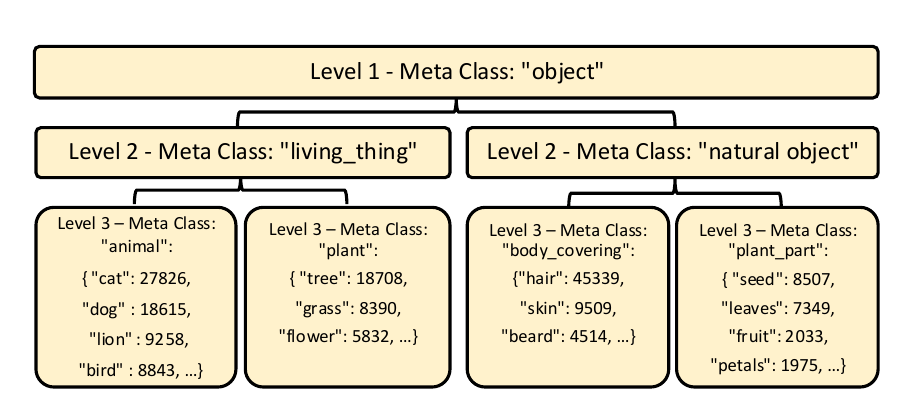}
    \vspace{-10pt}
    \caption{Illustration of multi-level meta class structure.}
    \vspace{-5pt}
    \label{fig:meta_class_level}
\end{figure}

\begin{figure}[htbp]
    \begin{minipage}[t]{0.5\linewidth}
        \centering
        \includegraphics[width=\textwidth]{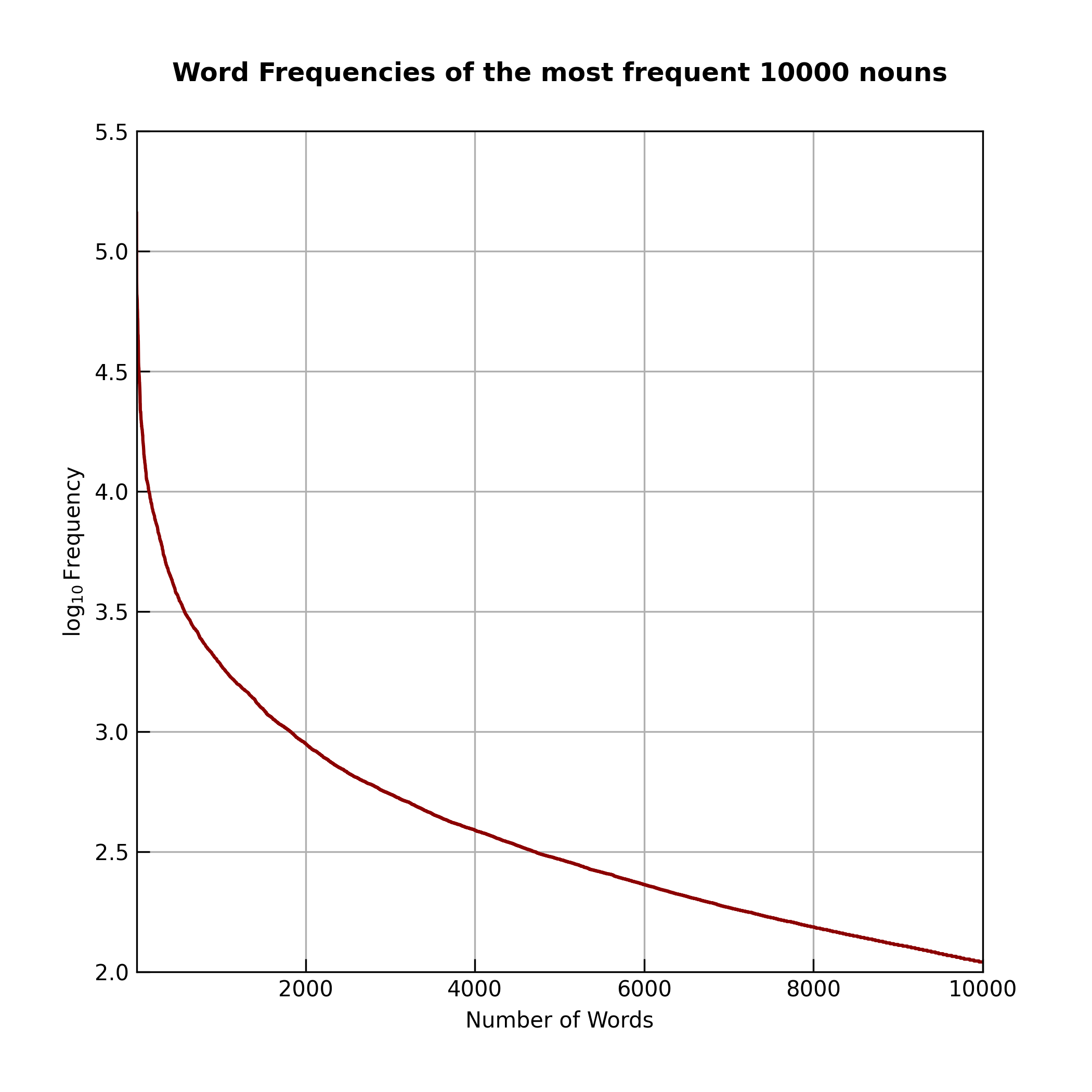}
        {\small
        \centerline{(a) noun frequency distribution}
        }
        \label{fig:vidprom_frequency_noun}
    \end{minipage}%
    \begin{minipage}[t]{0.5\linewidth}
        \centering
        \includegraphics[width=\textwidth]{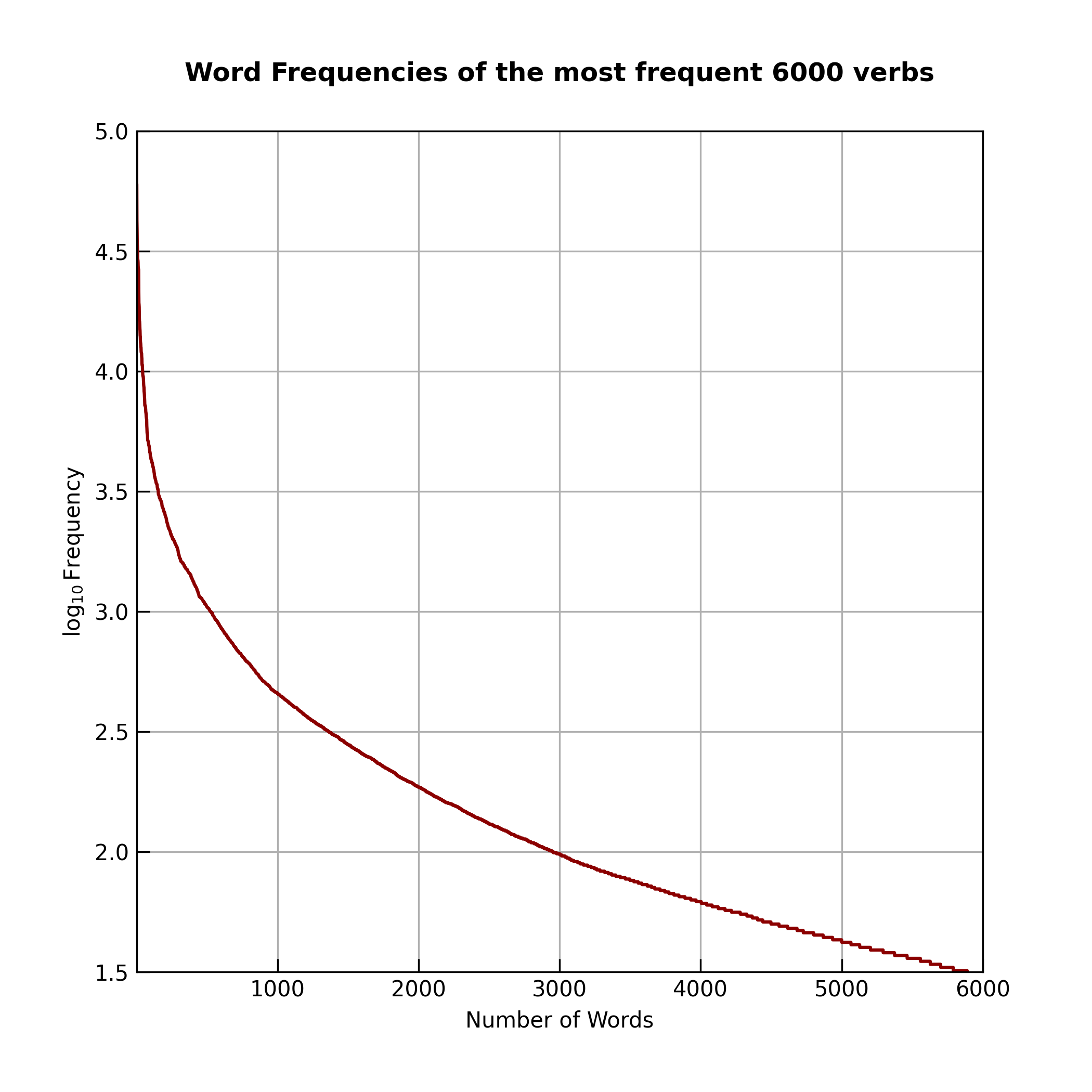}
        {\small
        \centerline{(b) verb frequency distribution}
        }
        \label{fig:vidprom_frequency_verb}
    \end{minipage}
    \vspace{-10pt}
    \caption{\textbf{Analysis of word frequencies.} We show the frequency of occurrence for nouns and verbs in real-user prompts.}
\label{fig:vidprom_frequency}
\end{figure}

\noindent \textbf{Nouns}. 
Figure~\ref{fig:meta_class_level} illustrates the structure of these multi-level classes. The nouns organized into multi-level meta classes come along with their frequencies of occurrence in real-user prompts. 
To select high-frequency words, we first analyze the frequency distribution of the entire real-user prompt dataset. Figure~\hyperref[fig:vidprom_frequency_noun]{\ref{fig:vidprom_frequency}(a)} displays the frequencies of the 10000 most frequent nouns, arranged in descending order. It is apparent that users do not use words with uniform frequency; rather, a significant number of words center around the first 2000 most frequent words, while the remaining words receive much less attention. 
In this dataset, all of the top 2000 words have frequencies greater than 900, so we primarily focus on words that exceed this threshold.
Following our word selection criteria outlined in Section~\ref{sec:new_dataset}, we start with words that meet our criteria, such as ``dog'' and ``car''. From there, we select additional words within the same class as ``dog'' or ``car'' to expand our noun list, for example, ``cat'', ``lion'' and ``truck'', ``boat''.
As a result, we identify a total of 260 object nouns, within which 66\% have frequencies greater than 900. Although the remaining words fall below this frequency, they are still commonly used in natural language and contribute to the diversity of our noun selection.

\noindent \textbf{Verbs}.
We apply the same method to group the verbs. Figure~\hyperref[fig:vidprom_frequency_noun]{\ref{fig:vidprom_frequency}(b)} illustrates the frequency distribution of verbs. The top 1200 words have frequencies over 370. Then, we select 200 active and vibrant verbs, with 60\% having frequencies greater than 370.

\noindent \textbf{Attributes}.
For words that describe attributes, we select from both adjectives and nouns.
When analyzing adjectives, we identify relevant words by looking for specific keywords in their definitions. If a definition includes terms like ``color'' or ``colored'', we classify the word as a color attribute. Similarly, if the definition contains ``made of'' or ``texture'', we categorize it under texture attributes. Word definitions including ``shape'' or ``shaped'' are classified as shape attributes. However, some words that should belong to these categories may be excluded if their definitions do not contain the relevant keywords.
To address this, we also consider attributes derived from nouns. Words that represent color or shape can function as either nouns or adjectives in a sentence, so we refer to the meta classes of ``color'' and ``shape'' within the noun classification. Additionally, there is a meta class for ``material'', which encompasses words that describe the texture of objects. We draw from these classes to supplement the attributes identified from adjectives. In total, we select 14 color attributes, 26 shape attributes, and 26 texture attributes.
For human-related attributes, we focus on nouns in meta classes such as ``body\_covering'', which includes words like ``hair'' and ``beard'', as well as ``clothing'', which features words like ``dress'' and ``hat''.

The list of the 260 nouns, 200 verbs and 100 attributes are provided in \href{https://github.com/KaiyueSun98/T2V-CompBench/blob/V2/word_list.py}{T2V-CompBench word list}.

\subsection{Prompt Generation with LLM}
\label{app:promptgen_llm}
When generating prompts with LLM, we provide the high-frequency nouns, verbs, attributes or other word lists to LLM. The templates used to generate prompts are displayed in section~\ref{app:template} Table~\ref{tab:template_prompt_1}, ~\ref{tab:template_prompt_2}, ~\ref{tab:template_prompt_3}, ~\ref{tab:template_prompt_4}, ~\ref{tab:template_prompt_5}, ~\ref{tab:template_prompt_6} and ~\ref{tab:template_prompt_7}. 

\subsection{Prompt Suite Statistics}
\label{t2v-compbench_metaclass}
\begin{figure}[htbp]
    \begin{minipage}[t]{0.5\linewidth}
        \centering
        \includegraphics[width=\textwidth]{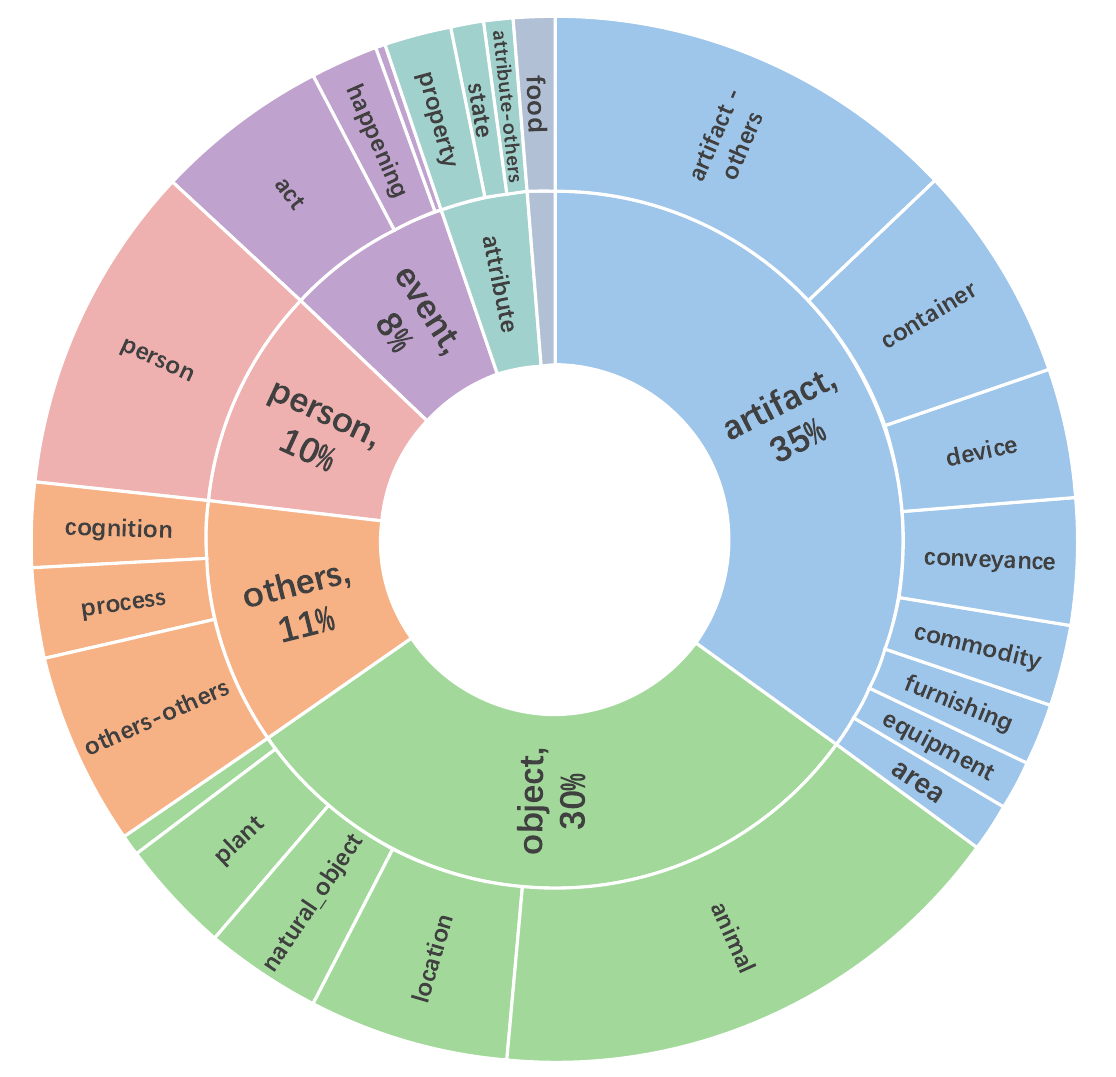}
        \centerline{(a) Meta types of nouns}
    \end{minipage}%
    \begin{minipage}[t]{0.5\linewidth}
        \centering
        \includegraphics[width=\textwidth]{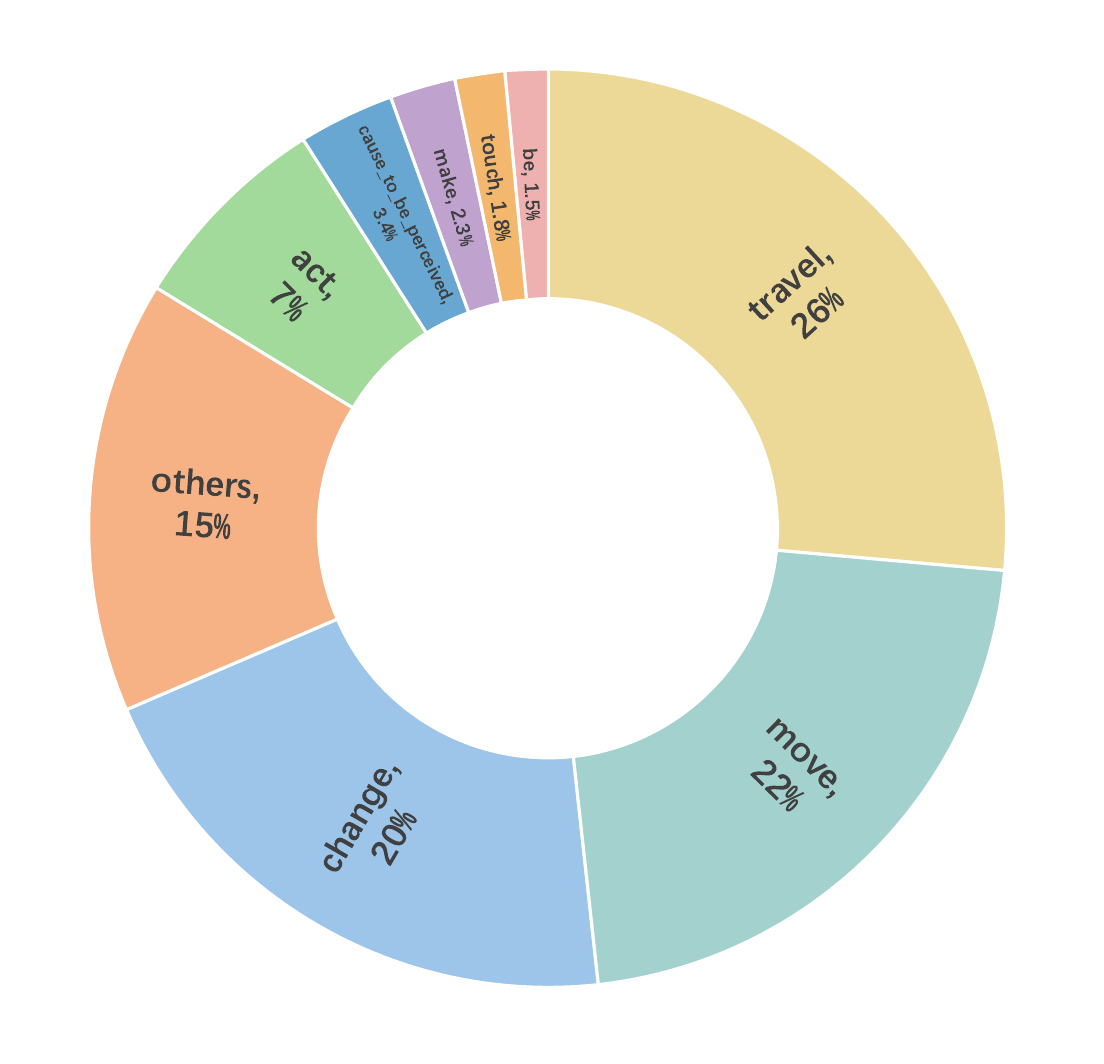}
        \centerline{(b) Meta types of verbs}
    \end{minipage}
    \caption{\textbf{Word distributions of T2V-CompBench prompts} We show the types of nouns and verbs of T2V-CompBench prompts.}
\label{fig:prompt_metaclass_dist}
\end{figure}

After obtaining the 1400 prompts of T2V-CompBench, we use WordNet~\cite{miller1995wordnet} to identify the metaclasses of nouns and verbs, and their distributions are visualized in Figure~\ref{fig:prompt_metaclass_dist}. 
In our prompt dataset, the high-frequency nouns are primarily concentrated in the categories of artifact, object, and person, most of which belong to the ``thing'' categories. Meanwhile, high-frequency verbs are mainly concentrated in actions like ``travel'', ``move'', and    ``change''. The occurrence of less dynamic verbs, such as those associated with ``express'' and ``think'', is nearly negligible in our dataset.

\subsection{Stability of 1400 Prompts and Videos}
\label{app:1400}
\begin{figure} [h!]
   \centering
   \vspace{-5pt}
   \includegraphics[width=0.8\linewidth]{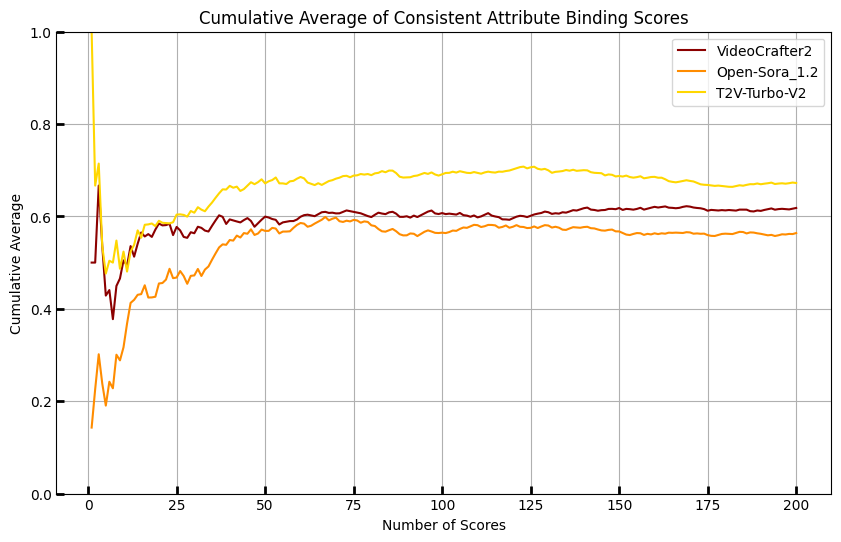}
   \vspace{0pt}
   \caption{The stability of \textit{Grid-LLaVA} with increasing number of videos in Consistent Attribute Binding.}
   \vspace{-10pt}
   \label{fig:plot_1}
\end{figure}

\begin{figure} [h!]
   \centering
   \vspace{-5pt}
   \includegraphics[width=0.8\linewidth]{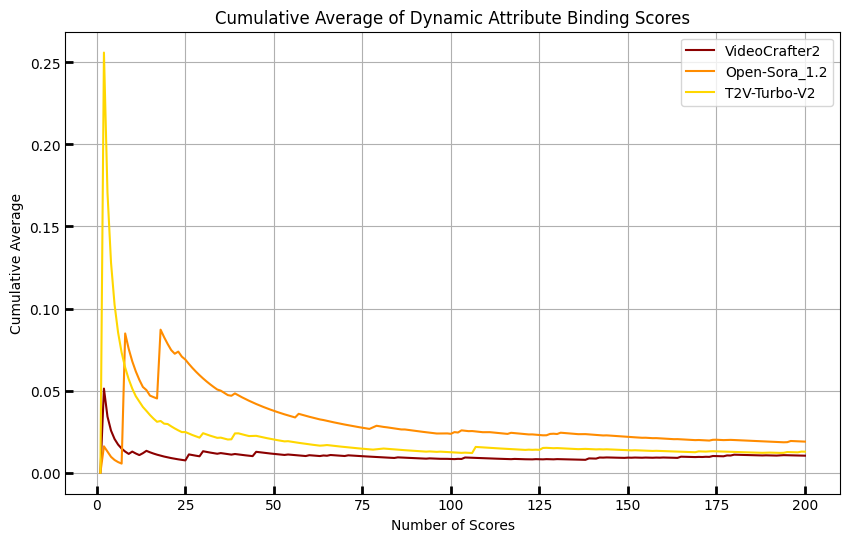}
   \vspace{0pt}
   \caption{The stability of \textit{D-LLaVA} with increasing number of videos in Dynamic Attribute Binding.}
   \vspace{-10pt}
   \label{fig:plot_2}
\end{figure}

\begin{figure} [h!]
   \centering
   \vspace{-5pt}
   \includegraphics[width=0.8\linewidth]{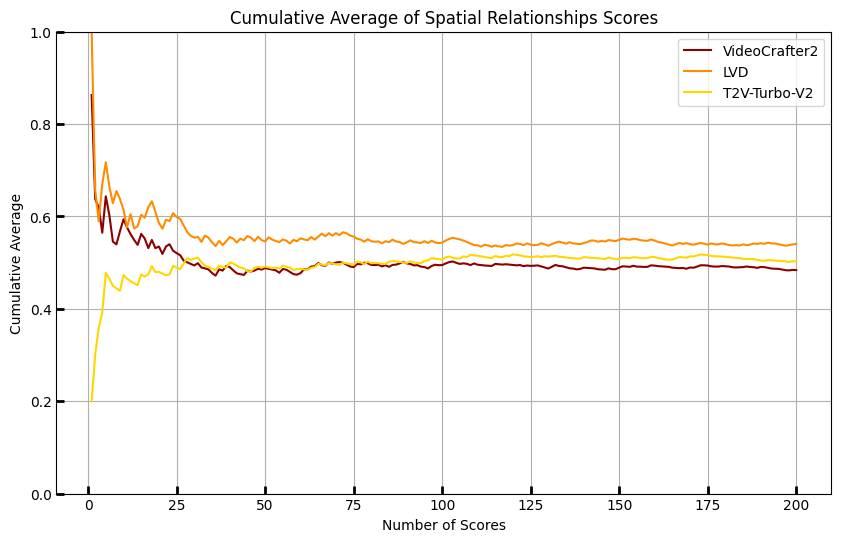}
   \vspace{0pt}
   \caption{The stability of \textit{G-Dino} with increasing number of videos in Spatial Relationships.}
   \vspace{-10pt}
   \label{fig:plot_3}
\end{figure}

\begin{figure} [h!]
   \centering
   \vspace{-5pt}
   \includegraphics[width=0.8\linewidth]{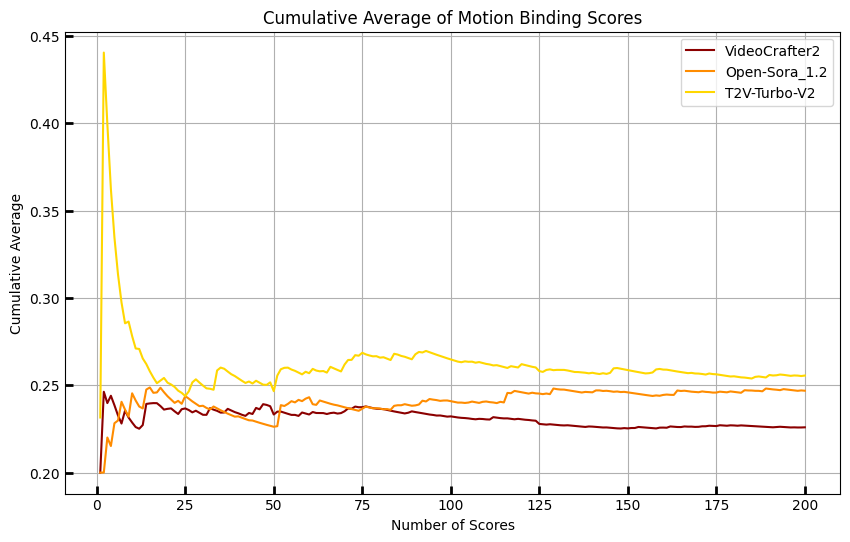}
   \vspace{0pt}
   \caption{The stability of \textit{DOT} with increasing number of videos in Motion Binding.}
   \vspace{-10pt}
   \label{fig:plot_4}
\end{figure}

\begin{figure} [h!]
   \centering
   \vspace{-5pt}
   \includegraphics[width=0.8\linewidth]{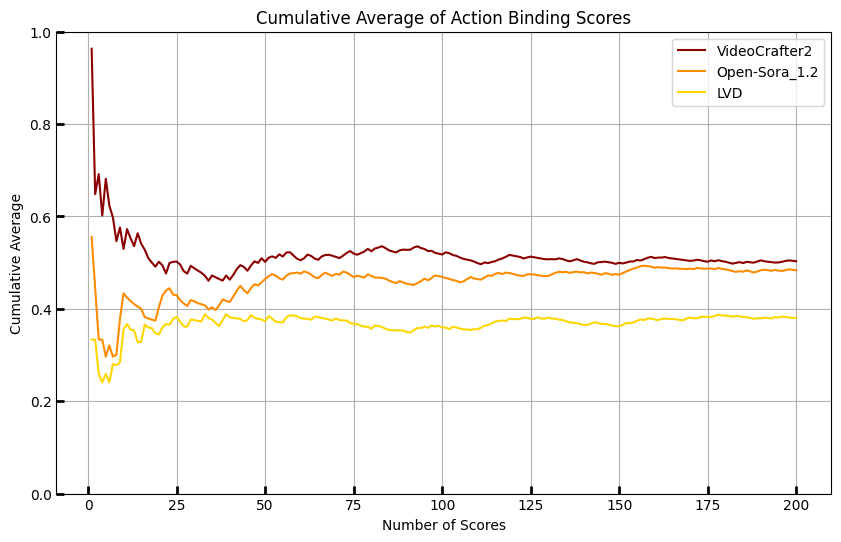}
   \vspace{0pt}
  \caption{The stability of \textit{Grid-LLaVA} with increasing number of videos in Action Binding.}
   \vspace{-10pt}
   \label{fig:plot_5}
\end{figure}

\begin{figure} [h!]
   \centering
   \vspace{-5pt}
   \includegraphics[width=0.8\linewidth]{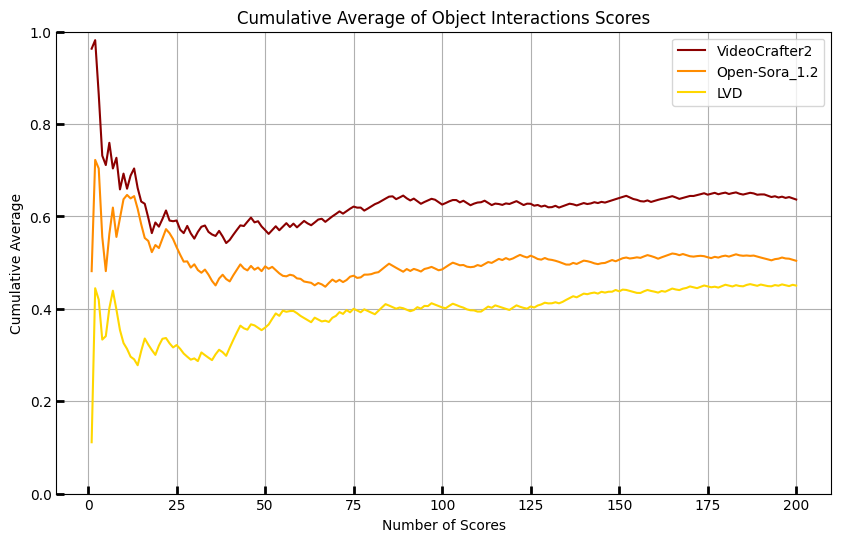}
   \vspace{0pt}
   \caption{The stability of \textit{Grid-LLaVA} with increasing number of videos in Object Interactions.}
   \vspace{-10pt}
   \label{fig:plot_6}
\end{figure}

\begin{figure} [h!]
   \centering
   \vspace{-5pt}
   \includegraphics[width=0.8\linewidth]{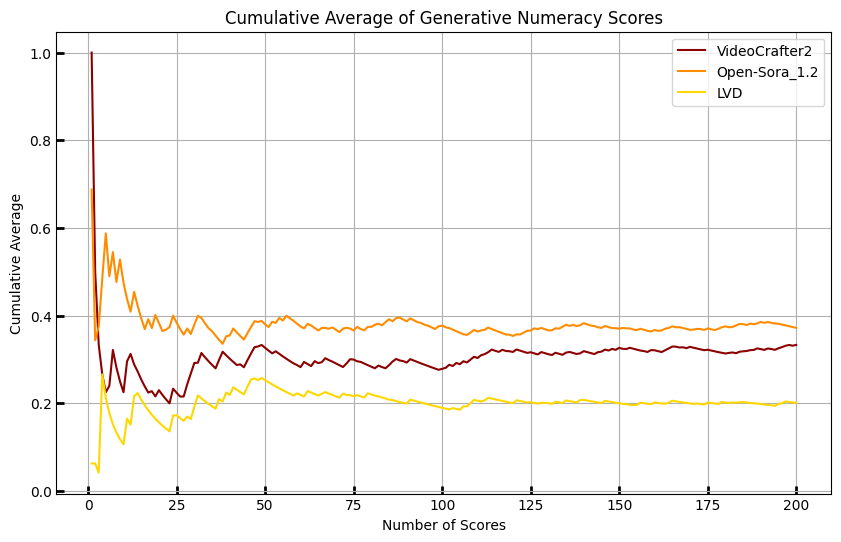}
   \vspace{0pt}
   \caption{The stability of \textit{G-Dino} with increasing number of videos in Generative Numeracy.}
   \vspace{0pt}
   \label{fig:plot_7}
\end{figure}

We conduct an analysis on the stability of our proposed metrics in evaluating the seven categories.
Figures~\ref{fig:plot_1}, ~\ref{fig:plot_2}, ~\ref{fig:plot_3}, ~\ref{fig:plot_4}, ~\ref{fig:plot_5}, ~\ref{fig:plot_6}, and ~\ref{fig:plot_7} illustrate how the average score changes as the number of videos increases within these categories. Here, we generate one video for each prompt. The analysis shows that the average score tends to stabilize as the number of videos increases. Some metrics, such as G-Dino for spatial relationships and numeracy, stabilize earlier, at approximately 125 videos, while others, such as Grid-LLaVA for consistent attribute binding, action binding and object interactions, stabilize around 150 videos. 
The stability of these metrics can also be influenced by different T2V models, as evidenced by the varying levels of fluctuations at the ends of the lines in the figures.
However, given that current T2V models typically take minutes to inference a video, it is more practical to limit the number of videos per category to 200, resulting in a total of 1400 prompts and videos. This ensures that the evaluation does not consume excessive computational time and resources.

\section{Implementation Details}
\label{app:implementation}
We follow the official and default implementations of the T2V models in evaluation. Details of the videos generated by the T2V models, including resolution, total frames, FPS, and duration are presented in Table~\ref{tab:fps}.

\begin{table}[ht]
\vspace{0pt}
\centering
\resizebox{\linewidth}{!}{%
\begin{tabular}{llcccc}
\toprule
\multicolumn{2}{c}{{\textbf{Model}}} & \textbf{Resolution} & \textbf{Frames}& \textbf{FPS} & \textbf{Duration (s)} \\
\midrule
& ModelScope~\cite{wang2023modelscope}                                       &   256$\times$256   &  16  &   8   &    2.0\\  
& ZeroScope~\cite{Zeroscope}                                                 &   576$\times$320   &  36  &   10  &    3.6\\  
& LVD~\cite{lian2023llm}                                                     &   256$\times$256   &  16  &   8   &    2.0\\  
\cmidrule{2-6}
& AnimateDiff~\cite{guo2023animatediff}                                      &   384$\times$256   &  16  &   8   &    2.0\\  
& MagicTime~\cite{yuan2024magictime}                                         &   512$\times$512   &  16  &   8   &    2.0\\ 
\cmidrule{2-6}
& Show-1~\cite{zhang2023show}                                                &   576$\times$320   &  29  &   8   &    3.6\\  
& VideoCrafter2~\cite{chen2024videocrafter2}                                 &   512$\times$320   &  16  &   8   &    2.0\\  
& VideoTetris~\cite{tian2024videotetris}                                              &   512$\times$320   &  16  &   8   &    2.0\\  
& Vico~\cite{yang2024vico}                                                   &   512$\times$320   &  16  &   8   &    2.0\\  
& T2V-Turbo-V2~\cite{li2024t2vturbo}                                         &   512$\times$320   &  16  &   8   &    2.0\\  
\midrule
& Latte~\cite{ma2024latte}                                                   &   512$\times$512   &  16  &   8   &    2.0\\  
& Open-Sora 1.1~\cite{opensora}                                              &   512$\times$512   &  16  &   8   &    2.0\\  
& Open-Sora 1.2~\cite{opensora}                                              &   640$\times$360   &  51  &   24  &    2.1\\  
& Open-Sora-Plan v1.0.0~\cite{pku_yuan_lab_and_tuzhan_ai_etc_2024_10948109}  &   512$\times$512   &  65  &   24  &    2.7\\  
& Open-Sora-Plan v1.3.0~\cite{pku_yuan_lab_and_tuzhan_ai_etc_2024_10948109}  &   640$\times$352   &  93  &   18  &    5.2\\  
& CogVideoX-5B~\cite{yang2024cogvideox}                                      &   512$\times$320   &  40  &   8   &    5.0\\  
& Mochi~\cite{genmo2024mochi}                                                &   512$\times$320   &  151  &  30  &    5.0\\  
\midrule
\midrule
& Pika-1.0~\cite{pika}                                                       &   1280$\times$720  &  72  &   24  &    3.0\\  
& Gen-2~\cite{gen2}                                                          &   1408$\times$768  &  96  &   24  &    4.0\\  
& Gen-3~\cite{gen3}                                                          &   1280$\times$768  &  128 &   24  &    5.3\\  
& Dreamina 1.2~\cite{Dreamina}                                                   &   1280$\times$720  &  24  &   8   &    3.0\\  
& PixVerse-V3~\cite{PixVerse}                                                &   1408$\times$768  &  161 &   30  &    5.4\\  
& Kling-1.0~\cite{kling}                                                     &   1280$\times$720  &  153 &   30  &    5.1\\ 
\bottomrule
\end{tabular}%
}
\caption{\textbf{Details of generated videos by T2V models}. The table shows resolution, total frames, FPS, and video duration in second for videos generated by T2V models.} 
\label{tab:fps}
\vspace{0pt}
\end{table}

\section{Using MLLMs as Evaluation Metrics}
\label{app:mllm}
\subsection{How to query MLLMs}
One challenge of using MLLMs as evaluation metrics is their tendency to generate hallucinations. This can manifest as mistakes in identifying visual content or selection of unmatched grades or scores.
\begin{figure*}[h]
    \centering
    \vspace{-5pt}
    \includegraphics[width=\linewidth]{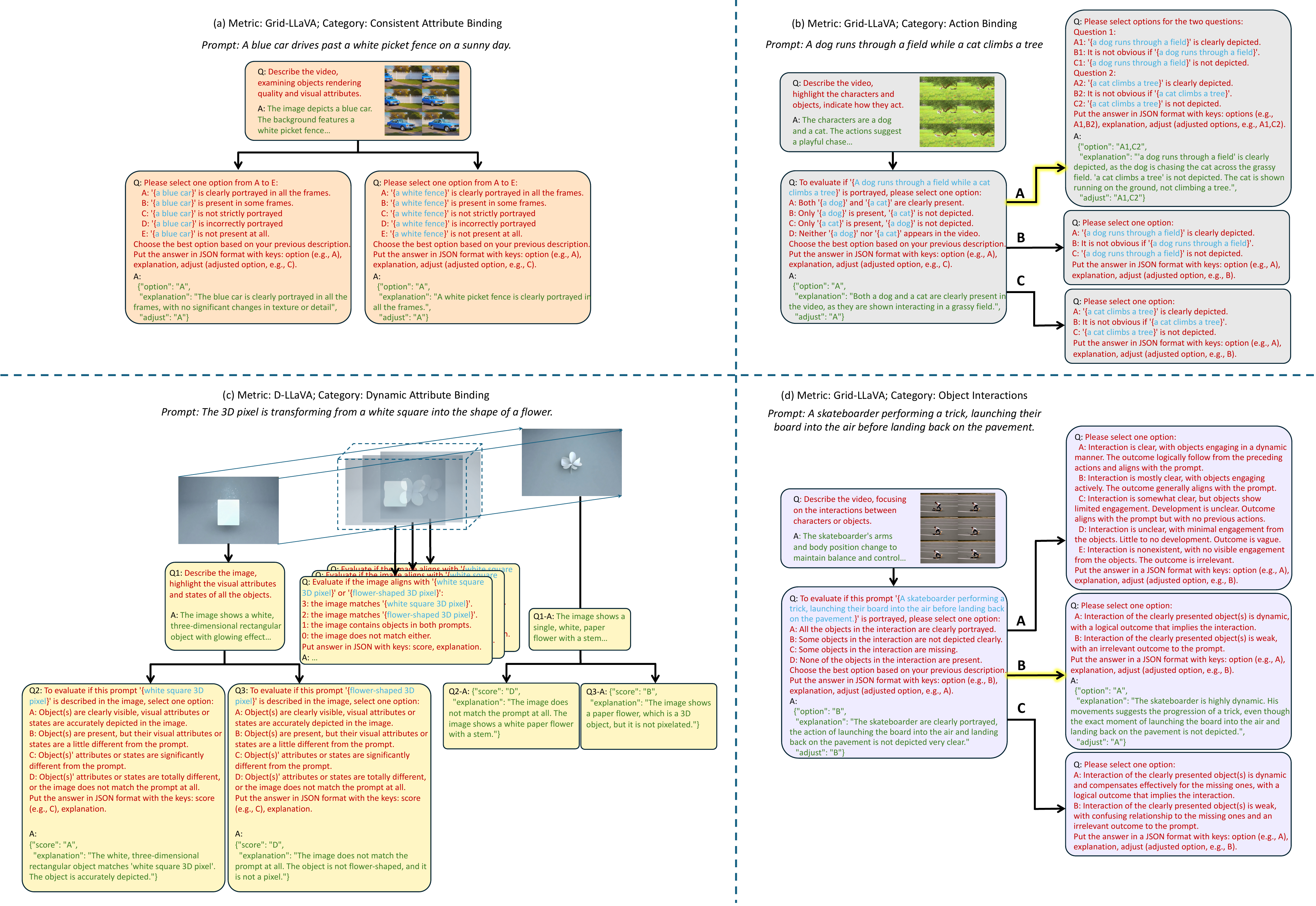}
    \vspace{-10pt}
    \caption{\textbf{Question and answer examples for MLLM-based evaluation metrics.} We replace the parts marked in \textcolor{cyan}{cyan} with metadata for different prompts. Answers are generated by LLaVA~\cite{liu2023improvedllava}. Note that the examples are simplified slightly for clarity and brevity.}
    \vspace{-5pt}
    \label{fig:llava}
\end{figure*}

To mitigate these hallucinations, we employ the chain-of-thought mechanism~\cite{wei2022chain}. We firstly ask the MLLM to describe the image without revealing the specific question we intend to ask. 
This allows the MLLM to independently describe the visual content, without being affected by the subsequent questions.

When querying the MLLM for evaluation, it is important to prepare sufficient grading options to differentiate performances of the models. However, too many options make it difficult for the MLLM to identify the subtle differences between them. To this end, we disentangle the evaluation aspects into parallel or sequential queries, asking about each aspect separately. This ensures a sufficient number of grades while preventing the MLLM from being overwhelmed with too many options at once. For consistent attribute binding, we ask whether each of the two objects possesses the correct attribute. For dynamic attribute binding, we ask whether the image depicts the initial or final state. For action binding, we first inquire about the presence of the two objects. Based on that response, we then ask if the actions of the present object(s) align with the prompt. For object interactions, we start by confirming the presence of objects, then evaluate the quality of their interaction. Figure~\ref{fig:llava} shows the question and answer examples for our proposed MLLM-based evaluation metrics.

\subsection{How to obtain reliable and reproducible results}
\begin{figure}[h]
    \centering
    \vspace{0pt}
    \includegraphics[width=\linewidth]{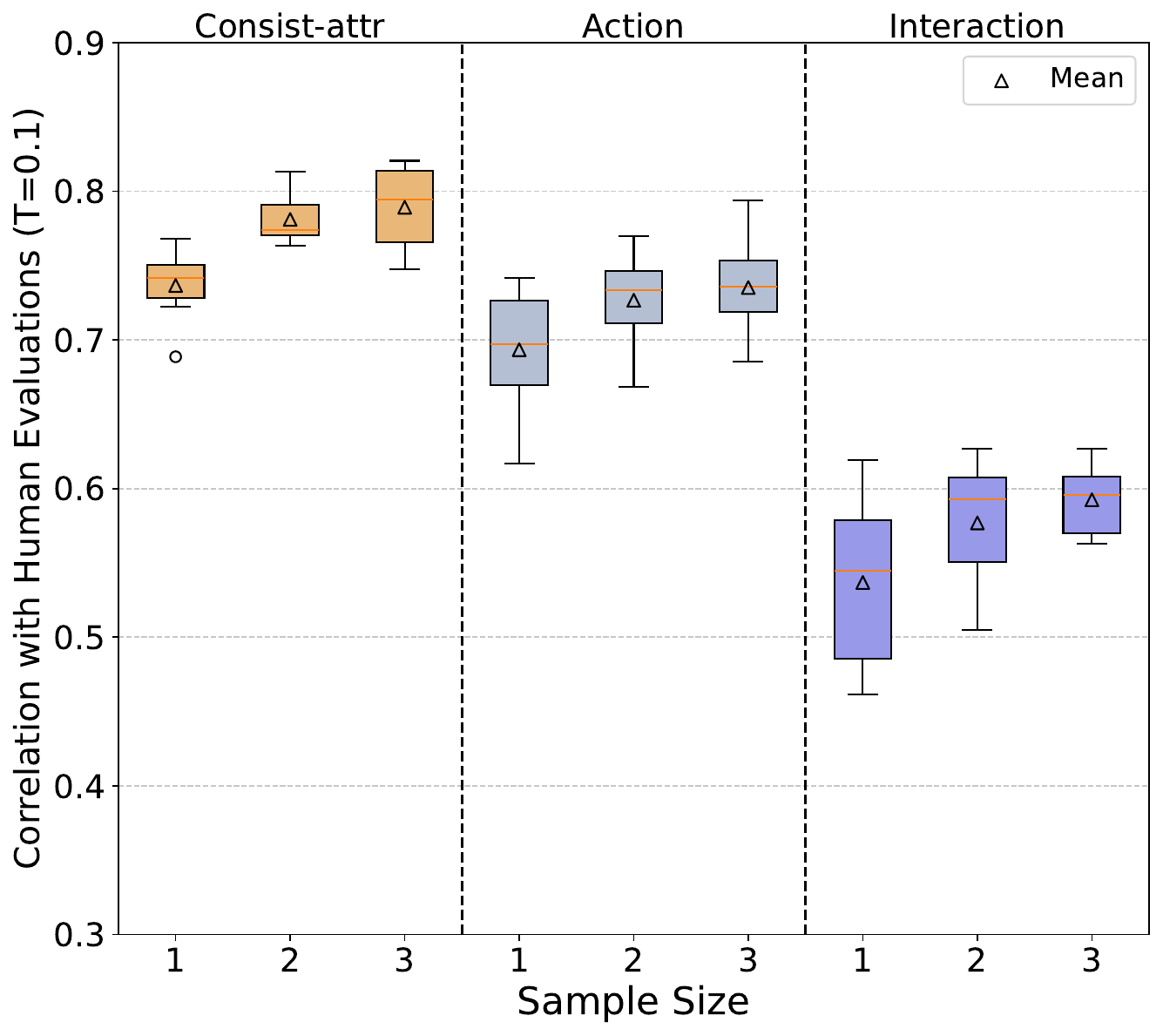}
    \vspace{0pt}
    \caption{\textbf{Box Plot Analysis of Correlation Between Human Evaluations and Grid-LLaVA.} We show the box plot of correlations (Spearman's $\rho$) between human evaluations and Grid-LLaVA in consistent attribute binding, action binding and object interactions at temperature 0.1 with sample size of 1, 2, and 3.}
    \vspace{0pt}
    \label{fig:box_avg}
\end{figure}

It is essential for evaluation metrics to produce reliable and reproducible results. To ensure that the results generated by an MLLM are reproducible, we fix its parameters and seed each time it evaluates a video. 
However, we empirically find that varying the seed during evaluations can change the ranking of the models, especially when their scores are close. 
To obtain a more reliable score for each video and ranking for the models, an effective approach is to query the MLLM multiple times and average the resulting scores. This method is similar to asking multiple annotators to rate each video, as they may have different opinions.
To decide how many times to query the MLLM, we run Grid-LLaVA 8 times with the same settings, varying only the seed. This is done on videos evaluated by humans for the categories of consistent attribute binding, action binding, and object interactions.
We randomly sample 2 or 3 experiments from the 8 conducted and calculate the average score for each video. We repeat this process 8 times and obtain 8 correlations with human scores for a sample size of 2 and 8 correlations for a sample size of 3. The results are summarized in Figure~\ref{fig:box_avg}.
As shown in the box plot, increasing the sample size leads to higher mean and median and generally more stable correlations, which in turn indicates that the scores are more stable. 
Therefore, considering both reliability and practicality, we decide to sample 3 scores for each video, and use their average as the final result.

\section{Evaluation Results Analysis}
\label{app:result}
As mentioned in Section~\ref{sec:new_dataset}, we divide each category in T2V-CompBench into several subgroups. In addition, for certain categories, the final score is determined through multiple stages. Here, we extract the scores for some important subgroups, and also extrapolate some meaningful data from the interim stages. The results for these sub-dimensions are documented in Table~\ref{tab:scores}. 
In this section, we provide comprehensive explanations for these sub-dimensions and analysis for their results, along with visualizations for each category.

\subsection{Consistent Attribute Binding}
\begin{figure}[htbp]
    \centering
    \vspace{-5pt}
    {\scriptsize Prompt: \textit{A blue car drives past a white picket fence on a sunny day.}}
    \vspace{0pt} 
    \begin{subfigure}{\linewidth}
        \centering
        \includegraphics[width=\linewidth]{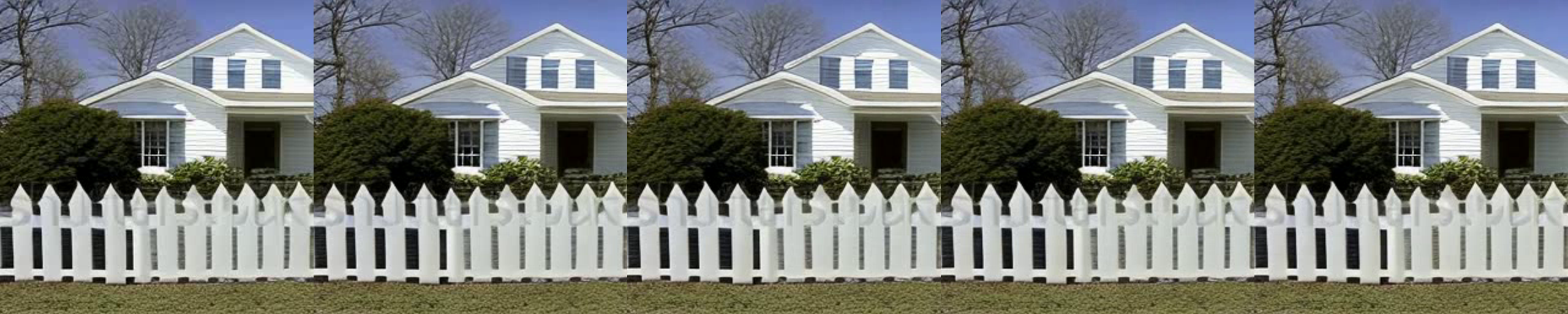} 
    \caption{T2V Model: ModelScope~\cite{wang2023modelscope}. Score: 0.5000}
    \end{subfigure}
    \begin{subfigure}{\linewidth}
        \centering
        \includegraphics[width=\linewidth]{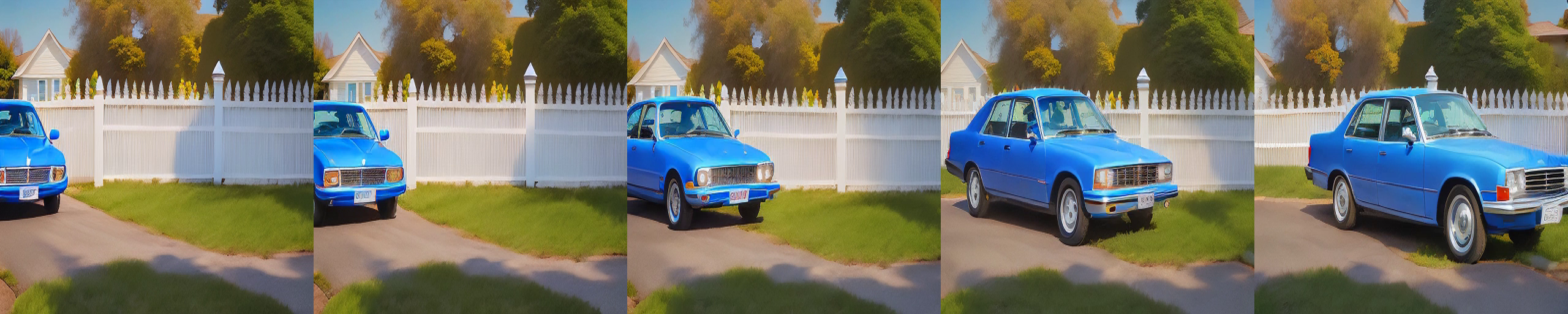} 
        \caption{T2V Model: T2V-Turbo-V2~\cite{li2024t2vturbo}. Score: 0.9524 }
    \end{subfigure}
    \vspace{-15pt}
    \caption{Visualization of Consistent Attribute Binding examples.}
    \label{fig:analysis_1_image}
    \vspace{-10pt}
\end{figure}

The first three columns of Table~\ref{tab:scores} present the average score for the subgroups of color, shape, and texture within the category of consistent attribute binding. Among the three subgroups, color is the easiest to manage, followed by texture, while shape proves to be the most challenging.
In Figure~\ref{fig:analysis_1_image}, we present a concatenation of frames from videos generated by ModelScope~\cite{wang2023modelscope} and T2V-Turbo-V2~\cite{li2024t2vturbo}. Not only does T2V-Turbo-V2~\cite{li2024t2vturbo} accurately represent the color-object binding in the prompt, but it also demonstrates noticeable object movement.

\subsection{Dynamic Attribute Binding}
\begin{figure}[h]
    \centering
    \vspace{-5pt}
    {\scriptsize Prompt: \textit{A yellow rubber duck in bath water turns a deep purple.}}
    \begin{subfigure}{\linewidth}
        \centering
        \includegraphics[width=\linewidth]{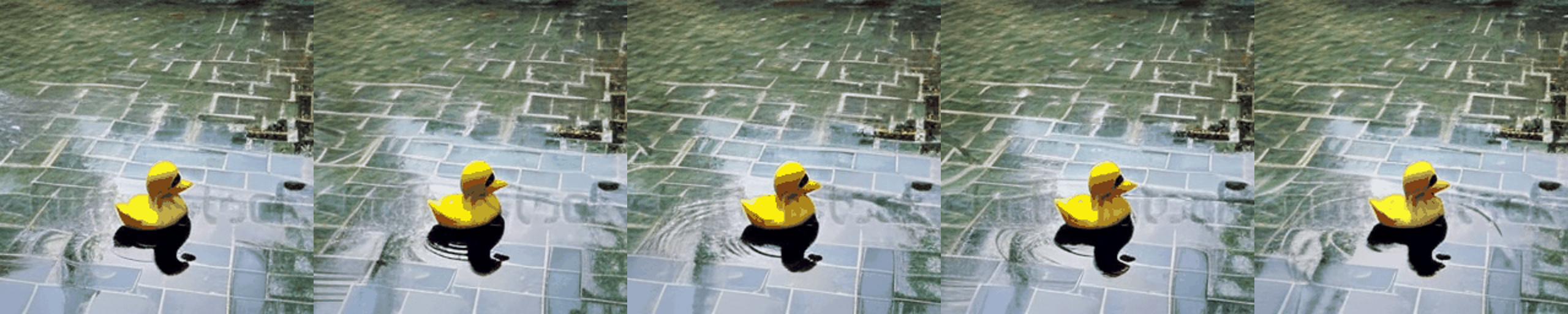} 
    \caption{T2V Model: LVD~\cite{lian2023llm}. Score: 0.0000}
    \end{subfigure}
    \begin{subfigure}{\linewidth}
        \centering
        \includegraphics[width=\linewidth]{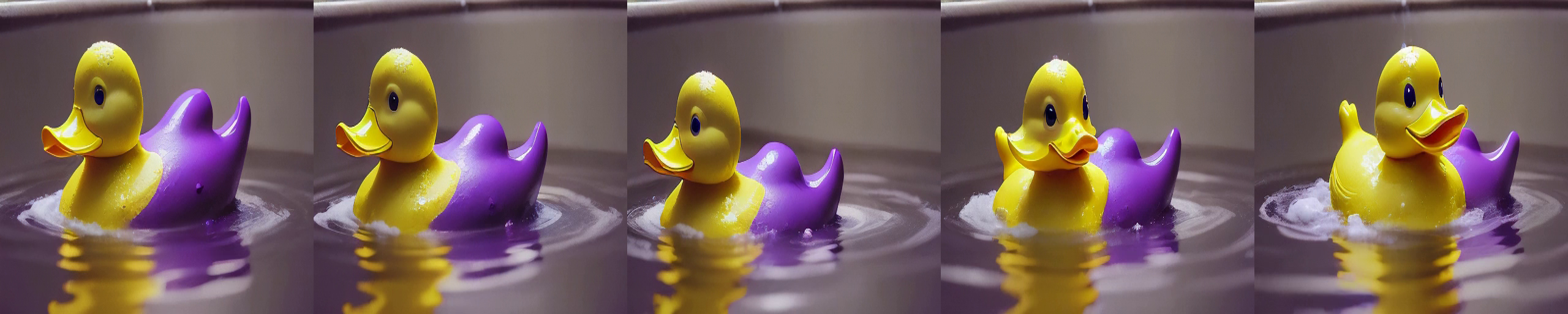}
        \caption{T2V Model: T2V-Turbo-V2~\cite{li2024t2vturbo}. Score: 0.0000}
    \end{subfigure}
    \begin{subfigure}{\linewidth}
        \centering
        \includegraphics[width=\linewidth]{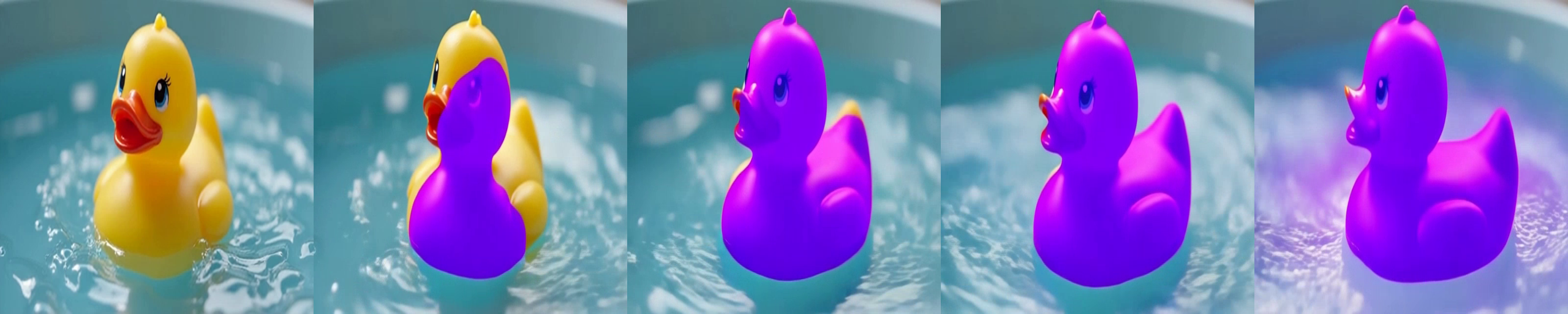} 
        \caption{T2V Model: PixVerse-V3~\cite{PixVerse}. Score: 0.8000}
    \end{subfigure}
    \vspace{-15pt}
    \caption{Visualization of Dynamic Attribute Binding examples.}
    \vspace{-10pt}
    \label{fig:analysis_2_image}
\end{figure}

For the category of dynamic attribute binding, T2V models rarely produce qualified videos. For example, PixVerse-V3~\cite{PixVerse} achieves the highest score of $0.0687$ in this category. Out of the 200 generated videos, only 31 show relevant elements or transitions, and just 9 of those exhibit meaningful transitions.
Models always struggle to generate changing attributes. LVD~\cite{lian2023llm} and T2V-Turbo-V2~\cite{li2024t2vturbo} exemplify this difficulty in Figure~\ref{fig:analysis_2_image}. They tend to produce static content that only reflects either the initial or final state of the objects described in the prompt, or both states simultaneously. In this case, PixVerse-V3~\cite{PixVerse} demonstrates better prompt understanding and temporal dynamics.

\begin{table*}[htbp]
\centering
\resizebox{\linewidth}{!}{%
\begin{tabular}
{llcccccccccccc}
\toprule 
\multicolumn{2}{c}{{\textbf{Model}}} & \multicolumn{3}{c}{\textbf{Consist-attr}} & \multicolumn{3}{c}{\textbf{Spatial-2D}} & \multicolumn{2}{c}{\textbf{Motion}} & \multicolumn{2}{c}{\textbf{Action}} & \multicolumn{2}{c}{\textbf{Interaction}}\\
\cmidrule(lr){1-2} \cmidrule(lr){3-5} \cmidrule(lr){6-8} \cmidrule(lr){9-10} \cmidrule(lr){11-12} \cmidrule(lr){13-14}
\multicolumn{2}{c}{Sub-dimension} & Color & Shape & Texture & Coexist & Acc. & Acc.Score & Motion Level & Acc. & Common & Uncommon & Physical & Social  \\
\midrule
\multicolumn{2}{l}{\hspace{-.5em} \textit{diffusion unet-based}} \\
 & ModelScope~\cite{wang2023modelscope}                                       & 0.5826 & 0.3440 & 0.4560 & 62\% & 44\% & 0.8011 & 16.98 & 32\% & 0.4025 & 0.2093 & 0.3815 & 0.5411   \\
 & ZeroScope~\cite{Zeroscope}                                                 & 0.4829 & 0.1905 & 0.3012 & 50\% & 65\% & 0.7836 & \cellcolor{mycolor_blue}{21.00} & 29\% & 0.4042 & 0.2139 & 0.2970 & 0.5422   \\
 & LVD~\cite{lian2023llm}                                                     & 0.6155 & 0.4139 & 0.4458 & 63\% & \cellcolor{mycolor_blue}{\textbf{90\%}} & \cellcolor{mycolor_blue}{\textbf{0.9054}} & 13.73 & \cellcolor{mycolor_blue}{\textbf{47\%}} & 0.4120 & 0.2528 & 0.3752 & 0.5252   \\
\cmidrule{2-14}
 & AnimateDiff~\cite{guo2023animatediff}                                      & 0.4619 & 0.3113 & 0.4196 & 50\% & 52\% & 0.8180 & 12.79 & 21\% & 0.3053 & 0.2009 & 0.3604 & 0.4337   \\
 \cmidrule{2-14}
 & Show-1~\cite{zhang2023show}                                                & 0.5964 & 0.4774 & 0.5601 & 62\% & 56\% & 0.8387 & 11.74 & 26\% & 0.4331 & 0.2083 & 0.5659 & 0.6830   \\
 & VideoCrafter2~\cite{chen2024videocrafter2}                                 & 0.6717 & 0.4792 & \cellcolor{mycolor_blue}{0.5780} & 73\% & 59\% & 0.7914 & 8.96  & 24\% & 0.5472 & 0.3259 & 0.5607 & \cellcolor{mycolor_blue}{0.7122}   \\
 & VideoTetris~\cite{tian2024videotetris}                                              & 0.6957 & 0.4667 & 0.5589 & 74\% & 53\% & 0.7831 & 9.94  & 21\% & 0.5410 & 0.3056 & \cellcolor{mycolor_blue}{0.6200} & 0.6956   \\
 & Vico~\cite{yang2024vico}                                                   & 0.6705 & 0.4476 & 0.4619 & \cellcolor{mycolor_blue}{80\%} & 50\% & 0.8083 & 9.12  & 20\% & 0.5609 & 0.3120 & 0.5263 & 0.6652   \\
 & T2V-Turbo-V2~\cite{li2024t2vturbo}                                         & \cellcolor{mycolor_blue}{0.7781} & \cellcolor{mycolor_blue}{0.5351} & 0.5452 & 77\% & 54\% & 0.7967 & 19.14 & 37\% & \cellcolor{mycolor_blue}{0.6694} & \cellcolor{mycolor_blue}{0.3657} & 0.5907 & 0.6970   \\
 \midrule
 \multicolumn{2}{l}{\hspace{-.5em} \textit{DiT-based}} \\
 & Latte~\cite{ma2024latte}                                                   & 0.4657 & 0.3613 & 0.5280 & 61\% & 51\% & 0.7856 & 12.76 & 14\% & 0.4623 & 0.2241 & 0.3904 & 0.4389   \\
 & Open-Sora 1.1~\cite{opensora}                                              & 0.5669 & 0.4220 & 0.5304 & 77\% & 69\% & 0.8323 & 10.87 & 25\% & 0.5613 & 0.2731 & 0.5070 & 0.6059   \\
 & Open-Sora 1.2~\cite{opensora}                                              & 0.6038 & 0.4321 & \cellcolor{mycolor_yellow}{0.5607} & 76\% & 50\% & \cellcolor{mycolor_yellow}{0.8905} & 15.98 & 30\% & 0.5069 & \cellcolor{mycolor_yellow}{0.3889} & 0.4396 & 0.5681   \\
 & Open-Sora-Plan v1.0.0~\cite{pku_yuan_lab_and_tuzhan_ai_etc_2024_10948109}  & 0.4357 & 0.3887 & 0.4107 & 66\% & 58\% & 0.7320 & 7.54  & 17\% & 0.4583 & 0.1713 & 0.4067 & 0.4233   \\
 & Open-Sora-Plan v1.3.0~\cite{pku_yuan_lab_and_tuzhan_ai_etc_2024_10948109}  & 0.7107 & 0.4244 & 0.5429 & 76\% & 59\% & 0.8726 & 15.89 & 22\% & 0.4938 & 0.2870 & 0.4393 & 0.4574   \\
 & CogVideoX-5B~\cite{yang2024cogvideox}                                      & \cellcolor{mycolor_yellow}{0.7202} & 0.4333 & 0.5167 & 71\% & 61\% & 0.8706 & \cellcolor{mycolor_yellow}{23.33} & \cellcolor{mycolor_yellow}{34\%} & \cellcolor{mycolor_yellow}{0.5824} & 0.3370 & \cellcolor{mycolor_yellow}{0.5607} & \cellcolor{mycolor_yellow}{0.6530}   \\
 & Mochi~\cite{genmo2024mochi}                                                & 0.7043 & \cellcolor{mycolor_yellow}{0.4393} & 0.4774 & \cellcolor{mycolor_yellow}{78\%} & \cellcolor{mycolor_yellow}{75\%} & 0.8045 & 18.55 & \cellcolor{mycolor_yellow}{34\%} & 0.5336 & 0.2454 & 0.4930 & 0.5833   \\
\midrule
\midrule
\multicolumn{2}{l}{\hspace{-.5em} \textit{commercial}} \\
 & Pika-1.0~\cite{pika}                                                       & 0.6095 & 0.4315 & 0.5119 & 72\% & 55\% & 0.7841 & 8.13  & 17\% & 0.4650 & 0.2648 & 0.4348 & 0.6048   \\
 & Gen-2~\cite{gen2}                                                          & 0.6721 & 0.3994 & 0.5131 & 83\% & 53\% & 0.8162 & 7.73  & 20\% & 0.4894 & 0.2491 & 0.5681 & 0.6607   \\
 & Gen-3~\cite{gen3}                                                          & 0.6652 & 0.5077 & 0.5280 & 68\% & 67\% & 0.8728 & 22.59 & \cellcolor{mycolor_red}{42\%} & 0.5639 & 0.3611 & 0.5663 & 0.6148   \\
 & Dreamina 1.2~\cite{Dreamina}                                                   & 0.7798 & 0.5601 & 0.5738 & \cellcolor{mycolor_red}{\textbf{90\%}} & 62\% & \cellcolor{mycolor_red}{0.8777} & 12.71 & 23\% & 0.6225 & 0.4722 & 0.6411 & 0.7237   \\
 & PixVerse-V3~\cite{PixVerse}                                                & 0.7898 & \cellcolor{mycolor_red}{\textbf{0.5768}} & \cellcolor{mycolor_red}{\textbf{0.6381}} & 89\% & \cellcolor{mycolor_red}{71\%} & 0.8231 &  \cellcolor{mycolor_red}{\textbf{28.09}} & 38\% & \cellcolor{mycolor_red}{\textbf{0.9016}} & \cellcolor{mycolor_red}{\textbf{0.7546}} & \cellcolor{mycolor_red}{\textbf{0.8219}} & \cellcolor{mycolor_red}{\textbf{0.8400}}   \\
 & Kling-1.0~\cite{kling}                                                     & \cellcolor{mycolor_red}{\textbf{0.7971}} & 0.4673 & 0.6149 & \cellcolor{mycolor_red}{\textbf{90\%}} & 57\% & 0.8634 & 17.27 & 35\% & 0.5863 & 0.5481 & 0.6530 & 0.7726   \\

\bottomrule
\end{tabular}%
}
\caption{\textbf{T2V-CompBench sub-dimension evaluation results using proposed metrics}. Scores are normalized between 0 and 1. A higher score indicates better performance.  \textbf{Bold} signifies the highest score within each category. \colorbox{mycolor_blue}{Blue} highlights the top score among diffusion unet-based models. \colorbox{mycolor_yellow}{Yellow} highlights the top score among DiT-based models. \colorbox{mycolor_red}{Red} highlights the top score among commercial models.
}
\label{tab:scores}
\end{table*}

\begin{figure*}[htbp]
\centering
\begin{minipage}{0.33\linewidth}
    \centering
    \includegraphics[width=\textwidth]{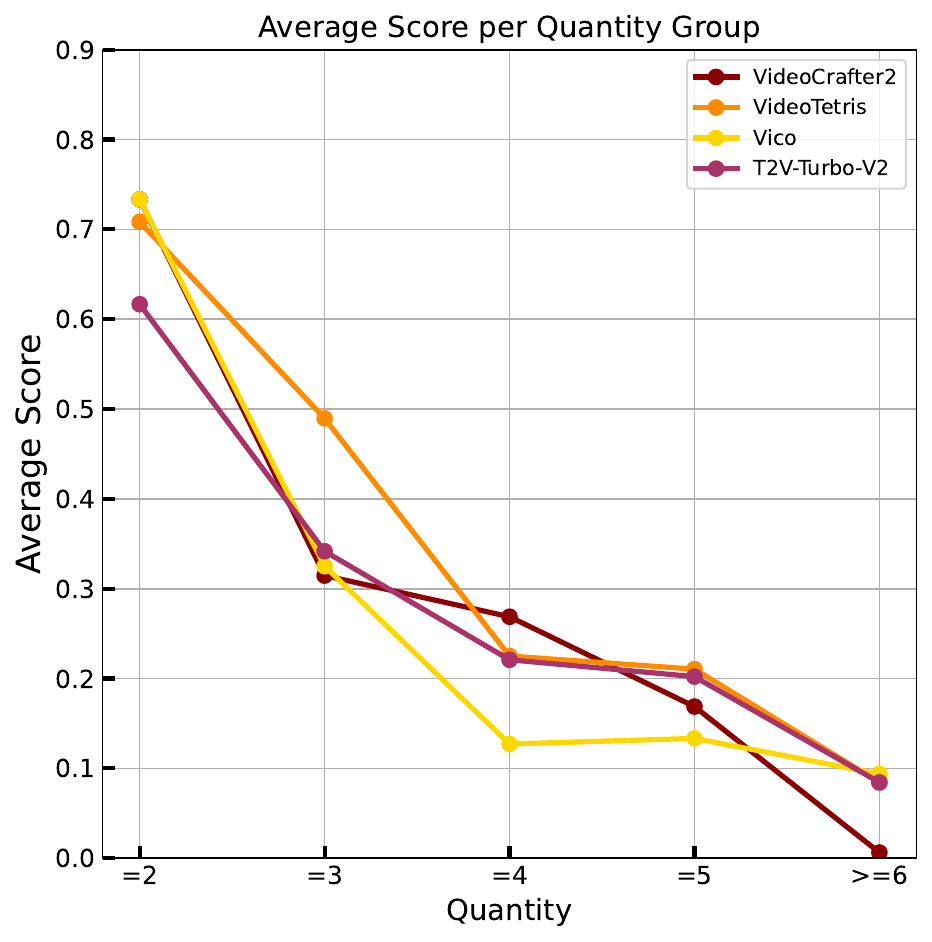}
    {\small
    \centerline{(a) diffusion unet-based models}
    }
    \label{fig:numeracy_diffusion}
\end{minipage}%
\begin{minipage}{0.33\linewidth}
    \centering
    \includegraphics[width=\textwidth]{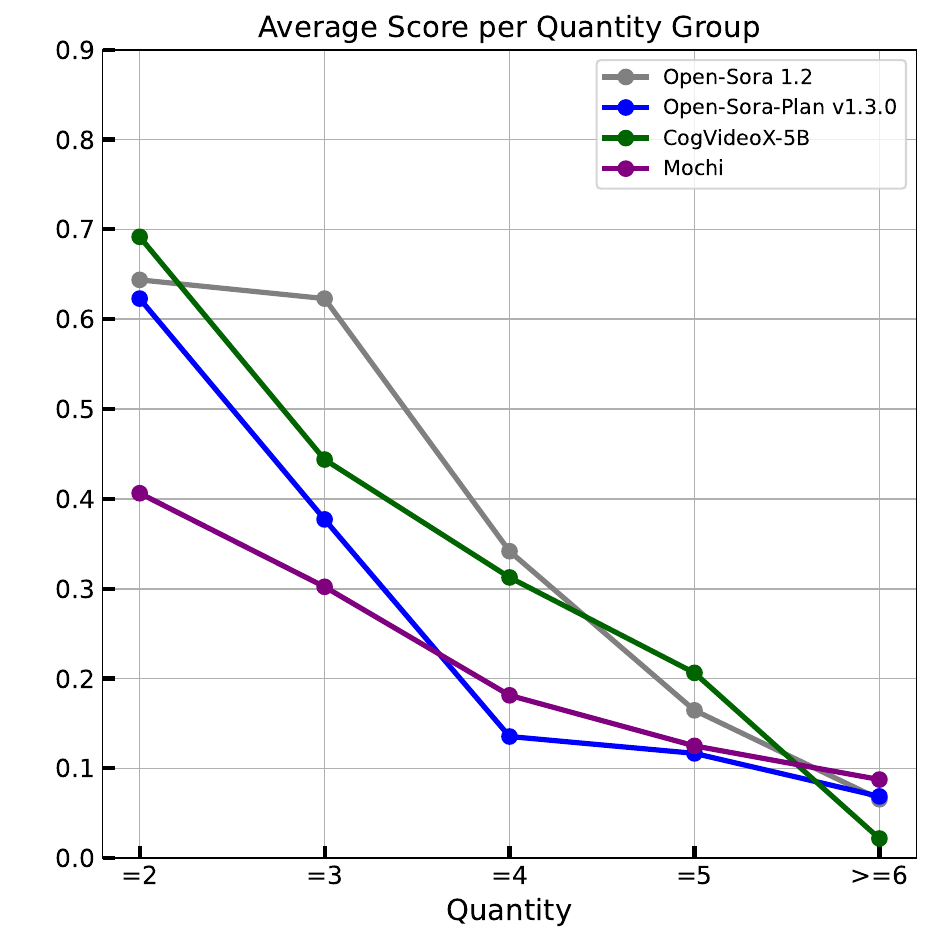}
    {\small
    \centerline{(b) DiT-based models}
    }
    \label{fig:numeracy_dit}
\end{minipage}
\begin{minipage}{0.33\linewidth}
    \centering
    \includegraphics[width=\textwidth]{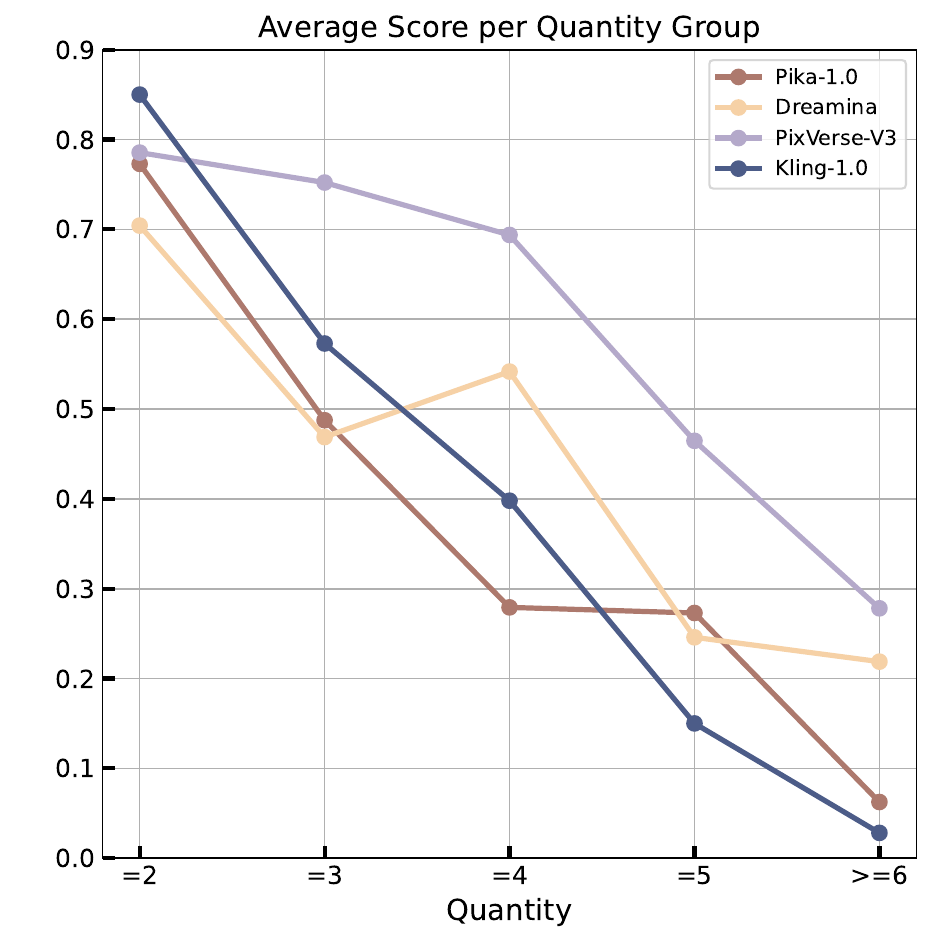}
    {\small
    \centerline{(c) commercial models}
    }
    \label{fig:numeracy_commercial}
\end{minipage}
\vspace{-10pt}
\caption{\textbf{Analysis of text-to-video generation accuracy by object quantity.} We plot the average score versus the quantity groups for videos generated by prompts with single object class in Generative Numeracy. This plot draws four T2V models from each of the following groups: (a) diffusion unet-based models, (b) DiT-based models, and (c) commercial models.}
\label{fig:analysis_numeracy}
\end{figure*}

\subsection{Spatial Relationships}
To investigate whether T2V models can understand 2D spatial relationships, we extract three key indicators from the evaluated spatial scores.
As shown in columns \#4-6 of Table~\ref{tab:scores}, the first indicator, ``Coexist'', represents the percentage of videos that successfully generate both objects involved in a specific 2D spatial relationship. A higher ``Coexist'' percentage indicates the model is better at generating multiple objects simultaneously.
The second indicator, ``Acc.'' (accuracy), reflects the percentage of videos that accurately depict the spatial relationships among all videos that successfully generate both objects. A higher ``Acc.'' suggests a greater likelihood that the model comprehends 2D spatial relationships.
The third indicator, ``Acc.Score'', is the average score of those videos with correct spatial relationships. A higher ``Acc.Score'' indicates that the two objects are more distinctly separated, leading to a clearer spatial layout. For the model to generate accurate spatial relationships, both ``Acc.'' and ``Acc.Score'' must be high.

According to the results in Table~\ref{tab:scores}, Vico~\cite{yang2024vico} and Mochi~\cite{genmo2024mochi} demonstrate the best performances in generating multiple objects among open-source models. Although LVD~\cite{lian2023llm} does not achieve a high ``Coexist'' percentage, it ranks highest in ``Acc.'' and ``Acc.Score'' among all models, which verifies its strong capability in layout planning.

\begin{figure}[h]
    \centering
    \begin{subfigure}{\linewidth}
        \centering
        \includegraphics[width=\linewidth]{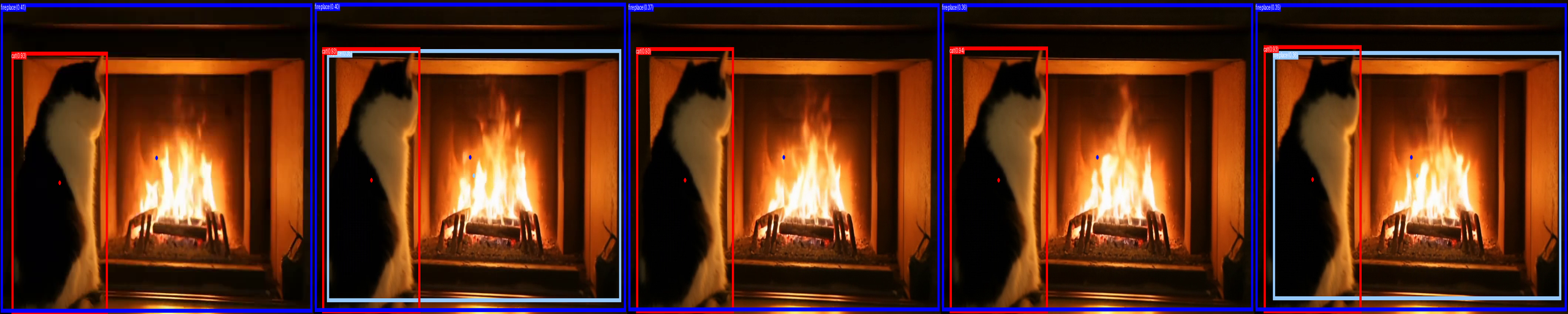} 
    \caption{Prompt:\textit{ A cat sitting on the \textbf{left} of a fireplace.} Score: 0.8394}
    \end{subfigure}
    \begin{subfigure}{\linewidth}
        \centering
        \includegraphics[width=\linewidth]{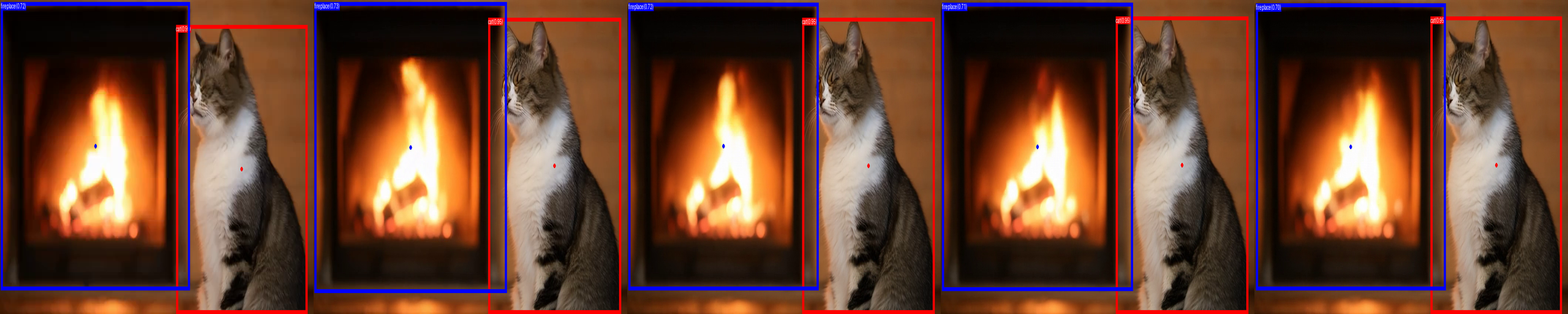} 
        \caption{Prompt:\textit{ A cat sitting on the \textbf{right} of a fireplace.} Score: 0.9706}
    \end{subfigure}
    \vspace{-15pt}
    \caption{Visualization of Spatial Relationships examples generated by Mochi~\cite{genmo2024mochi}.}
    \label{fig:analysis_3_image}
\end{figure}

Figure~\ref{fig:analysis_3_image} showcases a pair of videos generated by Mochi~\cite{genmo2024mochi}. In this example, the model accurately depicts the spatial relationship between the cat and the fireplace. Notably, the model also achieves a high score in ``Acc.''. This suggests that the model may possess some understanding of 2D spatial relationships.

\subsection{Motion Binding}
As described in Section~\ref{sec:evaluation_metrics}, motion vectors are obtained when evaluating motion binding. Therefore, we use these vectors to derive two meaningful indicators, ``Motion Level'' and ``Acc.'' for videos generated by prompts with a single object in motion binding. They are recorded in columns \#7-8 in Table~\ref{tab:scores}. We first normalize the image to a size of 100x100. ``Motion Level'' represents the displacement the object travels. Although this indicator does not account for the direction of movement, it reflects the overall motion level and the spatio-temporal dynamics of the video. ``Acc.'' in this category represents the percentage of videos in which the object moves at least 5 units in the correct direction, among all videos that successfully generate the object. 
LVD~\cite{lian2023llm} is particularly effective in planning the dynamic scene layouts, allowing it to determine the correct motion directions for objects. This is validated by its ``Acc.'' of 47\%, the highest percentage among all models. Another notable value is 42\% ``Acc.'' achieved by Gen-3~\cite{gen3}, which suggests it may have some understanding of motion direction.
Additionally, we can see from ``Motion Level'' that PixVerse-V3~\cite{PixVerse} and CogVideoX-5B~\cite{yang2024cogvideox} can generate significant object motion in their videos.

\begin{figure}[h]
    \centering
    {\scriptsize Prompt: \textit{A golden retriever scampering leftwards across a garden.}}
    \begin{subfigure}{\linewidth}
        \centering
        \includegraphics[width=\linewidth]{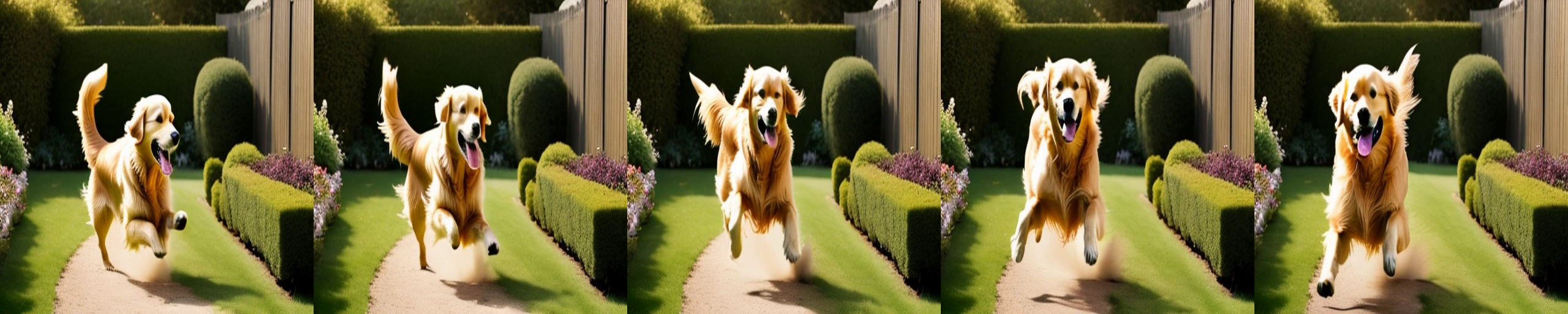} 
    \caption{T2V Model: VideoCrafter2~\cite{chen2024videocrafter2}. Score: 0.2688}
    \end{subfigure}
    \begin{subfigure}{\linewidth}
        \centering
        \includegraphics[width=\linewidth]{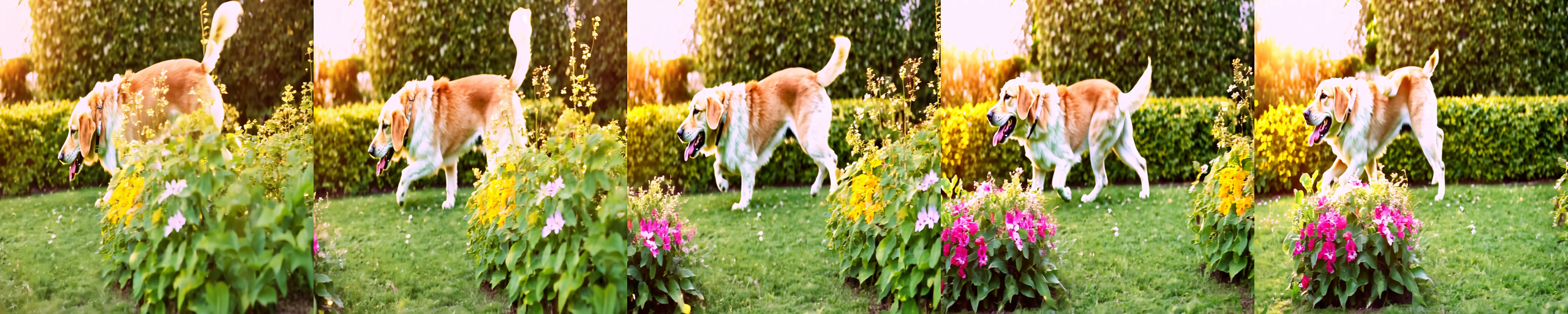} 
        \caption{T2V Model: Gen-3~\cite{gen3}. Score: 0.4381}
    \end{subfigure}
    \vspace{-15pt}
    \caption{Visualization of Motion Binding examples.}
    \vspace{-5pt}
    \label{fig:analysis_4_image}
\end{figure}

Figure~\ref{fig:analysis_4_image} compares two examples from VideoCrafter2~\cite{chen2024videocrafter2} and Gen-3~\cite{gen3}. The first features a static background with limited foreground motion, while the second displays noticeable background movement.

\subsection{Action Binding}
\begin{figure}[h]
    \centering
    {\scriptsize Prompt: \textit{A dog plays guitar while a cat takes a selfie.}}
    \begin{subfigure}{\linewidth}
        \centering
        \includegraphics[width=\linewidth]{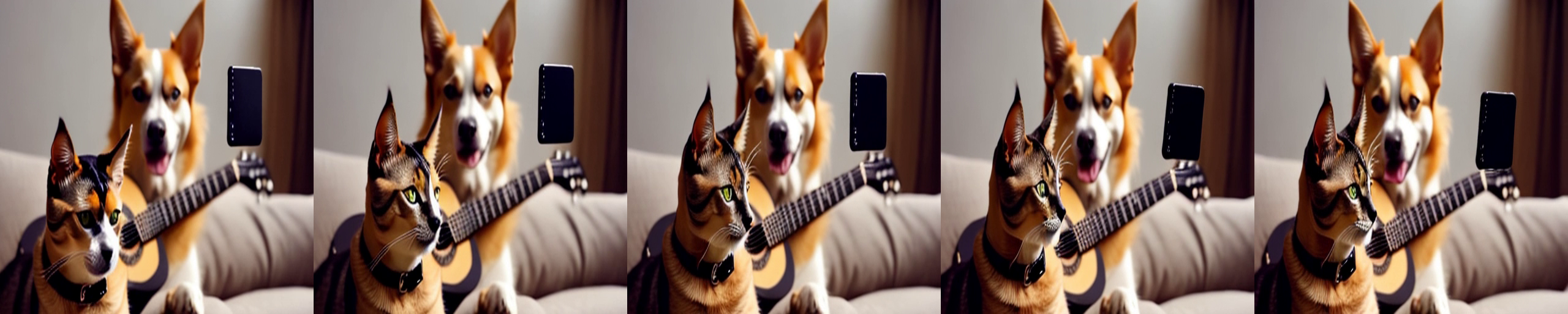} 
    \caption{T2V model: VideoCrafter2~\cite{chen2024videocrafter2}. Score: 0.7037}
    \end{subfigure}
    \begin{subfigure}{\linewidth}
        \centering
        \includegraphics[width=\linewidth]{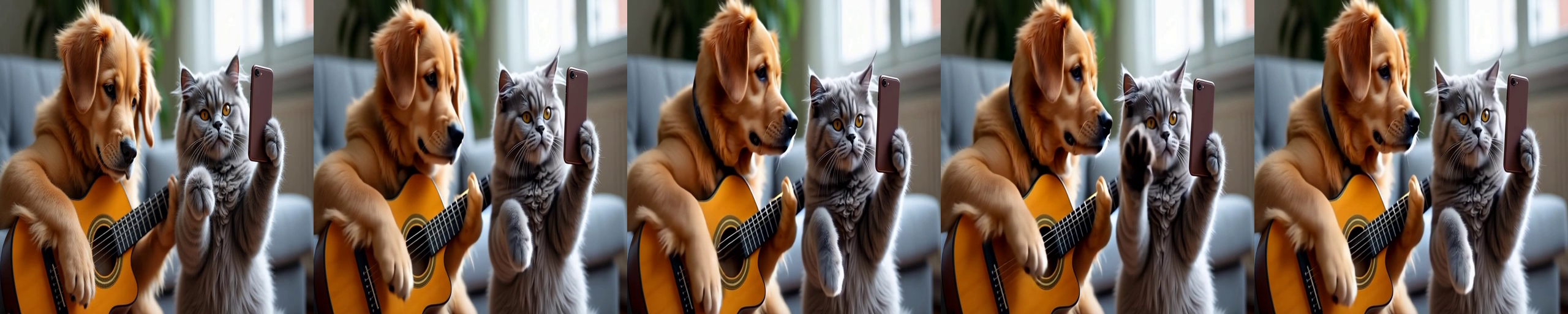} 
        \caption{T2V model: PixVerse-V3~\cite{PixVerse}. Score: 0.9630}
    \end{subfigure}
    \vspace{-15pt}
    \caption{Visualization of Action Binding examples.}
    \vspace{-5pt}
    \label{fig:analysis_5_image}
\end{figure}

Columns \#9 and \#10 in Table~\ref{tab:scores} show the average scores for the subgroups of common and uncommon prompts in the category of action binding. Some of the uncommon prompts instruct animals to perform anthropomorphic actions, which are shown in Figure~\ref{fig:analysis_5_image}. This subgroup is clearly more challenging than the common prompts because executing these uncommon actions is more difficult.

\subsection{Object Interactions}
\begin{figure}[h]
    \centering
    {\scriptsize Prompt: \textit{A skateboarder performing a trick, launching their board into the air before landing back on the pavement.}}
    \begin{subfigure}{\linewidth}
        \centering
        \includegraphics[width=\linewidth]{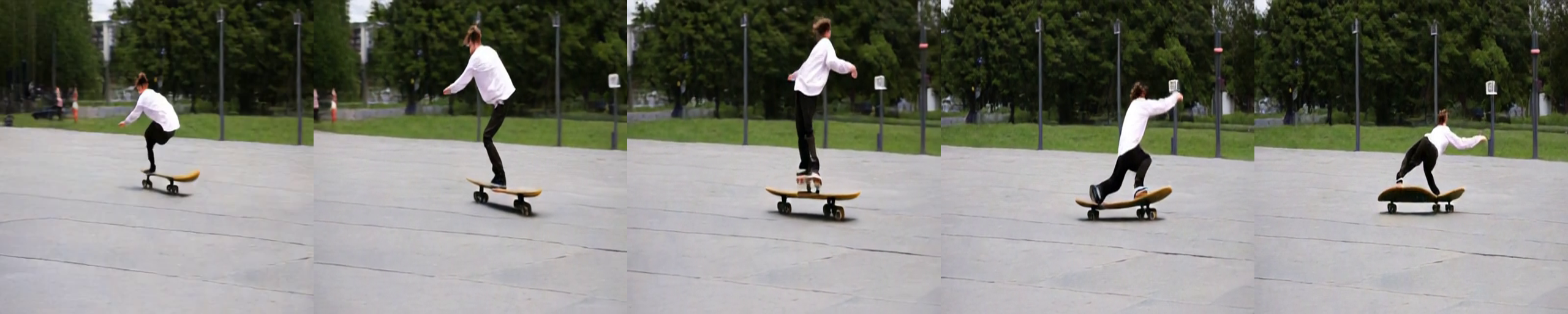} 
    \caption{T2V model: Open-Sora-Plan v1.3.0~\cite{pku_yuan_lab_and_tuzhan_ai_etc_2024_10948109}. Score: 0.5185}
    \end{subfigure}
    \begin{subfigure}{\linewidth}
        \centering
        \includegraphics[width=\linewidth]{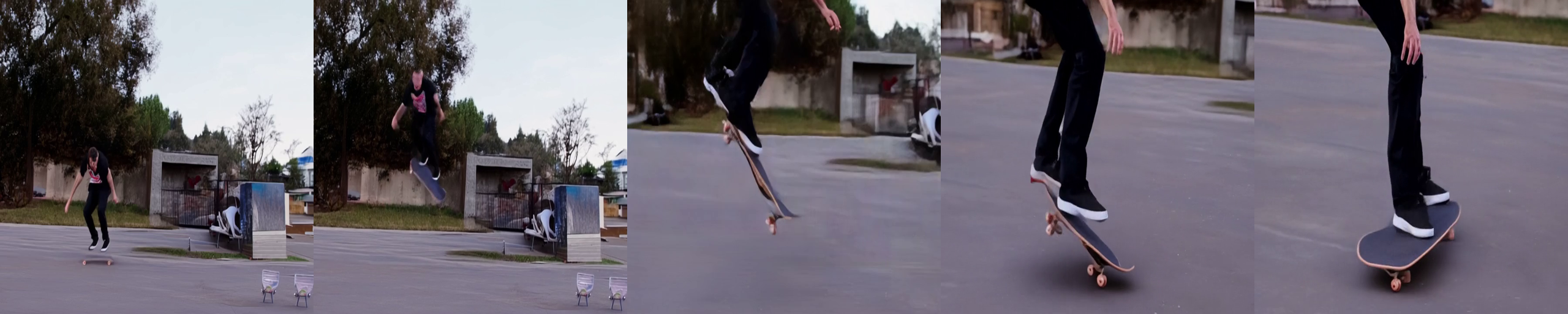} 
        \caption{T2V model: CogVideoX-5B~\cite{yang2024cogvideox}. Score: 1.000}
    \end{subfigure}
    \vspace{-15pt}
    \caption{Visualization of Object Interactions examples.}
    \vspace{-5pt}
    \label{fig:analysis_6_image}
\end{figure}

The last two columns in Table~\ref{tab:scores} record the average scores for the subgroups of physical and social interactions within the category of object interactions.
In these two subgroups, depicting physical interactions proves to be more challenging, as it requires an understanding of physical laws. Figure~\ref{fig:analysis_6_image} illustrates two examples of physical interactions generated by Open-Sora-Plan v1.3.0~\cite{pku_yuan_lab_and_tuzhan_ai_etc_2024_10948109} and VideoTetris~\cite{tian2024videotetris}. The first example fails to accurately represent the interaction process described in the prompt, while the second effectively captures both the progression and outcome of the interaction, following the prompt more accurately.

\subsection{Generative Numeracy}
Figure~\ref{fig:analysis_numeracy} illustrates how the score in generative numeracy changes in relation to the quantity of object specified in the prompt. It is evident that, as the quantity increases, the average score tends to decrease. 

Among all models, commercial models generally outperform open-source models. In particular, PixVerse-V3~\cite{PixVerse} achieves the highest scores across almost all quantity groups. The diffusion unet-based models in Figure~\hyperref[fig:numeracy_diffusion]{\ref{fig:analysis_numeracy}(a)} demonstrate comparable results in terms of numeracy. Among DiT-based models in Figure~\hyperref[fig:numeracy_dit]{\ref{fig:analysis_numeracy}(b)}, Open-Sora 1.2~\cite{opensora} stands out with the best overall performance.

\begin{figure}[h]
    \centering
    {\scriptsize Prompt: \textit{Four pandas munch bamboo in the garden.}}
    \begin{subfigure}{\linewidth}
        \centering
        \includegraphics[width=\linewidth]{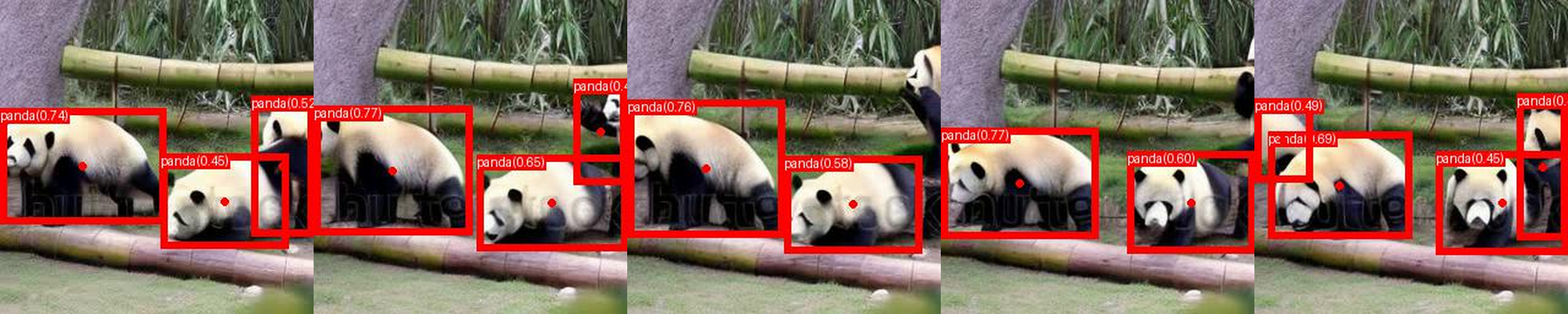} 
    \caption{T2V Model: ModelScope~\cite{wang2023modelscope}. Score: 0.0625}
    \end{subfigure}
    \begin{subfigure}{\linewidth}
        \centering
        \includegraphics[width=\linewidth]{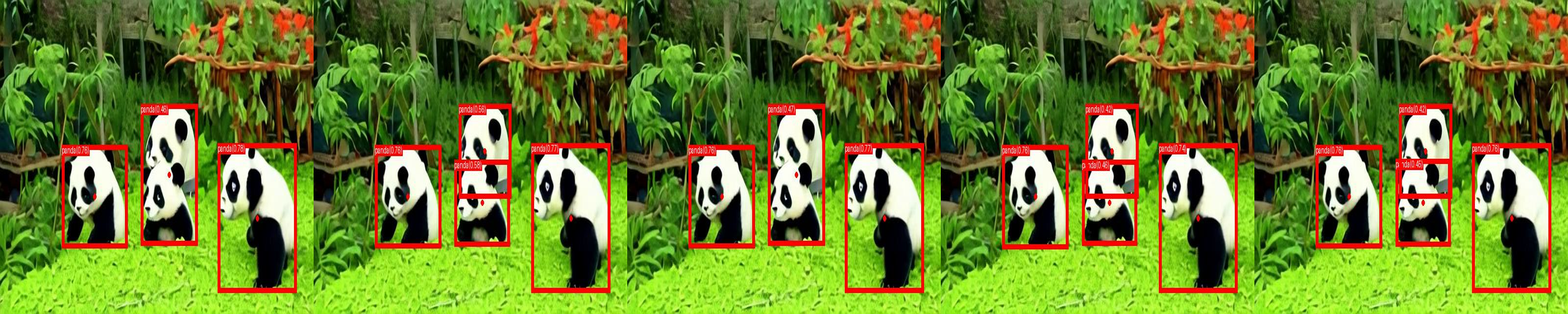} 
        \caption{T2V Model: Open-Sora 1.2~\cite{opensora}. Score: 0.8750}
    \end{subfigure}
    \vspace{-15pt}
    \caption{Visualization of Generative Numeracy examples.}
    \vspace{-5pt}
    \label{fig:analysis_7_image}
\end{figure}

Figure~\ref{fig:analysis_7_image} presents videos generated by ModelScope~\cite{wang2023modelscope} and Open-Sora 1.2~\cite{opensora}. Although the video generated by Open-Sora 1.2~\cite{opensora} is not of realistic style, it successfully represents the correct quantity of object.

\section{Human Evaluation}
\label{app:human_evaluation}
\begin{figure}[h]
    \centering
    \includegraphics[width=\linewidth]{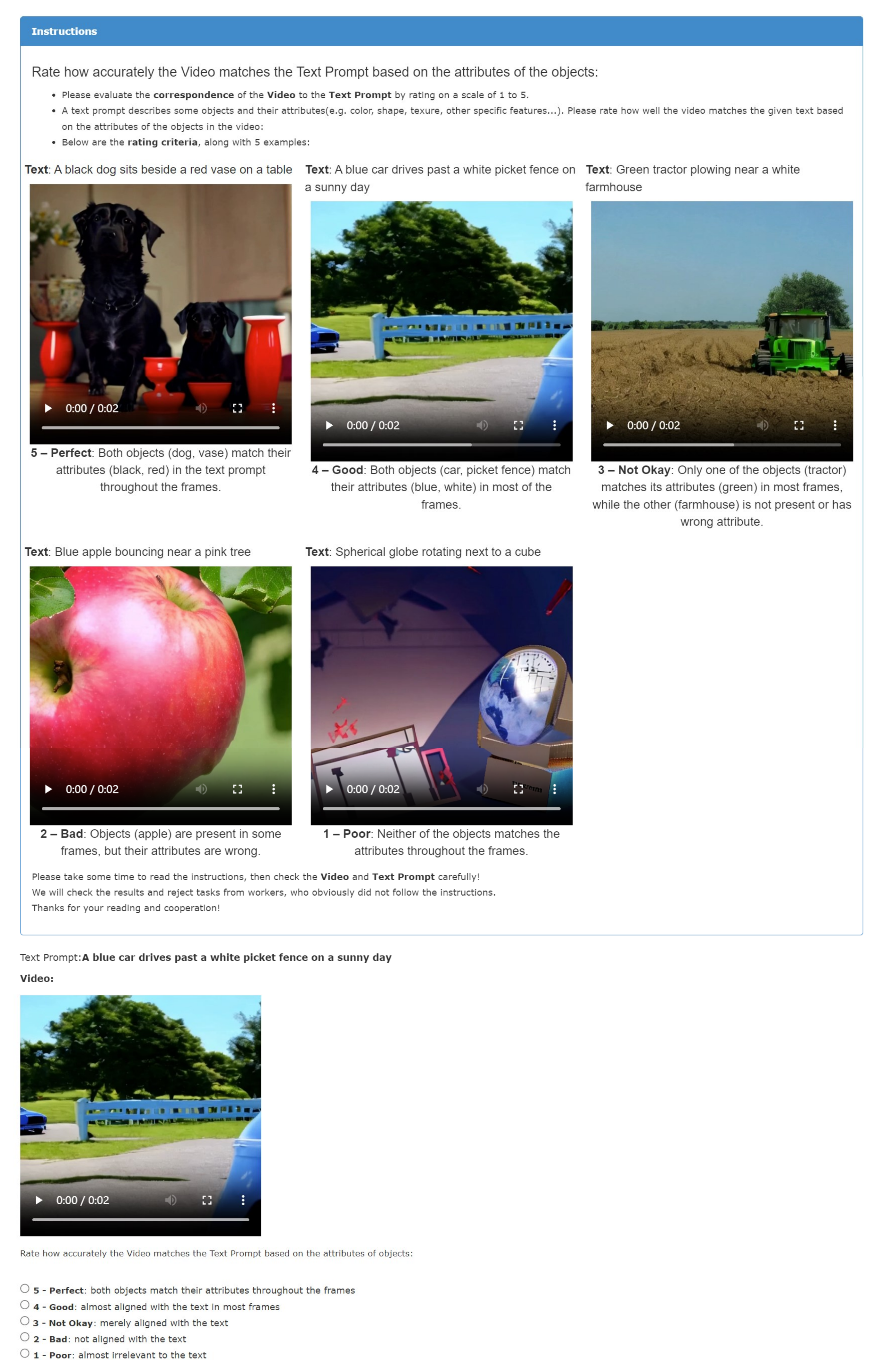}
    \vspace{0pt}
    \caption{AMT Interface for human evaluation of video-text alignment on consistent attribute binding.}
    \vspace{0pt}
    \label{mturk:consistent-attr}
\end{figure}

To assess the correlation of the scores given by T2V-CompBench metrics with human preferences, we prepare a human evaluation interface for each of the seven categories. The human evaluations are conducted on Amazon Mechanical Turk (AMT). Figure~\ref{mturk:consistent-attr} shows the interface for consistent attribute binding.\\

\noindent \textbf{Annotation Instruction}.
Each interface includes several key components. First, we clarify the evaluation dimension on which this category is focusing. 
At the top of each interface, we thoroughly explain the key information in the prompt and the specific dimension of interest. For example, in consistent attribute binding, annotators are instructed to focus solely on the objects and attributes.
Next, we provide clear rating criteria along with examples in this category. For each prompt-video example, we provide a detailed rationale for the assigned score based on the specific objects and attributes depicted in the video. Annotators may refer to these examples at any time when they are uncertain about the appropriate rating.
Finally, we present the video-text pair that needs evaluation, along with the scoring options (ranging from 5 to 1). To reinforce the rating criteria for the annotators, we include a concise summary of the criteria following each score. \\

\noindent \textbf{Strategies for Ensuring Quality}.
To ensure the quality of human evaluations, we employ several strategies:
\begin{itemize}
    \item Interface Notification: We include a note in the interface informing annotators that we will review the evaluation results and reject tasks from workers who obviously do not follow the instructions.
    \item Random Sampling: We randomly sample 20\% of the total tasks completed by each worker. Any task that evidently fails to follow the instructions will be rejected. If an annotator is rejected more than five times, we will block that individual from further participation, and the tasks completed by that annotator will be reassigned for evaluation.
    \item Multiple Raters: To balance different opinions, we require three human annotators to rate each task. The average score from these ratings is then assigned to the corresponding video-text pair.
    \item Selection of Annotators: We only select experienced and responsible AMT workers by establishing a high historical task acceptance threshold of 90\%.
    \item Compensation: The estimated hourly wage for each participant is 9.60 USD. We have communicated with several annotators, and they agree that this wage is sufficiently competitive to encourage reliable task ratings. In total, we spent 195 USD on participant compensation.
\end{itemize}

These measures collectively enhance the reliability and accuracy of our human evaluations.

\section{Societal Impacts}
\label{app:societal_impacts}
The community should be aware of the potential negative social impact that can arise from the misuse of video generation models, particularly in generating misleading or harmful content, which could exacerbate issues related to misinformation and deepfakes. Additionally, the biases inherent in the training data may lead to the perpetuation of stereotypes or exclusion of underrepresented groups, thereby influencing public perception and societal norms. Therefore, it is imperative that the evaluation of T2V models not only assesses their technical performances but also considers broader social implications, ensuring that these technologies contribute positively to society while mitigating potential risks. To this end, we plan to incorporate an evaluation dimension focused on unbiased composition in the future.

\section{Limitations and Future Work}
\label{app:limitations}
While we have made the first step in evaluating compositional T2V generation, there are still many challenges:

\begin{itemize}
\item A limitation of our work is the lack of a unified evaluation metric for all categories, and we believe that this limitation points out new challenges for better and larger multimodal LLMs or video understanding models.  If such models become available, we can utilize them as our evaluation metrics.

\item Our benchmark aims to evaluate videos within 2 to 5 seconds. For categories other than motion binding, we sample a fixed number of frames for evaluation, which may not be sufficient for videos longer than 5 seconds. In motion binding, longer videos may lead to greater object displacement and better performance. We leave the evaluation of long videos for future work.

\end{itemize}

\section{Templates for Generating Prompts and Metadata Using GPT-4}
\label{app:template}
This section provides the instructions used to prompt GPT-4~\cite{openai2024gpt4} to generate the text prompts for T2V-CompBench and the corresponding metadata for evaluation. The text prompts in T2V-CompBench and metadata generated for evaluation are available in the \href{https://github.com/KaiyueSun98/T2V-CompBench/tree/V2}{code repository}.

Table~\ref{tab:template_prompt_1},
~\ref{tab:template_prompt_2}, ~\ref{tab:template_prompt_3}, ~\ref{tab:template_prompt_4}, ~\ref{tab:template_prompt_5}, ~\ref{tab:template_prompt_6}, ~\ref{tab:template_prompt_7}
are the input templates used to generate the prompts for the seven categories, each with specific requirements. 

Table~\ref{tab:template_metadata_1},
~\ref{tab:template_metadata_2}, ~\ref{tab:template_metadata_3}, ~\ref{tab:template_metadata_4}, ~\ref{tab:template_metadata_5}, ~\ref{tab:template_metadata_7}
are the templates used to generate the metadata for evaluation.

\begin{table*}[!h]
\centering
\begin{tcolorbox}[colframe=black, colback=gray!20, arc=5mm, boxrule=0.5mm, width=\textwidth]
\begin{tabular}{p{\linewidth}}
\texttt{<USER>:} This is a list of object nouns: {[\textcolor{cyan}{{object nouns}}]}. Please reply `Yes' if received. \\ 
\texttt{<ASSISTANT>:} Yes\\
\texttt{<USER>:} This is a list of color attributes: {[\textcolor{cyan}{color attributes}]}. Please reply `Yes' if received. \\
\texttt{<ASSISTANT>:} Yes\\
\texttt{<USER>:} This is a list of active verbs: {[\textcolor{cyan}{verbs}]}. Please reply `Yes' if received. \\
\texttt{<ASSISTANT>:} Yes\\
\texttt{<USER>:} Please generate 50 prompts featuring two objects with color attributes. \\

Requirements: \\
1. Specify a unique color for each object in the prompt to enrich the visual narrative. Use the color attributes in the color attributes list provided above. \\
2. Select object nouns that have multiple possible colors from the list of object nouns provided above.\\
3. The objects can engage in separate or shared activities. Describe them using vibrant, active verbs, avoiding static verbs such as `see', `rest', or `park'. Select verbs from the list of active verbs and feel free to adjust their tense.\\

Here are some examples:\\
A blue car drives past a white picket fence on a sunny day.\\
Yellow rubber duck floating next to a blue bath towel.
\end{tabular}
\end{tcolorbox}
\caption{\textbf{Template for generating consistent attribute binding prompts using GPT-4.} The attribute type in the template can be replaced by shape or texture; here we take color as an example. The object nouns, color attributes, and verbs marked in \textcolor{cyan}{cyan} should be replaced with words derived from real user prompts.}
\label{tab:template_prompt_1}
\end{table*}

\begin{table*}[!h]
\centering
\begin{tcolorbox}[colframe=black, colback=gray!20, arc=5mm, boxrule=0.5mm, width=\textwidth]
\begin{tabular}{p{\linewidth}}

\texttt{<USER>:} Here are some prompts that describe  color \& light change, please generation 30 prompts by referring these examples:\\
The leaf is turning from green to yellow.\\
The sky is shifting from bright to dark.
\end{tabular}
\end{tcolorbox}
\caption{\textbf{Template for generating dynamic attribute binding prompts using GPT-4.} The attribute type in the template can be replaced by shape \& size change, texture change or combined change; here we take color \& light change as an example. The examples are sourced from TempCompass~\cite{liu2024tempcompass}.}
\label{tab:template_prompt_2}
\end{table*}

\begin{table*}[!h]
\centering
\begin{tcolorbox}[colframe=black, colback=gray!20, arc=5mm, boxrule=0.5mm, width=\textwidth]
\begin{tabular}{p{\linewidth}}
\texttt{<USER>:} This is a list of object nouns: {[\textcolor{cyan}{{object nouns}}]}. Please reply `Yes' if received. \\ 
\texttt{<ASSISTANT>:} Yes\\
\texttt{<USER>:} This is a list of active verbs: {[\textcolor{cyan}{verbs}]}. Please reply `Yes' if received. \\
\texttt{<ASSISTANT>:} Yes\\
\texttt{<USER>:} Please generate 50 pairs of prompts featuring two objects where, in each pair, one object is described as being \textcolor{cyan}{on the left of} the other in the first prompt, and the second prompt should state that the same object is \textcolor{cyan}{on the right of} the other.\\

Requirements: \\
1. Select object nouns from the list provided above.\\
2. The prompt must contain active verbs to ensure the scene is dynamic. You can select verbs from the list of active verbs and feel free to adjust their tense. \\
Here are some examples:\\
`A dog running on the left of a bicycle' and `A dog running on the right of a bicycle'.\\
`A llama walking on the left side of a motorcycle in a city street' and `A llama walking on the right side of a motorcycle in a city street'.
\end{tabular}
\end{tcolorbox}
\caption{\textbf{Templates for generating spatial relationships prompts using GPT-4.} 
The spatial relationship in the template can be replaced by above\&below or in front of\&behind; here we take left\&right as an example. The object nouns and verbs marked in \textcolor{cyan}{cyan} should be replaced with words derived from real user prompts.
}
\label{tab:template_prompt_3}
\end{table*}

\begin{table*}[!h]
\centering
\begin{tcolorbox}[colframe=black, colback=gray!20, arc=5mm, boxrule=0.5mm, width=\textwidth]
\begin{tabular}{p{\linewidth}}
\texttt{<USER>:} This is a list of object nouns: {[\textcolor{cyan}{{object nouns}}]}. Please reply `Yes' if received. \\ 
\texttt{<ASSISTANT>:} Yes\\
\texttt{<USER>:} Please generate 50 pairs of prompts where, in each pair, an object is being describe as \textcolor{cyan}{moving leftwards} in the first prompt and the second prompt should state the same object is \textcolor{cyan}{moving rightwards}.

Requirements: \\
1. Select object nouns from the list provided above.\\
2. Describe the objects using vibrant, active verbs.\\

Here are some examples:\\
`A golden retriever scampering leftwards across a garden' and `A golden retriever scampering rightwards across a garden'.\\
`A football rolling from the right to the left on the grass' and `A football rolling from the left to the right on the grass'.
\end{tabular}
\end{tcolorbox}
\caption{\textbf{Templates for generating motion binding prompts using GPT-4.} The moving direction in the template can be replaced by upwards\&downwards; here we take leftward\&rightwards as an example. The object nouns marked in \textcolor{cyan}{cyan} should be replaced with words derived from real user prompts.}
\label{tab:template_prompt_4}
\end{table*}

\begin{table*}[!h]
\centering
\begin{tcolorbox}[colframe=black, colback=gray!20, arc=5mm, boxrule=0.5mm, width=\textwidth]
\begin{tabular}{p{\linewidth}}
\texttt{<USER>:} This is a list of object nouns: {[\textcolor{cyan}{{object nouns}}]}. Please reply `Yes' if received. \\ 
\texttt{<ASSISTANT>:} Yes\\
\texttt{<USER>:} This is a list of active verbs: {[\textcolor{cyan}{verbs}]}. Please reply `Yes' if received. \\
\texttt{<ASSISTANT>:} Yes\\
\texttt{<USER>:} Please generate 50 prompts that describe an object engaging in an activity, while another object is engaging in a different activity.  \\
Requirements: \\
1. Select the two object nouns from the list of object nouns provided above.\\
2. Specify a unique verb for each object in the prompt.\\
3. Use vibrant, active verbs, avoiding static verbs such as `see', `rest', or `park'. You can select verbs from the list of active verbs and feel free to adjust their tense.\\

Here are some examples:\\
A dog runs through a field while a cat climbs a tree.\\
A man takes photos and a boy dances on the street.
\end{tabular}
\end{tcolorbox}
\caption{\textbf{Templates for generating action binding prompts using GPT-4.} The object nouns, marked in \textcolor{cyan}{cyan} should be replaced with words derived from real user prompts.}
\label{tab:template_prompt_5}
\end{table*}

\begin{table*}[!h]
\centering
\begin{tcolorbox}[colframe=black, colback=gray!20, arc=5mm, boxrule=0.5mm, width=\textwidth]
\begin{tabular}{p{\linewidth}}
\texttt{<USER>:} This is a list of object nouns: {[\textcolor{cyan}{{object nouns}}]}. Please reply `Yes' if received. \\ 
\texttt{<ASSISTANT>:} Yes\\
\texttt{<USER>:} Please generate 50 prompts describing physical interaction between two objects that can change their state of motion.\\

Requirements: \\
1. For the two objects involved in the interaction, select at least one object noun from the list of object nouns provided above.\\
2. Use vibrant, active verbs, avoiding static verbs such as `see', `rest', or `park'.\\

Here are some examples:\\
Two cars collide at an intersection.\\
A dog dragging a blanket off a bed, leaving it tangled on the floor.
\end{tabular}
\end{tcolorbox}
\caption{\textbf{Templates for generating object interactions prompts using GPT-4.} The physical interaction in the template can be replaced by social interaction; here we take physical interaction as an example. The object nouns marked in \textcolor{cyan}{cyan} should be replaced with words derived from real user prompts.}
\label{tab:template_prompt_6}
\end{table*}

\begin{table*}[!h]
\centering
\begin{tcolorbox}[colframe=black, colback=gray!20, arc=5mm, boxrule=0.5mm, width=\textwidth]
\begin{tabular}{p{\linewidth}}
\texttt{<USER>:} This is a list of object nouns: {[\textcolor{cyan}{{object nouns}}]}. Please reply `Yes' if received. \\ 
\texttt{<ASSISTANT>:} Yes\\
\texttt{<USER>:} This is a list of active verbs: {[\textcolor{cyan}{verbs}]}. Please reply `Yes' if received. \\
\texttt{<ASSISTANT>:} Yes\\
\texttt{<USER>:} Please generate one prompt for each object in the list {[\textcolor{cyan}{{object nouns}}]} with a quantity of \textcolor{cyan}{two}. Use vibrant and active verbs in the prompt. You can select verbs from the list provided above and feel free to adjust their tense.\\
Here are some examples:\\
Two cats walking on the floor. \\
Two dogs running through a field.
\end{tabular}
\end{tcolorbox}
\caption{\textbf{Templates for generating numeracy prompts using GPT-4.} The quantity of `two' in the template can be replaced by `three', `four', `five', `six', `seven' and `eight'; here we take `two' as an example. The object nouns marked in \textcolor{cyan}{cyan} should be replaced with words derived from real user prompts, here we restrict the length of the list to 30.}
\label{tab:template_prompt_7}
\end{table*}

\begin{table*}[!h]
\centering
\begin{tcolorbox}[colframe=black, colback=blue!5, arc=5mm, boxrule=0.5mm, width=\textwidth]
\begin{tabular}{p{\linewidth}}
\texttt{<USER>:} I will provide you with several prompts, each on a separate line. Each prompt will
describe two objects, each associated with an attribute. For each prompt, you are to create a JSON object. This JSON object should include two keys: ``prompt'' and ``phrases''. \\
The ``prompt'' key should simply replicate the original prompt text. The ``phrases'' key should contain the two described objects separated by a semicolon.\\
For instance, for the prompt ``A blue car drives past a white picket fence on a sunny day'', the value under ``phrases'' should be ``a blue car; a white picket fence''. \\
Below are the prompts for which you need to generate the JSON objects:\\
A blue car drives past a white picket fence on a sunny day.\\
Yellow rubber duck floating next to a blue bath towel.\\
......\\
\rule[0.25\baselineskip]{\textwidth}{1pt}
\texttt{<ASSISTANT>:} Here are the JSON representations for the prompts:\\
\{``prompt'': ``A blue car drives past a white picket fence on a sunny day'', ``phrases'': ``a blue car; a white picket fence''\}, \\
\{``prompt'': ``Yellow rubber duck floating next to a blue bath towel'', ``phrases'': ``yellow rubber duck; a blue bath towel''\}, \\
......\\
\end{tabular}
\end{tcolorbox}
\caption{
\textbf{Template for generating the metadata of consistent attribute binding prompts using GPT-4.}
}
\label{tab:template_metadata_1}
\end{table*}

\begin{table*}[!h]
\centering
\begin{tcolorbox}[colframe=black, colback=blue!5, arc=5mm, boxrule=0.5mm, width=\textwidth]
\begin{tabular}{p{\linewidth}}
\texttt{<USER>:} I will provide you with several prompts, each on a separate line. Each prompt describes a color, shape, texture or combined change of one or multiple objects. For each prompt, you are to create a JSON object. This JSON object should include the following keys:\\
``prompt'': simply replicate the original prompt text.\\
``state 0'': the object(s) with initial states.\\
``state 1'': the object(s) with final states.\\
For instance, for the prompt ``The sky is shifting from bright to dark'',  ``state 0'' should be ``bright sky'',  ``state 1'' should be ``dark sky''. For the prompt ``a leaf is transitioning from green to yellow and a fresh apple is decaying'', ``state 0'': ``a green leaf and a fresh apple''; ``state 1'': ``a yellow leaf a decayed apple''.\\
Below are the prompts for which you need to generate the JSON objects:\\
The sky is shifting from bright to dark.\\
The leaf is turning from green to yellow.\\
......\\
\rule[0.25\baselineskip]{\textwidth}{1pt}
\texttt{<ASSISTANT>:} Here are the JSON representations for the prompts:\\
\{``prompt'': ``The sky is shifting from bright to dark'', ``state 0'': ``bright sky'', ``state 1'': ``dark sky''\}, \\
\{``prompt'': ``The leaf is turning from green to yellow'', ``state 0'': ``green leaf'', ``state 1'': ``yellow leaf''\}, \\
......\\
\end{tabular}
\end{tcolorbox}
\caption{
\textbf{Template for generating the metadata of dynamic attribute binding prompts using GPT-4.}
}
\label{tab:template_metadata_2}
\end{table*}

\begin{table*}[!h]
\centering
\begin{tcolorbox}[colframe=black, colback=blue!5, arc=5mm, boxrule=0.5mm, width=\textwidth]
\begin{tabular}{p{\linewidth}}
\texttt{<USER>:} I will provide you with several prompts, each on a separate line. For each prompt, you are to create a JSON object. This JSON object should include the following keys:\\ ``prompt'': simply replicate the original prompt text. \\
``spatial'': contain the word from the prompt that represents the spatial relationship. Choose one from the following six options: left, right, above, below, in front of, and behind. 
For example, if the prompt uses the word ``under'',  set the value of ``spatial'' to ``below'' because these words have the same meaning.\\
``object\_1'' and ``object\_2'': These keys represent the two objects mentioned in the prompt. The value of ``object\_1'' should be positioned to the left of, right of, above, below, in front of, or behind ``object\_2'' based on the spatial word used in the prompt.\\
Here's an example to illustrate the format:\\
Prompt: ``A toddler walking on the left of a dog in a park''\\
JSON object: \{``prompt'': ``A toddler walking on the left of a dog in a park'', ``spatial'': ``left'', ``object\_1'': ``toddler'', ``object\_2'': ``dog''\}\\
Below are the prompts for which you need to generate the JSON objects:\\
A bird flying on the left side of a hot air balloon.\\
A llama walking on the left side of a motorcycle in a city street.\\
......\\
\rule[0.25\baselineskip]{\textwidth}{1pt}
\texttt{<ASSISTANT>:} Here are the JSON objects for each of your prompts, accurately capturing the spatial
relationship, object\_1, and object\_2:\\
\{``prompt'': ``A bird flying on the left side of a hot air balloon.'', ``spatial'': ``left'', ``object\_1'': ``bird'', ``object\_2'': ``hot air balloon''\}, \\
\{``prompt'': ``A llama walking on the left side of a motorcycle in a city street.'', ``spatial'': ``left'', ``object\_1'': ``llama'', ``object\_2'': ``motorcycle''\}, \\
......\\
\end{tabular}
\end{tcolorbox}
\caption{
\textbf{Template for generating the metadata of spatial relationships prompts using GPT-4.}
}
\label{tab:template_metadata_3}
\end{table*}

\begin{table*}[!h]
\centering
\begin{tcolorbox}[colframe=black, colback=blue!5, arc=5mm, boxrule=0.5mm, width=\textwidth]
\begin{tabular}{p{\linewidth}}
\texttt{<USER>:} I will provide you with several prompts, each on a separate line. Each prompt describes the movement of one or two objects in different directions.
For each prompt, you are to create a JSON object. 
This JSON object should include the following keys:\\
``prompt'': simply replicate the original prompt text.\\
``object\_1'': the first object mentioned in the prompt that moves in a direction.\\
``d\_1'': motion direction for ``object\_1'' and can only be one of the following: left, right, up, or down.\\
``object\_2'': If there is a second object mentioned in the prompt that moves in a direction, this key will contain its name. Otherwise, the value will be an empty string.\\
``d\_2'': If there is ``object\_2'', this key will represent its motion direction. Otherwise, the value will be an empty string.

Below are the prompts for which you need to generate the JSON objects:

A golden retriever scampering leftwards across a garden\\
A ball rolls left while a puppy dashes right\\
......\\
\rule[0.25\baselineskip]{\textwidth}{1pt}
\texttt{<ASSISTANT>:} Here are the JSON representations for each prompt:
 
 \{``prompt'': ``A golden retriever scampering leftwards across a garden'', ``object\_1'': ``golden retriever'', ``d\_1'': ``left'', ``object\_2'': ``'', ``d\_2'': ``''\}

\{``prompt'': ``A ball rolls left while a puppy dashes right'', ``object\_1'': ``ball'', ``d\_1'': ``left'', ``object\_2'': ``puppy'', ``d\_2'': ``right''\},\\
......\\
\end{tabular}
\end{tcolorbox}
\caption{
\textbf{Template for generating the metadata of motion binding prompts using GPT-4.}
}
\label{tab:template_metadata_4}
\end{table*}

\begin{table*}[!h]
\centering
\begin{tcolorbox}[colframe=black, colback=blue!5, arc=5mm, boxrule=0.5mm, width=\textwidth]
\begin{tabular}{p{\linewidth}}
\texttt{<USER>:} I will provide you with several prompts, each on a separate line. For each prompt, 
your task is to extract the prompt into a noun phrase and a noun phrase with an action.\\
For example, given the prompt ``A dog runs through a field while a cat climbs a tree'', the output should be: 
\{``prompt'': ``A dog runs through a field while a cat climbs a tree'', ``phrase\_0'': [``a dog?'', ``a dog runs through a field?''], ``phrase\_1'': [``a cat?'', ``a cat climbs a tree?'']\}.

Below are the prompts for which you need to generate the JSON objects:

A dog runs through a field while a cat climbs a tree.

A man takes photos and a boy dances on the street.

......\\
\rule[0.25\baselineskip]{\textwidth}{1pt}
\texttt{<ASSISTANT>:} Here are the extracted phrases for each prompt:
 
\{``prompt'': ``A dog runs through a field while a cat climbs a tree'', ``phrase\_0'': [``a dog?'', ``a dog runs through a field?''], ``phrase\_1'': [``a cat?'', ``a cat climbs a tree?'']\},

\{``prompt'': ``A man takes photos and a boy dances on the street'', ``phrase\_0'': [``a man?'', ``a man takes photos?''], ``phrase\_1'': [``a boy?'', ``a boy dances on the street?'']\},

......\\
\end{tabular}
\end{tcolorbox}
\caption{
\textbf{Template for generating the metadata of action binding prompts using GPT-4.}
}
\label{tab:template_metadata_5}
\end{table*}

\begin{table*}[!h]
\centering
\begin{tcolorbox}[colframe=black, colback=blue!5, arc=5mm, boxrule=0.5mm, width=\textwidth]
\begin{tabular}{p{\linewidth}}
\texttt{<USER>:} I will provide you with several prompts, each on a separate line.
Each prompt describes a certain number of objects engaging in an activity, where each object is associated with a specific quantity. For example, the prompt may include phrases like ``two cat'' or ``three dogs''.

For each prompt, you are to create a JSON object. 
This JSON object should include the following keys:

``prompt'': simply replicate the original prompt text.

``objects'': This key will contain the objects mentioned in the prompt. If there are multiple objects, they will be separated by commas.

``numbers'': This key will represent the corresponding numbers associated with each object. The numbers will be concatenated with commas.

When describing the action the objects are performing and the environment, there may be mentions of other objects, articles like ``a road'', ``the woods'' or other unnecessary details. Please ignore these and focus only on extracting the main objects and their respective quantities.

Below are the prompts for which you need to generate the JSON objects:

Three cars speed along a road.

Two pandas munch on bamboo, and three elephants spray water.

......\\
\rule[0.25\baselineskip]{\textwidth}{1pt}
\texttt{<ASSISTANT>:} Here are the JSON representations for each prompt:
 
\{``prompt'': ``Three cars speed along a road'', ``objects'': ``car'', ``numbers'': ``3''\},

\{``prompt'': ``Two pandas munch on bamboo, and three elephants spray water'', ``objects'': ``panda,elephants'', ``numbers'': ``2,3''\},

......\\
\end{tabular}
\end{tcolorbox}
\caption{
\textbf{Template for generating the metadata of generative numeracy prompts using GPT-4.}
}
\label{tab:template_metadata_7}
\end{table*}

\end{document}